\documentclass[11pt]{article}

\usepackage{microtype}
\usepackage{inconsolata}
\usepackage{times}
\usepackage{latexsym}
\usepackage[T1]{fontenc}
\usepackage[utf8]{inputenc}
\usepackage[final]{acl}
\usepackage{graphicx}
\usepackage{subcaption}
\usepackage{booktabs} 
\usepackage{tcolorbox}
\usepackage{comment}
\usepackage{amsmath}
\usepackage{amssymb}
\usepackage{mathtools}
\usepackage{amsthm}
\usepackage[capitalize,noabbrev]{cleveref}
\usepackage{multirow}
\usepackage{threeparttable}

\title{In-Context Learning Beyond Copying: A Training-Time Analysis of Abstractive ICL}

\author{
  \textbf{Kerem Şahin\textsuperscript{1,*}},
  \textbf{Sheridan Feucht\textsuperscript{1}},
  \textbf{Adam Belfki\textsuperscript{1}},
  \textbf{Jannik Brinkmann\textsuperscript{2}},
\\
  \textbf{Aaron Mueller\textsuperscript{3}},
  \textbf{David Bau\textsuperscript{1}},
  \textbf{Chris Wendler\textsuperscript{1}}
\\
\\
  \textsuperscript{1}Northeastern University,
  \textsuperscript{2}University of Mannheim,
  \textsuperscript{3}Boston University
\\
}

\begin{document}
\maketitle
{\renewcommand\thefootnote{}\footnotetext{\textsuperscript{*}Correspondence to
\href{mailto:sahin.ke@northeastern.edu}{\texttt{sahin.ke@northeastern.edu}}. Code, model
checkpoints, and evaluation data can be found at \url{https://hapax.baulab.info}}}
\begin{abstract}
    The sudden emergence of induction circuits and the ability to perform in-context copying in LLMs coincides with a sharp drop in training loss over a short span of the training process. This co-occurrence has been cited as evidence that it marks the point at which a wide range of in-context learning (ICL) capabilities begin to emerge. To analyze the training dynamics of few-shot abstractive ICL capabilities (i.e., tasks where the answer is not contained in the input context), we propose \textsc{Hapax}, a training regime that omits the loss contribution of tokens predictable by the induction circuit to control training pressure for copying. Using this intervention, we find that abstractive ICL capabilities can still emerge, despite a substantial reduction in inductive copying and the absence of the characteristic phase changes observed during Vanilla training. Analysis of attention heads shows that \textsc{Hapax}-trained models develop fewer and weaker induction heads. Together, these results suggest that the developmental link between emergent phenomena associated with induction circuits and the emergence of abstractive ICL is weaker than previously hypothesized. More broadly, \textsc{Hapax} provides a controlled framework for analyzing how copying-related training signals shape the development of in-context learning.
    
\end{abstract}

\section{Introduction}

Language modeling is fundamentally repetitive: in a coherent document, many sequences appear more than once, like ``the Dursleys'' in Harry Potter, or ``public static void'' in Java code. Words that \textit{do} appear only once in a given text are so rare that corpus linguists give them a special name, \textit{hapax legomena}, transliterated from Ancient Greek for ``said once''. Prior work has identified \textit{induction heads} \citep{elhage2021mathematical}, attention heads that exploit this repetition by matching patterns from earlier context and copying their continuations. \citet{olsson2022context} observed that the emergence of induction circuits, where many attention heads rapidly attain high prefix-matching scores and inductive copying capabilities within a narrow training window, coincides with an increase in ICL performance. They hypothesized that this phase change marks the emergence of a wide range of ICL capabilities, driven by the formation of induction circuits.

\begin{figure}[t]
    \centering
    \includegraphics[width=\linewidth]{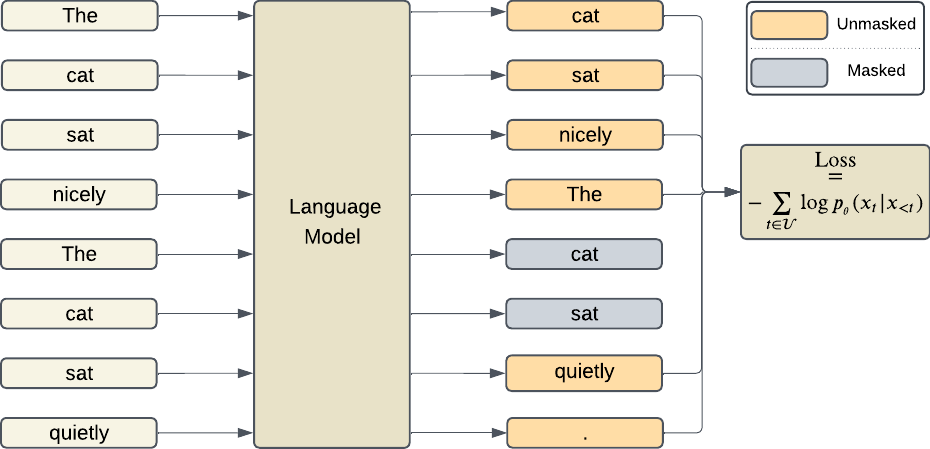}
    \caption{An overview of the \textsc{Hapax} training regime. To suppress inductive copying, we introduce \textsc{Hapax} where positions predictable by induction heads within a context window do not contribute to the loss (gray positions).}
    \label{fig:bigram_loss_masking}
\end{figure}

Subsequent work has found other attention heads that are causally more important for various ICL capabilities \citep{todd2024function, feucht2025the, yin2025attentionheadsmatterincontext}. \citet{yin2025attentionheadsmatterincontext} find that function vector heads are causally more important than induction heads for ICL task performance, but provide correlational evidence that the attention heads with high prefix-matching scores later transform into other ICL-related heads during training. In order to understand the  implications of the phenomena observed with the emergence of induction circuits, a causal intervention on the training process is necessary.

A natural thought experiment to analyze the training dynamics is to ask what would happen to ICL capabilities if every $n$-gram within a Large Language Model's (LLM) context window were a previously unseen hapax legomenon, removing the very repetition that induction heads exploit. 

Our contributions are as follows. (1) We introduce \textsc{Hapax}, a novel training regime which provides a causal methodology for analyzing the emergent phenomena associated with induction circuits. (2) Using \textsc{Hapax}, we provide an alternative view of ICL training dynamics in LLMs: despite the absence of the characteristic phase change and a substantial drop in inductive copying capacity, abstractive ICL capabilities are retained. (3) We show that the token-loss difference metric, previously used to track ICL emergence, primarily reflects gains from inductive copying, and propose computing it over non-extractive token positions to track abstractive capabilities. (4) Together, our results show that the few-shot abstractive ICL capabilities can develop without the canonical induction-circuit phase transition observed under standard training.

\section{Preliminaries}
\label{background}

\begin{figure*}[h]
    \centering
    \includegraphics[width=\linewidth]{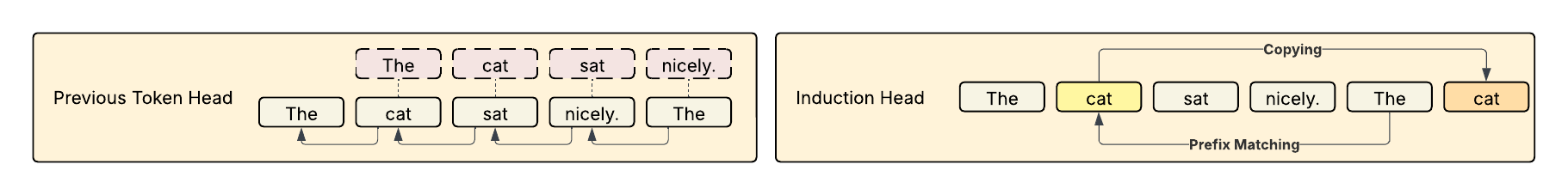}
    \caption{Demonstration of the induction circuit. Previous token heads allow each token to store which token came previously. Induction heads do a match-and-copy operation to reproduce the subsequence that appeared earlier in the context.}
    \label{fig:induction_circuit}
\end{figure*}

Following \citet{todd2024function}, we distinguish \textit{extractive} ICL tasks, where the model must extract the answer from the input, (e.g., ``foil car purple : foil, pen cloud window :'' $\rightarrow$ ``pen''), from \textit{abstractive} tasks, where the model needs to generate new answers rather than copying (e.g., ``Greece : Athens, China : Beijing, Egypt'' $\rightarrow$ ``Cairo''). Abstractive tasks cannot be solved by copying from context, making them a natural testbed for our interest in the emergence of non-copying based ICL tasks.

Induction circuits provide a mechanistic account of copying-based ICL behavior. Induction circuits consist of three steps: (1) \textit{previous token heads} that allow each token to store which token came before it, (2a) \textit{induction heads} that attend to the previous token information in earlier contexts, resulting in a ``prefix-matching'' attention pattern, and (2b) increasing the probability of the attended token in the output, which we refer to as inductive copying (Figure~\ref{fig:induction_circuit}). Formally, given input tokens $(x_1, \dots, x_j)$, induction circuits operate by searching for tokens that hold information of the current token $x_{j}$ (i.e., searching $x_{i+1}$ where $x_{i} = x_{j}$, $i < j$) \citep{elhage2021mathematical}. If there is a matching $x_{i+1}$, the induction head increases its logit for the next prediction. 

To quantify prefix matching, we use the score from \citet{nanda2022transformerlens}. On a sequence $x = (r_1 r_2 \ldots r_s r_1 r_2 \ldots r_{s-1})$ of randomly sampled tokens repeated twice, the prefix-matching score for head $(l, h)$ is
\begin{equation}
\text{PrefixMatching}(l,h) = \frac{1}{s-1} \sum_{i=1}^{s-1} A^{(l,h)}_{s+i,\, i+1}
\end{equation}
where $A^{(l,h)}$ is the attention map of the attention head at layer $l$ and index $h$.
\section{\textsc{Hapax}}

\footnotetext[1]{\hypertarget{fn:sig}{}Results reported in the text include only the subset of tasks with statistically significant differences ($p < 0.05$, McNemar's test) unless stated otherwise.}

\subsection{Training Protocol}
\label{sec:training_pro}

\textsc{Hapax} masks the loss contribution of induction-predictable positions, i.e. positions where the correct token can be retrieved by copying the continuation of an earlier matching bigram, reducing the training pressure for inductive copying. Single-token repetitions are not masked because they cannot be predicted by induction. 

Formally, let $\mathbf{x} = (x_t)_{t \in S}$ be an input sequence of tokens from vocabulary $\mathcal{V}$, where $S = \{1,2, \ldots, T\}$ denotes the set of all token positions and $x_t \in \mathcal{V}$ for all $t \in S$. The probability assigned by the model to the correct prediction at a position $t$ is defined as:
\begin{equation}
p_\theta(x_t | x_{<t}) = \text{softmax}(f_\theta(x_{<t}))_{x_t}
\label{eq:prediction}
\end{equation}
where $f_\theta$ denotes the model with parameters $\theta$. We define the set of masked positions as:
\begin{equation}
\begin{aligned}
\mathcal{M}
= \{\, j \in \mathcal{S} :\;&
\exists\, 1 < i < j \text{ such that} \\
& (x_{i-1}, x_i) = (x_{j-1}, x_j)
\,\}.
\end{aligned}
\label{eq:masked_positions}
\end{equation}
The set of unmasked positions is then $\mathcal{U} = \mathcal{S} \setminus \mathcal{M}$. We compute the \textsc{Hapax} loss, a negative log-likelihood, only over the unmasked positions:
\begin{equation}
\mathcal{L}_{\text{Hapax}} = -\frac{1}{|\mathcal{U}|} \sum_{t \in \mathcal{U}} \log p_\theta(x_t | x_{<t})
\label{eq:loss_computation}
\end{equation}

We train Vanilla and \textsc{Hapax} 1B models from scratch using the transformers library \citep{wolf-etal-2020-transformers} implementation of GPT-NeoX \citep{gpt-neox-library}. We use the same hyperparameter and training configuration as the Pythia models \citep{pmlr-v202-biderman23a}. We use the Pile dataset to train our models \citep{gao2020pile800gbdatasetdiverse}. We train our models for 19900 steps. Using the estimates of \citet{NEURIPS2022_c1e2faff}, a 1B-parameter model reaches the compute-optimal regime after training on approximately 20.2B tokens. Our main 1B Vanilla and \textsc{Hapax} models are trained beyond this threshold. 

Checkpoints are saved every 100 steps, and evaluations are computed from step 100 through step 19900. The training data consists of 41.7B tokens, of which 13.2B (31.7\%) tokens are masked for the \textsc{Hapax} model due to loss masking. Our main results are based on a single random seed for each training condition, following the Pythia training configuration and its default seed (1234). We also train two supplemental runs (seeds 1337 and 2807) under a different precision and GPU ecosystem, whose details can be found in Appendix~\ref{app:additional_seeds}.

Reported confidence intervals are computed using a normal approximation over evaluation examples. Further training variants that we explored but found ineffective are provided in Appendix~\ref{app:training_vars} to inform future work on training dynamics of LLMs.

\subsection{Similarity-Thresholded \textsc{Hapax}}

Certain token pairs can have a high cosine similarity in the input embedding space. As an example, the tokens ``National" and `` National" (with a leading space) have a cosine similarity of 0.84 in the input embedding space of the vanilla model. Although they are not exactly the same token, the model can still acquire a high copying signal through these similar tokens. Referring back to Equation~\ref{eq:masked_positions}, we modify the exact matching with a thresholded matching logic to create a stricter suppression. We mask a position if the corresponding tokens have a cosine similarity higher than $\tau$:
\begin{multline}
\mathcal{M} = \{\, j \in \mathcal{S} : \exists\, i < j \text{ s.t. }
S_{\cos}(e_{x_{i-1}}, e_{x_{j-1}})  \\
> \tau  \text{ and } S_{\cos}(e_{x_i}, e_{x_j}) > \tau \,\}.
\end{multline}
where $e_{x_i}$ refers to the embedding for the token $x_i$. We choose $\tau=0.3$. Consequently, 21.8B (52.5\%) out of 41.7B tokens are masked for the Thresholded-\textsc{Hapax} model. This variant allows us to test whether the copying-related training signals are effectively reduced in \textsc{Hapax}. However, because this criterion masks substantially more token positions than exact-match \textsc{Hapax}, it also risks removing useful non-copying training signal; we therefore treat it as a more aggressive stress test. See Appendix \ref{app:cos_thresh} for details.

\section{Effects of \textsc{Hapax} on ICL}

In the following sections we verify that \textsc{Hapax} suppresses inductive copying but abstractive ICL capabilities emerge despite the substantial reduction in inductive copying. We further analyze loss values throughout training and find that abstractive capabilities emerge even without the characteristic phase shift in loss observed for the vanilla model. Comparisons reported in the text include the tasks with statistically significant differences ($p < 0.05$, McNemar's test; see Appendix \ref{app:choice_of_statistical_test} for our test choice). 

\subsection{Suppression of Inductive Copying}
\label{sec:exact_copy_perf}

\begin{figure}[tbp]
    \centering
    \includegraphics[width=1\linewidth]{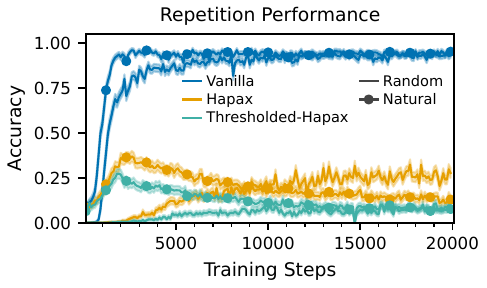}
    
    \caption{Repetition performance in both random token and natural text settings. \textsc{Hapax} models struggle with repetition, which is solvable with induction circuits. This shows the \textsc{Hapax} training can control copying behavior. Accuracy is measured over 1000 randomly generated samples with $s = 25$.}
    \label{fig:random_repetition_combined}
\end{figure}

We evaluate inductive copying on a random repetition task where the model is given 1000 sequences of random repeated tokens $r_1r_2\ldots r_{s}r_1r_2\ldots r_{s - 1}$ and is expected to predict $r_{s}$. This synthetic task does not occur in natural language but is solvable through the induction circuit. Figure~\ref{fig:random_repetition_combined} demonstrates that \textsc{Hapax} causes a major drop in random repetition performance. The \textsc{Hapax} model experiences a 66\% drop and Thresholded-\textsc{Hapax} experiences an 89\% drop in accuracy relative to the vanilla model at the end of training. We also evaluate performance on natural text repetition in Figure~\ref{fig:random_repetition_combined}, where the models are given 1000 sequences of the form $r_1r_2 \ldots r_{s}r_1r_2 \ldots r_{s-1}$ with $r_1r_2 \ldots r_s$ taken from whole sequences in the WikiText dataset \citep{merity2016pointer}. Natural text repetition accuracy decreases over time, suggesting the model's increased incentive to not repeat natural text. The initial increase of natural text repetition is due to the model's increased language modeling capacity, as a non-repeated sequence $r_1r_2\ldots r_{s - 1}$ has the same initial accuracy for the \textsc{Hapax} model (Figure~\ref{fig:natural_text_non_repeating_clean_1b_vs_masked_bigram_loss_1b}) as the maximum attained natural text repetition accuracy. For extractive tasks, \textsc{Hapax} and Thresholded-\textsc{Hapax} show reduced performance compared to vanilla on all but one task with significant differences (Tables \ref{tab:extractive_results}, \ref{tab:extractive_results_thresh}). We also observe this reduction in copying behavior in model generations. See Figure~\ref{fig:sentence_completion_examples} for examples and Appendix~\ref{app:fluency} for fluency evaluations.

Lastly, to test whether the reduction in copying stems from the specific positions we select or simply from the number of tokens masked, we add a Random-Masked baseline that masks the same number of positions per context but chooses them at random, using the same configuration and seed, for the first 5000 steps. Random repetition accuracy at step 5000 is 93.7\% for Vanilla and 93.5\% for the Random-Masked baseline, but 13.3\% for \textsc{Hapax}. The reduction in copying therefore cannot be explained by the number of tokens masked alone, and the positions selected by \textsc{Hapax} have direct influence on the copying signal. 

These results provide evidence that the \textsc{Hapax} training regime can control inductive copying.

\subsection{Abstractive ICL Capabilities}
\label{sec:abstract_copy_perf}

\begin{figure}[tbp]
    \centering
    \includegraphics[width=1\linewidth]{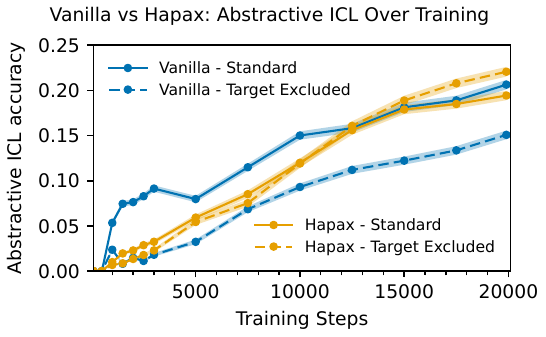}
    
    \caption{The mean absolute gap between the Standard and Target Excluded evaluation is 4.7\% for Vanilla versus 0.8\% for \textsc{Hapax} over training, demonstrating that \textsc{Hapax} is less affected by induction-driven label copying. The target-excluded curves of both models emerge at similar times, indicating that \textsc{Hapax} does not delay the onset of abstractive ICL learning despite the reduction of copying.}
    \label{fig:abstractive_icl_vanilla_vs_hapax_incl_excl_target_ci_step19900}
\end{figure}

\begin{table}[t]
  \centering
  \caption{Abstractive task performance summary at step~19900 over 34 tasks. \textit{\textbf{Sig.}} gives the number of tasks with significant accuracy differences under McNemar tests with joint Benjamini--Hochberg correction (FDR level $0.05$). The \textit{\textbf{$>$ Vanilla}} column gives the number and fraction of those significant tasks where the corresponding variant has higher accuracy than vanilla. \textit{Standard} does not exclude the target label in few-shot demonstrations; \textit{Target excl.} excludes the target label. Individual task names, accuracies and task category explanations are given in Appendix~\ref{app:icl_tasks}.}
  \label{tab:abstractive_task_summary}

  \begin{threeparttable}
  \footnotesize
  \setlength{\tabcolsep}{3pt}
  \renewcommand{\arraystretch}{1.12}

  \begin{tabular*}{\linewidth}{@{\extracolsep{\fill}}lrrrr@{}}
    \toprule
    & \multicolumn{4}{c}{\textbf{Vanilla vs.\ \textsc{Hapax}}} \\
    \cmidrule(lr){2-5}
    \textbf{Task Category}
      & \textbf{Vanilla} & \textbf{\textsc{Hapax}} & \textbf{Sig.} & \textbf{$>$ Vanilla} \\
    \midrule

    \multicolumn{5}{@{}l}{\textit{Algorithmic} ($n=8$)} \\
    \quad Standard & 28.6 $\pm$ 0.9 & 27.3 $\pm$ 0.8 &  5 & 1/5\, (20\%) \\
    \quad Target excl. & 25.7 $\pm$ 0.8 & 31.7 $\pm$ 0.9 &  8 & 7/8\, (88\%) \\
    \addlinespace[4pt]
    \multicolumn{5}{@{}l}{\textit{Linguistic} ($n=4$)} \\
    \quad Standard & 30.3 $\pm$ 2.2 & 39.8 $\pm$ 1.9 &  4 & 4/4\, (100\%) \\
    \quad Target excl. & 30.3 $\pm$ 2.2 & 39.9 $\pm$ 1.9 &  4 & 4/4\, (100\%) \\
    \addlinespace[4pt]
    \multicolumn{5}{@{}l}{\textit{Knowledge} ($n=11$)} \\
    \quad Standard & 18.1 $\pm$ 1.2 & 18.8 $\pm$ 1.2 &  6 & 4/6\, (67\%) \\
    \quad Target excl. & 12.4 $\pm$ 1.1 & 21.4 $\pm$ 1.3 &  9 & 9/9\, (100\%) \\
    \addlinespace[4pt]
    \multicolumn{5}{@{}l}{\textit{Translation} ($n=8$)} \\
    \quad Standard & 4.3 $\pm$ 0.4 & 6.1 $\pm$ 0.5 &  4 & 4/4\, (100\%) \\
    \quad Target excl. & 4.3 $\pm$ 0.4 & 6.1 $\pm$ 0.5 &  4 & 4/4\, (100\%) \\
    \addlinespace[4pt]
    \multicolumn{5}{@{}l}{\textit{Sentence-level} ($n=3$)} \\
    \quad Standard & 39.2 $\pm$ 1.1 & 9.4 $\pm$ 0.8 &  3 & 0/3\, (0\%) \\
    \quad Target excl. & 4.7 $\pm$ 0.4 & 18.1 $\pm$ 1.0 &  3 & 2/3\, (67\%) \\

    \midrule
    \multicolumn{5}{@{}l}{\textbf{All} ($n=34$)} \\
    \quad Standard & 20.6 $\pm$ 0.5 & 19.5 $\pm$ 0.5 & 22 & 13/22\, (59\%) \\
    \quad Target excl. & 15.1 $\pm$ 0.5 & 22.1 $\pm$ 0.5 & 28 & 26/28\, (93\%) \\

    \bottomrule
  \end{tabular*}
  \end{threeparttable}
\end{table}

Abstractive tasks require generating information not contained in the context. We evaluate our models on 34 abstractive tasks, where 26 are from \citet{todd2024function} and 8 are word-level translation tasks from the Open Multilingual Wordnet, accessed via NLTK \citep{bird-loper-2004-nltk} (Appendix~\ref{app:translation_tasks}). We organize tasks into categories following the categorization of \citet{hendel-etal-2023-context} (Table~\ref{tab:icl_category_definitions}).

We evaluate each model under two settings. In the \emph{standard} setting, we use the original 5-shot prompts. Some tasks from \citet{todd2024function} have small label spaces, so the target prediction frequently appears in the few-shot demonstrations as well, and a model can succeed by matching the context's label distribution rather than by abstractive generalization. To control for this, we also introduce a \emph{target-excluded} setting that samples few-shot examples from data points that do not have the same target prediction (see Appendix~\ref{app:low_cardinal_label_space}).

Table~\ref{tab:abstractive_task_summary} summarizes abstractive ICL across 34 tasks. Macro-averaged accuracy is 19.5\% for \textsc{Hapax} and 20.6\% for Vanilla under standard evaluation, and 22.1\% and 15.1\% respectively under target exclusion. At the task level, 22 tasks show a statistically significant difference under standard evaluation (13 favoring \textsc{Hapax}, 9 favoring Vanilla) and 28 under target exclusion (26 favoring \textsc{Hapax}, 2 favoring Vanilla). This indicates that part of Vanilla's standard-evaluation accuracy comes from copying target predictions present in the context, and that abstractive ICL capabilities can emerge under \textsc{Hapax} despite the suppression of inductive copying. Consistent with prior work \citep{todd2024function, feucht2025the, yin2025attentionheadsmatterincontext}, we verify in Appendix~\ref{app:fv_analysis} that function-vector and concept heads still develop under Hapax and  function-vector heads remain causally important for abstractive ICL. 

For Thresholded-\textsc{Hapax}, 24 tasks show a statistically significant difference under standard evaluation (7 favoring Thresholded-\textsc{Hapax}, 17 favoring Vanilla) and 26 under target exclusion (16 favoring Thresholded-\textsc{Hapax}, 10 favoring Vanilla) (Appendix~\ref{app:icl_tasks}). Thresholded-\textsc{Hapax} applies heavier masking and produces lower standard-evaluation scores than \textsc{Hapax} and Vanilla. However, under target exclusion, it achieves higher accuracy than the Vanilla model on 62\% of significant tasks, which suggests that abstractive capabilities can emerge despite a stronger reduction in induction capability. 

In Appendix~\ref{app:2.8b_run} we report an additional 2.8B comparison between Vanilla and \textsc{Hapax} with similar trends.  Overall, these results allow us to examine how abstractive ICL changes when inductive copying is suppressed. Rather than observing broad degradation of ICL, we find that abstractive capabilities are retained, and Figure~\ref{fig:abstractive_icl_vanilla_vs_hapax_incl_excl_target_ci_step19900} shows that Hapax and Vanilla acquire abstractive ICL at similar points in training under target-excluded evaluation.

\section{Loss Dynamics Beyond N-gram Copying}

\begin{figure*}[htbp]
    \centering       \includegraphics[width=\linewidth]{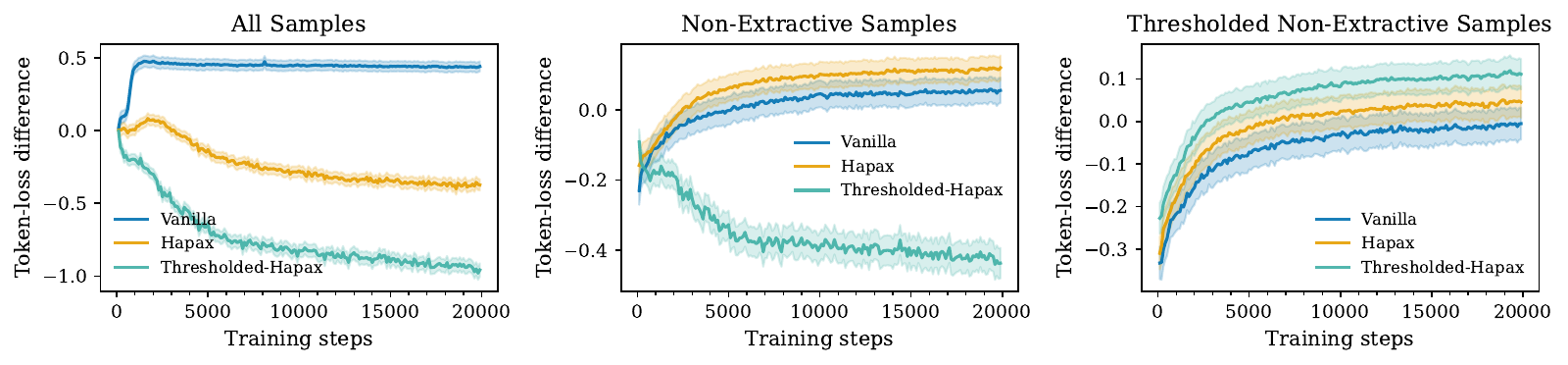}   
    \caption{Comparison of token-loss metrics across training. Unlike the regular (All Samples) TLD, each \textsc{Hapax} variant is able to achieve non-trivial ICL capabilities on its corresponding distribution. Since the regular distribution is dominated by induction-predictable positions, its gains primarily reflect exact copying rather than abstractive ICL. Computing the metric on targeted subsets thus reveals distinct emergence patterns across the distributions of interest. Metrics are computed on randomly sampled validation data.}
    \label{fig:tld_plot}
\end{figure*}

Sections \ref{sec:exact_copy_perf} and \ref{sec:abstract_copy_perf} show that abstractive ICL capabilities can emerge under \textsc{Hapax} despite a substantial reduction in inductive copying. This raises the question of whether the loss-based metrics previously used to track ICL emergence reflect this. 

We use the token-loss difference metric, which is defined by the differences of the cross-entropy loss of arbitrary token positions, conventionally using the 500$^{\text{th}}$ and 50$^{\text{th}}$ token positions. Formally, let $L_t = - \log p_\theta(x_t \mid x_{<t})$ denote the cross-entropy loss at token position $t$. The token-loss difference (TLD) between the 50$^{\text{th}}$ and 500$^{\text{th}}$ tokens is then given by $\Delta TLD = L_{50} - L_{500}$. Intuitively, token-loss difference measures improvement across increasing token positions. If the loss at token 500 is lower and $TLD > 0$, it shows that the model's predictions improved with increasing context.

\begin{figure}[htbp]
    \centering
    \includegraphics[width=1\linewidth]{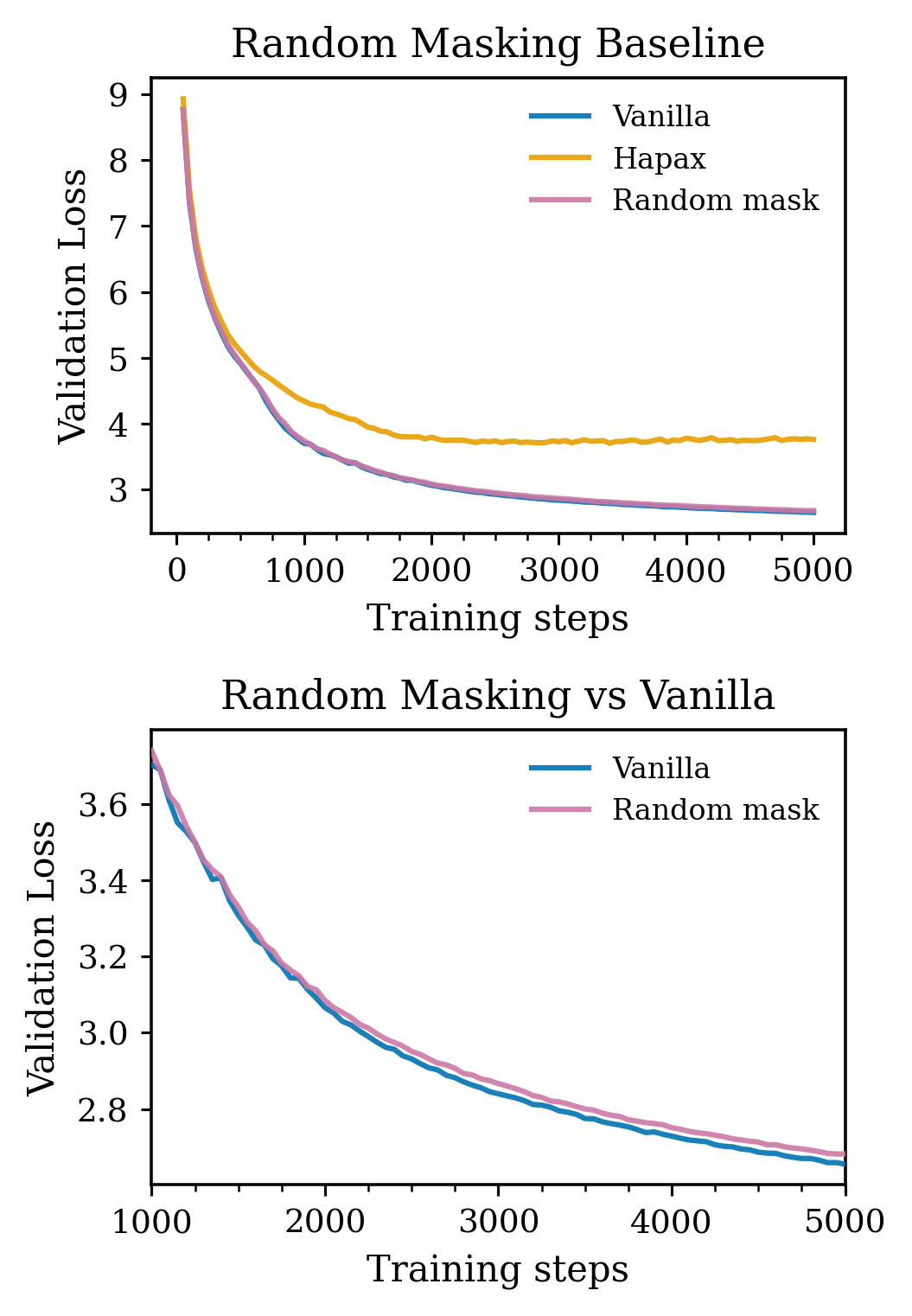}
    
    \caption{Validation loss over the first 5000 steps for the Vanilla, \textsc{Hapax}, and Random-Masked models. Random masking removes the same number of positions per context as \textsc{Hapax} (31.7\% of positions), but selects them at random. At step 5000 the Random-Masked model reaches a validation loss of 2.70, against 2.67 for Vanilla and 3.76 for \textsc{Hapax}. The gap between Vanilla and \textsc{Hapax} therefore cannot be attributed solely to the reduction in supervised positions.}
    \label{fig:val_loss_first5000}
\end{figure}

\citet{yin2025attentionheadsmatterincontext} provided evidence that at the final checkpoint of a model, the metric is strongly influenced by induction heads but does not correlate well with in-context learning task performance. Figure~\ref{fig:tld_plot} (All Samples) shows that the Vanilla model exhibits a sudden improvement in the metric, while \textsc{Hapax} and Thresholded-\textsc{Hapax} appear to worsen over training. As \textsc{Hapax} models perform worse at exact copying tasks but maintain their capabilities on abstractive ICL tasks, our results demonstrate that the sudden increase in the metric for the Vanilla model is indicative of the emergence of inductive copying capabilities (consistent with observations from \citet{lv-etal-2025-language}), but lacks indicative power for the emergence of abstractive ICL capabilities. As \textsc{Hapax} masks 31.7\% of tokens, a substantial fraction of positions in the data are ones where copying-based mechanisms are useful, which explains why the data distribution is dominated by repetition and why this results in a sudden loss drop as soon as induction is learned. Additionaly, in Figure \ref{fig:val_loss_first5000} at step 5000, the Random-Masked model reaches a validation loss of 2.70, against 2.67 for Vanilla and 3.76 for HAPAX. The gap between Vanilla and HAPAX therefore cannot be attributed solely to the reduction in supervised positions.

To investigate this hypothesis further, we propose evaluating TLD on samples where neither the 500\textsuperscript{th} nor the 50\textsuperscript{th} token can be predicted correctly through inductive copying, alongside a Thresholded variant that also excludes positions predictable through highly similar tokens ($\tau=0.3$). 

Figure~\ref{fig:tld_plot} shows that, contrary to the regular TLD metric, the Vanilla model does not exhibit a sudden phase shift on non-extractive samples and is statistically indifferent to \textsc{Hapax}. The individual losses at the non-extractive positions (Figure~\ref{fig:position_losses_combined}) are lower and the TLD metric itself is higher in the \textsc{Hapax} model compared to the Vanilla model, showing that despite the reduction in repetition and inductive copying, the TLD metric improves for non-extractive samples. 

A similar pattern holds for Thresholded-\textsc{Hapax}, where the TLD metric is worse on the first two distributions (since this variant also masks non-extractive but similar tokens) yet shows an increasing trend on the thresholded variant. Overall, our results show that the TLD metric, across all variants, captures the coarse trends of the distribution on which it is computed, so its implications for different types of ICL capabilities depend on the choice of distribution. The regular TLD metric is dominated by induction-predictable positions, and evaluating it on different sample distributions reveals different insights.

\section{Development of Prefix-Matching Under \textsc{Hapax}}

\begin{figure*}[h]
    \centering
    \includegraphics[width=\linewidth]{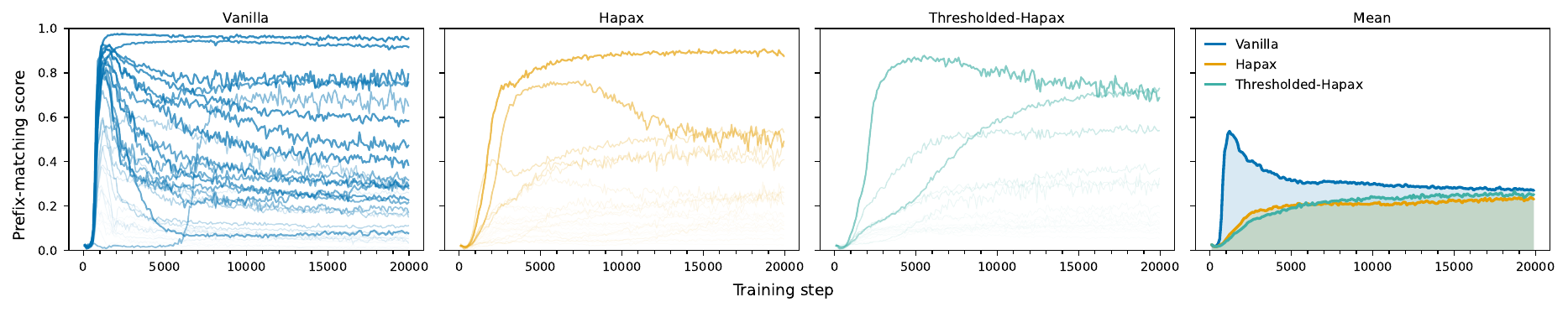}
    \caption{Prefix-matching scores of attention heads across training. Each line is one head. Opacity is proportional to peak score attained throughout training (heads with peak $<0.1$ omitted). Many heads in the Vanilla model peak suddenly and then decay (Mean plot), with 12 of 128 reaching > 80\% prefix-matching score. \textsc{Hapax} variants do not display such a transition and only 1 head in \textsc{Hapax} ever exceeds 80\%.}
    \label{fig:induction_scores_over_time_comparison}
\end{figure*}

\begin{figure}[h]
    \centering
    \includegraphics[width=\linewidth]{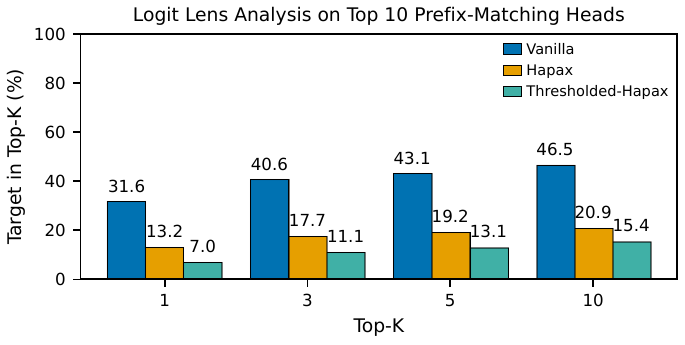}
    \caption{Logit lens on top-10 attention heads with the highest prefix-matching scores on the random repetition task (Section~\ref{sec:exact_copy_perf}). For each head, we measure whether the target token $r_s$ appears among its top-$k$ promoted tokens. Bars show the percentage averaged over the top-10 attention heads with the highest prefix-matching scores. Although these heads attend to $r_s$, \textsc{Hapax} variants increasingly fail to promote it at the output. See Figure~\ref{fig:logit_lens_per_head} for per-head contributions.}
    \label{fig:logit_lens_compare}
\end{figure}

Within Vanilla training, the sudden emergence of induction heads with high prefix-matching scores is a well-documented phenomenon \citep{olsson2022context}. One possibility is that this phase transition is a developmental prerequisite for abstractive ICL, where the sharp emergence of prefix-matching is a necessary transition only after which abstractive capabilities can be learned. Such developmental prerequisites have been documented elsewhere; for example, \citet{chen2024sudden} analyzed the sudden emergence of Syntactic Attention Structure in MLMs, and its necessity for the later acquisition of grammatical capabilities. 

We analyze the trajectory of prefix-matching scores in the Vanilla and  \textsc{Hapax} models. In Figure~\ref{fig:induction_scores_over_time_comparison}, the Vanilla model contains 12 out of 128 attention heads that attain a prefix-matching score over 80\%, (See Section~\ref{background} for formula) spike early in training and then decay, while the \textsc{Hapax} variants display only a single head that attains above 80\% prefix-matching score. This emergence unfolds gradually over 1800 steps in \textsc{Hapax}, in contrast to the sudden emergence in Vanilla. 

Beyond their reduced number and strength, many remaining attention heads with high prefix-matching scores in \textsc{Hapax} also depart from the canonical induction-head behavior. We apply logit lens \citep{nostalgebraist2020logitlens} to the top 10 attention heads with the highest prefix-matching score by taking the corresponding slice of the output matrix $W_o$ for each attention head’s output, then applying the final layer norm and unembedding matrix $W_u$ to approximate the distribution over tokens promoted by that head. 

As shown in Figure~\ref{fig:logit_lens_compare}, inclusion of the token to be copied in the top-$k$ promoted tokens worsens from Vanilla to \textsc{Hapax}, indicating that reduced copying behavior also holds at the attention-head level. This also holds under mean-ablation studies measuring each head’s contribution to predicting the target token (Appendix~\ref{app:ablation_stud}). Among the top 10 attention heads with the highest prefix-matching score in \textsc{Hapax}, 6 suppress rather than promote the correct token (compared to 2 in Vanilla), functionally behaving closer to anti-induction heads \citep{mcdougall-etal-2024-copy,olsson2022context} than to regular induction heads (Appendix~\ref{app:ablation_stud}).

Taken together, these findings suggest that abstractive ICL can emerge without the abrupt prefix-matching phase transition observed during Vanilla training. 

We additionally give supplemental runs with different seeds which differ from the reference in precision (bf16 vs fp16) and GPU ecosystem (ROCm vs CUDA). We observe similar qualitative trends in the emergence of abstractive ICL, prefix-matching scores, and TLD. See Appendix~\ref{app:additional_seeds} for details.

\section{Cross-Checkpoint Patching Reveals Initialization-Level Prefix-Matching Structure}

\begin{figure*}[t]
    \centering

    \begin{subfigure}[t]{0.60\linewidth}
        \centering
        \includegraphics[width=1.1\linewidth]{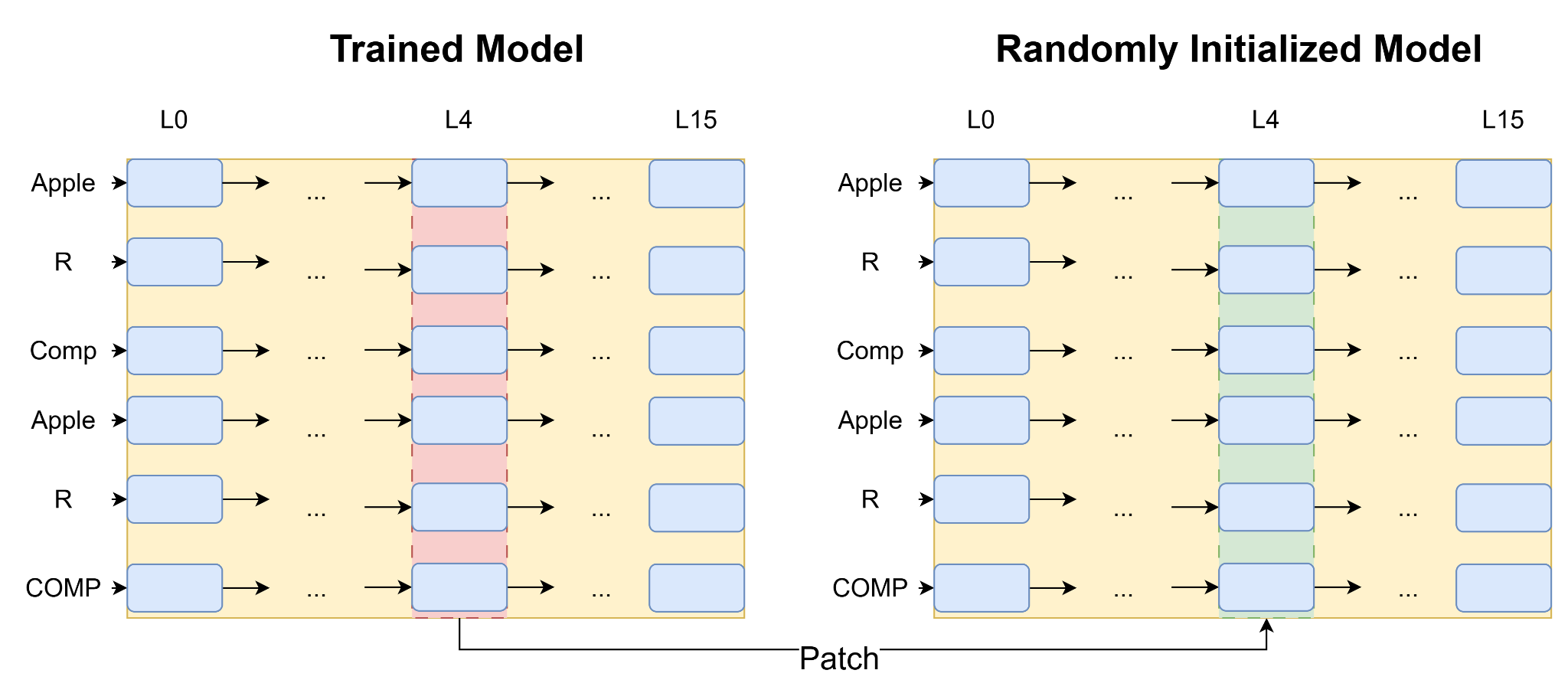}
        \caption{Visualization of the patching setting.}
    \end{subfigure}
    \hfill
    \begin{subfigure}[t]{0.30\linewidth}
        \centering
        \includegraphics[width=0.9\linewidth]{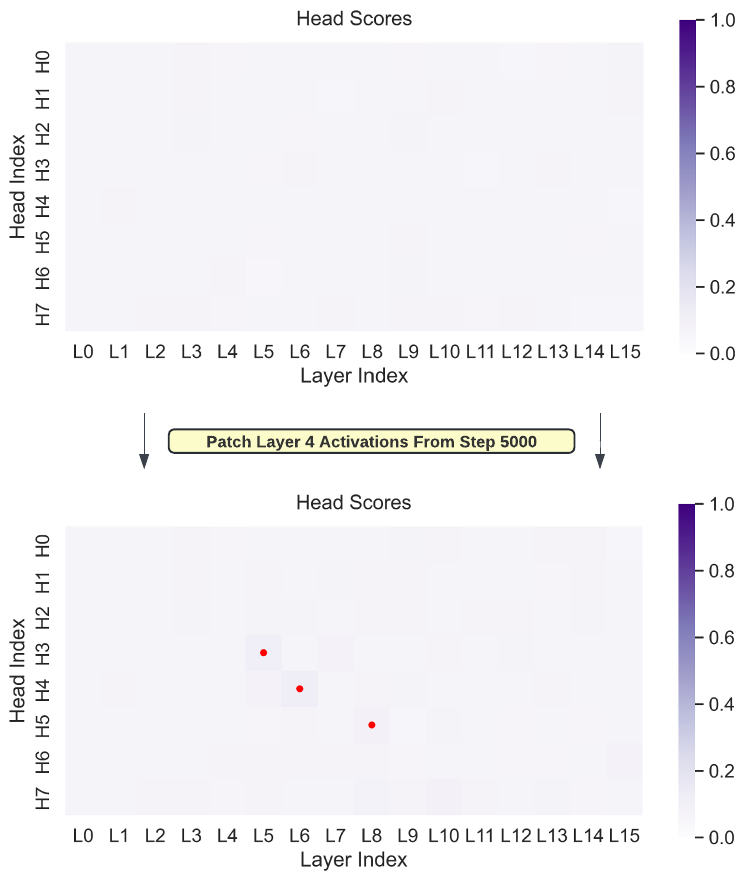}
        \caption{Prefix-Matching scores before and after the patching.}
    \end{subfigure}

    \caption{(a) Demonstration of the patching setup. (b) Induction scores for the randomly initialized Vanilla model before and after patching layer L4 from the step-5000 Vanilla model. At this checkpoint, L4 is the earliest layer with a strong previous-token head (Figure~\ref{fig:prefix-matching_step5000}). Patching L4 increases the maximum prefix-matching score from 0.076 to 0.122. Red dots mark the top three prefix-matching heads after patching, which also rank highly at the final checkpoint.}
    \label{fig:random_init_patching}
\end{figure*}

Under \textsc{Hapax}, the baseline phase transitions do not appear, and prefix-matching scores are attenuated. However, prefix matching is not eliminated entirely. We explore how these patterns can persist when the training objective removes the loss contribution of induction-predictable positions. One possibility involves previous-token heads, which can be learned for reasons beyond in-context copying, such as detokenization  \citep{lad2025remarkablerobustnessllmsstages, gurnee2023findingneuronshaystackcase, feucht-etal-2024-token, kaplan2025tokenswordsinnerlexicon}. If previous-token information is useful for reasons other than in-context copying, then later heads may still exploit this information and produce prefix-matching patterns even without explicit optimization for in-context copying.
To test this, we compare the Vanilla model at step 5000 with the same model at random initialization. At step 5000, layer L4 contains previous-token heads, but no strong prefix-matching heads appear in L4 or earlier layers (Figure~\ref{fig:prefix-matching_step5000}). We patch the layer-L4 output from step 5000 into the randomly initialized model and measure whether later randomly initialized heads begin to exhibit prefix matching.
Figure~\ref{fig:random_init_patching} shows that adding previous-token information increases the maximum prefix-matching score from 0.076 to 0.122\footnote{Highlighted heads are statistically significant under a paired $t$-test comparing each head's prefix-matching score to the mean score of all other heads across the same 1000 samples.}. The highlighted heads also rank among the top five prefix-matching heads at the final Vanilla checkpoint (Figure~\ref{fig:comparison_across_models}), suggesting that a degree of prefix-matching tendency may already be present at initialization. We observe a similar pattern in \textsc{Hapax}: two of its top three final prefix-matching heads also increase their prefix-matching scores in this patching experiment. Since Vanilla and \textsc{Hapax} share the same initialization seed, this overlap suggests that the remaining prefix matching scores can partly reflect initialization-level attention tendency.

\section{Related Work}

\textbf{ICL and Attention Heads}
Despite being trained to do simple next-token prediction, LLMs exhibit the capability to do in-context learning (ICL)~\citep{NEURIPS2020_1457c0d6, JMLR:v24:22-1144}, where they can perform tasks demonstrated in-context ``on-the-fly'' without additional training. Early work identified induction circuits responsible for inductive copying \citep{elhage2021mathematical} and analyzed their influence on ICL \citep{olsson2022context,crosbie-shutova-2025-induction}, hypothesizing they are the basis of in-context learning. Subsequent studies found various attention heads important for ICL, including function vector heads that trigger tasks \citep{todd2024function}, concept induction heads that copy lexical units \citep{feucht2025the}, semantic induction heads extracting relations \citep{ren-etal-2024-identifying}, symbolic induction heads inducing over abstract variables \citep{yang2025emergent}, and n-gram generalizations of induction heads \citep{pmlr-v235-akyurek24a}. Other work has studied how data structure shapes these phenomena, showing that parallel structure in pretraining data can support ICL \citep{chen-etal-2024-parallel}. Our work complements prior studies by examining the training dynamics underlying abstractive ICL when the training pressure for inductive copying is reduced, with the goal of better isolating the role of emergent induction-circuit phenomena.

\textbf{Training Dynamics and Induction Heads.} 

In concurrent work, \citet{wang2026patterningdualinterpretability} introduce patterning, a susceptibility-based framework that derives data interventions to steer training toward a target internal structure. They show in a 3M-parameter language model that targeted re-weighting of tokens can accelerate, delay, or prevent induction formation. They study how to control circuit formation. We instead use \textsc{Hapax} to suppress copying in order to study abstractive ICL's emergence relative to the induction-circuit phase transition at 1B and 2.8B scale.

Theoretic work on synthetic setups has provided insights into induction head development and training dynamics. Several studies analyze phase transitions where \citet{pmlr-v235-singh24c} use clamping to study subcomponents of induction heads and their effect on phase changes. \citet{NEURIPS2022_77c6ccac} shows how skewed Zipfian distributions lead to the emergence of ICL. \citet{bietti2023birth} demonstrates the evolution from global bigram statistics to induction head solutions over training, and \citet{chen2024unveiling, edelman2024the} further analyze convergence to induction-like solutions in Markov chain data with  \citet{edelman2024the} demonstrating the different phases where the model learns more complex strategies with more training.  \citet{minegishi2025beyond} studies a meta-learning setting to investigate mechanisms in non-exact copying scenarios. \citet{singh2025strategy} show that induction heads can be reused across competing strategies over training. In our work, we instead focus on a natural language setting and create a data distribution where induction cannot predict the correct next token to suppress inductive copying.

\textbf{Loss Masking.} Prior work incorporated loss masking strategies to prevent memorization for information of interest \citep{hans2024be, kosireddy-lucas-2025-empirical}. We incorporate a loss masking strategy to create a distribution which allows us to suppress induction behavior throughout training.  

\textbf{Repetition.} Prior work studied repetition from the perspective of data distribution and model outputs. \citet{zucchet2025emergencesparseattentionimpact} analyzes how training data repetition speeds up emergent behavior in language models. \citet{aoyama2026predictingemergenceinductionheads} characterizes statistical conditions under which induction heads emerge. Several works analyze the ``repetition curse'' observed within models \citep{hiraoka-inui-2025-repetition, yao-etal-2025-understanding}. \citet{Welleck2020Neural} presents unlikelihood loss training to mitigate repetitions in the model. As repetition is closely tied to induction heads, \textsc{Hapax} discourages repeated sequences, aiming to suppress inductive copying and, as a result, repetition.

\section{Conclusion}

Our results suggest that the phase transition observed in standard training is closely tied to the learning of repeated, induction-predictable structure in the data. When this copying signal is suppressed through Hapax, abstractive ICL capabilities still develop and are retained, despite the absence of the canonical phase transition. More broadly, our approach illustrates how targeted loss masking can disentangle training dynamics that emerge together under standard training, letting us test whether a phenomenon is tied to the capability being studied or to other predictable structure in the data.

\section*{Limitations}

\textsc{Hapax} tests whether the abstractive ICL capabilities studied here can emerge when copying-related training signal is reduced. The results characterize this specific question rather than final behavior of the model across all model sizes, training budgets, and masking variants. Our main analysis is limited at the 1B setting with the official Pythia configuration, covering Vanilla, \textsc{Hapax}, and Thresholded-\textsc{Hapax} across training, with the 2.8B run as a scale check. We also report two seeds in Appendix~\ref{app:additional_seeds}, though they differ from the main run in precision and GPU ecosystem. Scaling across larger models, longer training, more seeds, and alternative masking variants is an important future direction.

Models reach non-trivial few-shot accuracy on several tasks and remain low on others (e.g., Polish-English) (Table~\ref{tab:abstractive_results}). The comparisons measure whether HAPAX retains performance comparable to Vanilla, rather than absolute capability on every task.

Results are scoped to the few-shot abstractive settings used here. These settings test whether certain forms of abstractive ICL can emerge despite the absence of the canonical induction-circuit phase transition and reduced copying behavior. Other forms of abstractive ICL which require larger models or more complex reasoning may exhibit different dynamics. Similarly, our experiments show that function-vector and concept heads emerge and remain causally important under \textsc{Hapax}. Additional mechanisms may also be at play, which have not been analyzed within the scope of this work.

The Random-Masked baseline that separates the effect of masking induction predictable positions from that of reducing supervision is trained at the 1B setting for 5000 steps. This covers the window in which the emergence of inductive copying is observable, but the comparison is not carried through to the end of training due to computational limits.

\section*{Ethics Considerations}
This paper aims to advance the foundational understanding of in-context learning mechanisms. While such research may influence future model development and deployment, we cannot meaningfully anticipate these downstream impacts within the scope of this work.

\section*{Acknowledgments}
We thank Andy Arditi, Amir Zur, Arnab Sen Sharma, Constanza Fierro, and Nikhil Prakash for valuable feedback and discussions.  AB and DB are supported by the NSF National Deep Inference Fabric Grant \#2408455, and CW and DB are supported by NSF Research Grant \#2403304. SF and DB and KŞ are funded by a grant from Coefficient Giving. KŞ is supported by the Fulbright Program, which is sponsored by the U.S. Department of State and Turkish Fulbright Commission. The contents of this work are solely the responsibility of the authors.

The authors acknowledge the MGHPCC AI Computing Resource (AICR), with support from the Commonwealth of Massachusetts and the AICR participating institutions, for providing AI and high-performance computing resources used in the results reported in this publication. The authors thank the Northeastern University Research Computing team for their assistance with using the AICR cluster.

This research used resources of the Oak Ridge Leadership Computing Facility at the Oak Ridge National Laboratory, which is supported by the Office of Science of the U.S. Department of Energy under Contract No. DE-AC05-00OR22725.

\bibliography{custom}

\appendix

\setcounter{figure}{0}
\setcounter{table}{0}
\renewcommand{\thefigure}{\Alph{section}\arabic{figure}}
\renewcommand{\thetable}{\Alph{section}\arabic{table}}

\section{Additional Model Results}
\label{app:additional_mod}

\subsection{ICL Task Results}\label{app:icl_tasks}
Within this section, we provide the full table summaries of the tasks.
\begin{table*}[t]
  \centering
    \caption{Abstractive task performance summary for the 1B model at step~19900 over 34 tasks. Macro-averaged accuracies (\%) within each family are shown with 95\% confidence intervals (normal approximation on the macro mean per category). \textit{\textbf{Sig.}} gives the number of tasks with significant accuracy differences under McNemar tests with joint Benjamini--Hochberg correction ($\alpha=0.05$). The \textit{\textbf{$>$ Vanilla}} column gives the number and fraction of those significant tasks where the corresponding variant has higher accuracy than vanilla. \textit{Standard} includes the target label in demonstrations; \textit{Target excl.} withholds the target label.}
  \label{tab:abstractive_task_summary_with_thresh}

  \begin{threeparttable}
  \footnotesize
  \setlength{\tabcolsep}{3pt}
  \renewcommand{\arraystretch}{1.12}

  \begin{tabular*}{\linewidth}{@{\extracolsep{\fill}}lrrrrrrrr@{}}
    \toprule
    & \multicolumn{4}{c}{\textbf{\textsc{Hapax}}}
    & \multicolumn{4}{c}{\textbf{Thresh-\textsc{Hapax}}} \\
    \cmidrule(lr){2-5}\cmidrule(l){6-9}
    \textbf{Task Category}
      & \textbf{Van.} & \textbf{\textsc{Hapax}} & \textbf{Sig.} & \textbf{$>$ Van.}
      & \textbf{Van.} & \textbf{Thresh} & \textbf{Sig.} & \textbf{$>$ Van.} \\
    \midrule

    \multicolumn{9}{@{}l}{\textit{Algorithmic} ($n=8$)} \\
    \quad Standard & 28.6 $\pm$ 0.9 & 27.3 $\pm$ 0.8 &  5 & 1/5\, (20\%) & 28.6 $\pm$ 0.9 & 15.3 $\pm$ 0.8 &  7 & 0/7\, (0\%) \\
    \quad Target excl. & 25.7 $\pm$ 0.8 & 31.7 $\pm$ 0.9 &  8 & 7/8\, (88\%) & 25.7 $\pm$ 0.8 & 19.1 $\pm$ 0.8 &  7 & 3/7\, (43\%) \\
    \addlinespace[4pt]
    \multicolumn{9}{@{}l}{\textit{Linguistic} ($n=4$)} \\
    \quad Standard & 30.3 $\pm$ 2.2 & 39.8 $\pm$ 1.9 &  4 & 4/4\, (100\%) & 30.3 $\pm$ 2.2 & 18.3 $\pm$ 2.0 &  4 & 2/4\, (50\%) \\
    \quad Target excl. & 30.3 $\pm$ 2.2 & 39.9 $\pm$ 1.9 &  4 & 4/4\, (100\%) & 30.3 $\pm$ 2.2 & 18.3 $\pm$ 2.0 &  4 & 2/4\, (50\%) \\
    \addlinespace[4pt]
    \multicolumn{9}{@{}l}{\textit{Knowledge} ($n=11$)} \\
    \quad Standard & 18.1 $\pm$ 1.2 & 18.8 $\pm$ 1.2 &  6 & 4/6\, (67\%) & 18.1 $\pm$ 1.2 & 10.0 $\pm$ 0.9 &  5 & 0/5\, (0\%) \\
    \quad Target excl. & 12.4 $\pm$ 1.1 & 21.4 $\pm$ 1.3 &  9 & 9/9\, (100\%) & 12.4 $\pm$ 1.1 & 12.9 $\pm$ 1.0 &  7 & 5/7\, (71\%) \\
    \addlinespace[4pt]
    \multicolumn{9}{@{}l}{\textit{Translation} ($n=8$)} \\
    \quad Standard & 4.3 $\pm$ 0.4 & 6.1 $\pm$ 0.5 &  4 & 4/4\, (100\%) & 4.3 $\pm$ 0.4 & 7.9 $\pm$ 0.6 &  5 & 5/5\, (100\%) \\
    \quad Target excl. & 4.3 $\pm$ 0.4 & 6.1 $\pm$ 0.5 &  4 & 4/4\, (100\%) & 4.3 $\pm$ 0.4 & 7.9 $\pm$ 0.6 &  5 & 5/5\, (100\%) \\
    \addlinespace[4pt]
    \multicolumn{9}{@{}l}{\textit{Sentence-level} ($n=3$)} \\
    \quad Standard & 39.2 $\pm$ 1.1 & 9.4 $\pm$ 0.8 &  3 & 0/3\, (0\%) & 39.2 $\pm$ 1.1 & 0.4 $\pm$ 0.1 &  3 & 0/3\, (0\%) \\
    \quad Target excl. & 4.7 $\pm$ 0.4 & 18.1 $\pm$ 1.0 &  3 & 2/3\, (67\%) & 4.7 $\pm$ 0.4 & 1.1 $\pm$ 0.2 &  3 & 1/3\, (33\%) \\

    \midrule
    \multicolumn{9}{@{}l}{\textbf{All} ($n=34$)} \\
    \quad Standard & 20.6 $\pm$ 0.5 & 19.5 $\pm$ 0.5 & 22 & 13/22\, (59\%) & 20.6 $\pm$ 0.5 & 10.9 $\pm$ 0.4 & 24 & 7/24\, (29\%) \\
    \quad Target excl. & 15.1 $\pm$ 0.5 & 22.1 $\pm$ 0.5 & 28 & 26/28\, (93\%) & 15.1 $\pm$ 0.5 & 12.8 $\pm$ 0.5 & 26 & 16/26\, (62\%) \\

    \bottomrule
  \end{tabular*}
  \end{threeparttable}
\end{table*}

\begin{table*}[htbp]
\centering
\caption{Performance comparison on \textbf{abstractive} in-context learning tasks (5-Shot) for Vanilla and \textsc{Hapax}. 
Values show accuracy $\pm$ 95\% CI margin (binomial normal-approximation per task). Bold indicates higher performance; $\dagger$ denotes statistical significance ($p < 0.05$) according to McNemar's test.}
\label{tab:abstractive_results}
\scriptsize
\begin{tabular}{lcc@{\hspace{1.8em}}lcc}
\toprule
\textbf{Task} & \textbf{Vanilla (\%)} & \textbf{\textsc{Hapax} (\%)} &
\textbf{Task} & \textbf{Vanilla (\%)} & \textbf{\textsc{Hapax} (\%)} \\
\midrule
AG News & \textbf{32.7 $\pm$ 1.3}$^\dagger$ & 5.6 $\pm$ 0.6 & Antonym & 0.7 $\pm$ 0.3 & \textbf{2.2 $\pm$ 0.6}$^\dagger$ \\
Capitalize (Full Word) & \textbf{78.2 $\pm$ 2.8}$^\dagger$ & 59.0 $\pm$ 3.4 & Capitalize First Letter & 37.1 $\pm$ 3.3 & \textbf{66.8 $\pm$ 3.2}$^\dagger$ \\
Capitalize Last Letter & \textbf{9.3 $\pm$ 2.0}$^\dagger$ & 3.8 $\pm$ 1.3 & Capitalize Second Letter & \textbf{11.9 $\pm$ 2.3}$^\dagger$ & 2.4 $\pm$ 1.1 \\
Chinese\textrightarrow{English} & 4.1 $\pm$ 1.2 & \textbf{6.8 $\pm$ 1.6}$^\dagger$ & CommonsenseQA & \textbf{20.4 $\pm$ 1.1}$^\dagger$ & 5.2 $\pm$ 0.6 \\
Country-Capital & 29.6 $\pm$ 6.5 & \textbf{42.3 $\pm$ 7.0}$^\dagger$ & Country-Currency & 5.4 $\pm$ 4.6 & 5.4 $\pm$ 4.6 \\
French\textrightarrow{English} & 11.0 $\pm$ 2.0 & \textbf{11.9 $\pm$ 2.0} & Indonesian\textrightarrow{English} & \textbf{1.0 $\pm$ 0.6} & 0.6 $\pm$ 0.5 \\
Italian\textrightarrow{English} & 4.8 $\pm$ 1.3 & \textbf{5.1 $\pm$ 1.4} & Japanese\textrightarrow{English} & 0.4 $\pm$ 0.4 & \textbf{2.3 $\pm$ 0.9}$^\dagger$ \\
Landmark-Country & 32.8 $\pm$ 3.2 & \textbf{36.6 $\pm$ 3.3}$^\dagger$ & Lowercase First Letter & \textbf{71.1 $\pm$ 3.1}$^\dagger$ & 66.5 $\pm$ 3.2 \\
Lowercase Last Letter & 7.0 $\pm$ 1.8 & \textbf{7.2 $\pm$ 1.8} & National Parks & 18.2 $\pm$ 3.6 & \textbf{20.8 $\pm$ 3.7} \\
Next Capital Letter & \textbf{6.5 $\pm$ 1.7} & 6.3 $\pm$ 1.7 & Next Item & 12.4 $\pm$ 4.3 & \textbf{26.7 $\pm$ 5.8}$^\dagger$ \\
Park-Country & 12.8 $\pm$ 2.4 & \textbf{16.8 $\pm$ 2.7}$^\dagger$ & Person-Instrument & \textbf{23.9 $\pm$ 3.7}$^\dagger$ & 1.4 $\pm$ 1.0 \\
Person-Occupation & \textbf{18.5 $\pm$ 2.7}$^\dagger$ & 4.4 $\pm$ 1.4 & Person-Sport & 23.3 $\pm$ 4.6 & \textbf{28.6 $\pm$ 5.0} \\
Polish\textrightarrow{English} & 1.7 $\pm$ 0.8 & \textbf{2.2 $\pm$ 0.9} & Portuguese\textrightarrow{English} & 4.7 $\pm$ 1.3 & \textbf{8.1 $\pm$ 1.7}$^\dagger$ \\
Present-Past & 55.3 $\pm$ 5.7 & \textbf{78.2 $\pm$ 4.7}$^\dagger$ & Previous Item & 3.1 $\pm$ 2.3 & \textbf{4.4 $\pm$ 2.7} \\
Product-Company & 18.6 $\pm$ 3.3 & \textbf{19.5 $\pm$ 3.4} & Sentiment & \textbf{64.6 $\pm$ 2.7}$^\dagger$ & 17.4 $\pm$ 2.2 \\
Singular-Plural & 63.9 $\pm$ 6.6 & \textbf{76.6 $\pm$ 5.8}$^\dagger$ & Spanish\textrightarrow{English} & 7.0 $\pm$ 1.6 & \textbf{11.7 $\pm$ 2.0}$^\dagger$ \\
Synonym & 1.2 $\pm$ 0.4 & \textbf{2.4 $\pm$ 0.6}$^\dagger$ & Word Length & \textbf{7.7 $\pm$ 1.8} & 6.6 $\pm$ 1.7 \\

\bottomrule
\end{tabular}
\vspace{-0.5em}
\end{table*}

\begin{table*}[htbp]
\centering
\caption{Performance comparison on \textbf{extractive} in-context learning tasks (5-Shot) for Vanilla and \textsc{Hapax} models. Values show accuracy $\pm$ 95\% CI margin. Bold indicates higher performance; $\dagger$ denotes statistical significance ($p < 0.05$) according to McNemar's test.}
\label{tab:extractive_results}
\scriptsize
\begin{tabular}{lcc@{\hspace{1.8em}}lcc}
\toprule
\textbf{Task} & \textbf{Vanilla (\%)} & \textbf{\textsc{Hapax} (\%)} &
\textbf{Task} & \textbf{Vanilla (\%)} & \textbf{\textsc{Hapax} (\%)} \\
\midrule
Adjective V Verb 3 & \textbf{58.8 $\pm$ 3.1}$^\dagger$ & 41.7 $\pm$ 3.1 &
Adjective V Verb 5 & \textbf{52.7 $\pm$ 3.1}$^\dagger$ & 31.2 $\pm$ 2.9 \\

Alphabetically Last 5 & \textbf{18.6 $\pm$ 2.4} & 15.5 $\pm$ 2.2 &
Animal V Object 3 & \textbf{49.4 $\pm$ 3.1}$^\dagger$ & 26.0 $\pm$ 2.7 \\

Animal V Object 5 & \textbf{37.1 $\pm$ 3.0}$^\dagger$ & 17.5 $\pm$ 2.4 &
Choose First Of 3 & \textbf{94.9 $\pm$ 1.4}$^\dagger$ & 19.1 $\pm$ 2.4 \\

Choose First Of 5 & \textbf{95.6 $\pm$ 1.3}$^\dagger$ & 17.1 $\pm$ 2.3 &
Alphabetically First 3 & \textbf{35.5 $\pm$ 3.0}$^\dagger$ & 26.8 $\pm$ 2.7 \\

Alphabetically First 5 & \textbf{19.9 $\pm$ 2.5}$^\dagger$ & 13.8 $\pm$ 2.1 &
Alphabetically Last 3 & \textbf{32.6 $\pm$ 2.9}$^\dagger$ & 24.4 $\pm$ 2.7 \\

Choose Middle Of 5 & \textbf{25.8 $\pm$ 2.7}$^\dagger$ & 13.4 $\pm$ 2.1 &
Color V Animal 3 & \textbf{63.4 $\pm$ 3.0} & 60.2 $\pm$ 3.0 \\

Color V Animal 5 & 48.4 $\pm$ 3.1 & \textbf{49.2 $\pm$ 3.1} &
Choose Last Of 3 & \textbf{65.6 $\pm$ 2.9}$^\dagger$ & 46.5 $\pm$ 3.1 \\

Choose Last Of 5 & \textbf{51.5 $\pm$ 3.1}$^\dagger$ & 44.0 $\pm$ 3.1 &
Choose Middle Of 3 & \textbf{39.7 $\pm$ 3.0}$^\dagger$ & 26.9 $\pm$ 2.7 \\

Concept V Object 3 & \textbf{56.8 $\pm$ 3.1} & 54.2 $\pm$ 3.1 &
Conll2003 Person & \textbf{72.0 $\pm$ 3.0}$^\dagger$ & 19.9 $\pm$ 2.7 \\

Fruit V Animal 3 & \textbf{43.7 $\pm$ 3.1}$^\dagger$ & 17.7 $\pm$ 2.4 &
Concept V Object 5 & 45.3 $\pm$ 3.1 & \textbf{52.2 $\pm$ 3.1}$^\dagger$ \\

Conll2003 Location & \textbf{58.5 $\pm$ 3.1}$^\dagger$ & 34.7 $\pm$ 3.0 &
Object V Concept 5 & \textbf{68.9 $\pm$ 2.9}$^\dagger$ & 36.3 $\pm$ 3.0 \\

Conll2003 Organization & \textbf{54.7 $\pm$ 3.3}$^\dagger$ & 17.8 $\pm$ 2.5 &
Fruit V Animal 5 & \textbf{28.0 $\pm$ 2.8}$^\dagger$ & 8.8 $\pm$ 1.8 \\

Object V Concept 3 & \textbf{73.5 $\pm$ 2.7}$^\dagger$ & 43.0 $\pm$ 3.1 &
Squad Val & \textbf{14.5 $\pm$ 2.3}$^\dagger$ & 5.4 $\pm$ 1.5 \\

Verb V Adjective 3 & \textbf{70.1 $\pm$ 2.8}$^\dagger$ & 44.9 $\pm$ 3.1 &
Verb V Adjective 5 & \textbf{61.0 $\pm$ 3.0}$^\dagger$ & 36.3 $\pm$ 3.0 \\

\bottomrule
\end{tabular}
\vspace{-0.5em}
\end{table*}

\begin{table*}[htbp]
\centering
\caption{Performance comparison on \textbf{abstractive} in-context learning tasks (5-Shot) for Vanilla and Thresholded-\textsc{Hapax}. 
Values show accuracy $\pm$ 95\% CI margin. Bold indicates higher performance; $\dagger$ denotes statistical significance ($p < 0.05$) according to McNemar's test.}
\label{tab:abstractive_results_thresh}
\scriptsize
\begin{tabular}{lcc@{\hspace{1.8em}}lcc}
\toprule
\textbf{Task} & \textbf{Vanilla (\%)} & \textbf{\textsc{Hapax Thresh} (\%)} &
\textbf{Task} & \textbf{Vanilla (\%)} & \textbf{\textsc{Hapax Thresh} (\%)} \\
\midrule
AG News & \textbf{32.7 $\pm$ 1.3}$^\dagger$ & 1.1 $\pm$ 0.3 & Antonym & 0.7 $\pm$ 0.3 & \textbf{7.3 $\pm$ 1.0}$^\dagger$ \\
Capitalize (Full Word) & \textbf{78.2 $\pm$ 2.8}$^\dagger$ & 35.9 $\pm$ 3.3 & Capitalize First Letter & \textbf{37.1 $\pm$ 3.3}$^\dagger$ & 24.0 $\pm$ 2.9 \\
Capitalize Last Letter & \textbf{9.3 $\pm$ 2.0}$^\dagger$ & 4.1 $\pm$ 1.4 & Capitalize Second Letter & \textbf{11.9 $\pm$ 2.3}$^\dagger$ & 3.7 $\pm$ 1.3 \\
Chinese\textrightarrow{English} & 4.1 $\pm$ 1.2 & \textbf{10.1 $\pm$ 1.9}$^\dagger$ & CommonsenseQA & \textbf{20.4 $\pm$ 1.1}$^\dagger$ & 0.0 $\pm$ 0.0 \\
Country-Capital & \textbf{29.6 $\pm$ 6.5} & 29.1 $\pm$ 6.5 & Country-Currency & \textbf{5.4 $\pm$ 4.6} & 2.2 $\pm$ 2.5 \\
French\textrightarrow{English} & 11.0 $\pm$ 2.0 & \textbf{14.5 $\pm$ 2.2}$^\dagger$ & Indonesian\textrightarrow{English} & 1.0 $\pm$ 0.6 & 1.0 $\pm$ 0.6 \\
Italian\textrightarrow{English} & 4.8 $\pm$ 1.3 & \textbf{6.6 $\pm$ 1.6} & Japanese\textrightarrow{English} & 0.4 $\pm$ 0.4 & \textbf{3.8 $\pm$ 1.2}$^\dagger$ \\
Landmark-Country & \textbf{32.8 $\pm$ 3.2}$^\dagger$ & 21.8 $\pm$ 2.8 & Lowercase First Letter & \textbf{71.1 $\pm$ 3.1}$^\dagger$ & 46.6 $\pm$ 3.4 \\
Lowercase Last Letter & \textbf{7.0 $\pm$ 1.8}$^\dagger$ & 3.3 $\pm$ 1.2 & National Parks & \textbf{18.2 $\pm$ 3.6} & 16.9 $\pm$ 3.5 \\
Next Capital Letter & \textbf{6.5 $\pm$ 1.7} & 4.2 $\pm$ 1.4 & Next Item & \textbf{12.4 $\pm$ 4.3} & 8.9 $\pm$ 3.7 \\
Park-Country & 12.8 $\pm$ 2.4 & \textbf{15.0 $\pm$ 2.6} & Person-Instrument & \textbf{23.9 $\pm$ 3.7}$^\dagger$ & 0.8 $\pm$ 0.8 \\
Person-Occupation & \textbf{18.5 $\pm$ 2.7}$^\dagger$ & 4.8 $\pm$ 1.5 & Person-Sport & \textbf{23.3 $\pm$ 4.6}$^\dagger$ & 0.0 $\pm$ 0.5 \\
Polish\textrightarrow{English} & 1.7 $\pm$ 0.8 & \textbf{3.0 $\pm$ 1.1} & Portuguese\textrightarrow{English} & 4.7 $\pm$ 1.3 & \textbf{9.9 $\pm$ 1.9}$^\dagger$ \\
Present-Past & \textbf{55.3 $\pm$ 5.7}$^\dagger$ & 38.9 $\pm$ 5.6 & Previous Item & \textbf{3.1 $\pm$ 2.3} & 1.3 $\pm$ 1.4 \\
Product-Company & \textbf{18.6 $\pm$ 3.3}$^\dagger$ & 8.8 $\pm$ 2.4 & Sentiment & \textbf{64.6 $\pm$ 2.7}$^\dagger$ & 0.0 $\pm$ 0.1 \\
Singular-Plural & \textbf{63.9 $\pm$ 6.6}$^\dagger$ & 23.4 $\pm$ 5.8 & Spanish\textrightarrow{English} & 7.0 $\pm$ 1.6 & \textbf{14.2 $\pm$ 2.2}$^\dagger$ \\
Synonym & 1.2 $\pm$ 0.4 & \textbf{3.7 $\pm$ 0.7}$^\dagger$ & Word Length & \textbf{7.7 $\pm$ 1.8}$^\dagger$ & 0.9 $\pm$ 0.6 \\

\bottomrule
\end{tabular}
\vspace{-0.5em}
\end{table*}

\begin{table*}[htbp]
\centering
\caption{Performance comparison on \textbf{extractive} in-context learning tasks (5-Shot) for Vanilla and Thresholded-\textsc{Hapax}. Values show accuracy $\pm$ 95\% CI margin. Bold indicates higher performance; $\dagger$ denotes statistical significance ($p < 0.05$) according to McNemar's test.}
\label{tab:extractive_results_thresh}
\scriptsize
\begin{tabular}{lcc@{\hspace{1.8em}}lcc}
\toprule
\textbf{Task} & \textbf{Vanilla (\%)} & \textbf{\textsc{Hapax Thresh} (\%)} &
\textbf{Task} & \textbf{Vanilla (\%)} & \textbf{\textsc{Hapax Thresh} (\%)} \\
\midrule
Adjective V Verb 3 & \textbf{58.8 $\pm$ 3.1} & 56.1 $\pm$ 3.1 &
Adjective V Verb 5 & \textbf{52.7 $\pm$ 3.1}$^\dagger$ & 39.4 $\pm$ 3.0 \\
Alphabetically Last 5 & \textbf{18.6 $\pm$ 2.4}$^\dagger$ & 13.0 $\pm$ 2.1 &
Animal V Object 3 & \textbf{49.4 $\pm$ 3.1}$^\dagger$ & 27.3 $\pm$ 2.8 \\
Animal V Object 5 & \textbf{37.1 $\pm$ 3.0}$^\dagger$ & 16.3 $\pm$ 2.3 &
Choose First Of 3 & \textbf{94.9 $\pm$ 1.4}$^\dagger$ & 30.7 $\pm$ 2.9 \\
Choose First Of 5 & \textbf{95.6 $\pm$ 1.3}$^\dagger$ & 27.0 $\pm$ 2.8 &
Alphabetically First 3 & \textbf{35.5 $\pm$ 3.0}$^\dagger$ & 28.1 $\pm$ 2.8 \\
Alphabetically First 5 & \textbf{19.9 $\pm$ 2.5}$^\dagger$ & 14.5 $\pm$ 2.2 &
Alphabetically Last 3 & \textbf{32.6 $\pm$ 2.9}$^\dagger$ & 24.5 $\pm$ 2.7 \\
Choose Middle Of 5 & \textbf{25.8 $\pm$ 2.7}$^\dagger$ & 7.4 $\pm$ 1.6 &
Color V Animal 3 & \textbf{63.4 $\pm$ 3.0}$^\dagger$ & 18.6 $\pm$ 2.4 \\
Color V Animal 5 & \textbf{48.4 $\pm$ 3.1}$^\dagger$ & 9.5 $\pm$ 1.8 &
Choose Last Of 3 & \textbf{65.6 $\pm$ 2.9}$^\dagger$ & 53.6 $\pm$ 3.1 \\
Choose Last Of 5 & 51.5 $\pm$ 3.1 & \textbf{58.7 $\pm$ 3.1}$^\dagger$ &
Choose Middle Of 3 & \textbf{39.7 $\pm$ 3.0}$^\dagger$ & 21.5 $\pm$ 2.5 \\
Concept V Object 3 & \textbf{56.8 $\pm$ 3.1}$^\dagger$ & 35.9 $\pm$ 3.0 &
Conll2003 Person & \textbf{72.0 $\pm$ 3.0}$^\dagger$ & 25.8 $\pm$ 1.4 \\
Fruit V Animal 3 & \textbf{43.7 $\pm$ 3.1}$^\dagger$ & 21.5 $\pm$ 2.5 &
Concept V Object 5 & \textbf{45.3 $\pm$ 3.1}$^\dagger$ & 18.3 $\pm$ 2.4 \\
Conll2003 Location & \textbf{58.5 $\pm$ 3.1}$^\dagger$ & 17.5 $\pm$ 1.1 &
Object V Concept 5 & \textbf{68.9 $\pm$ 2.9}$^\dagger$ & 51.9 $\pm$ 3.1 \\
Conll2003 Organization & \textbf{54.7 $\pm$ 3.3}$^\dagger$ & 21.2 $\pm$ 1.3 &
Fruit V Animal 5 & \textbf{28.0 $\pm$ 2.8}$^\dagger$ & 9.9 $\pm$ 1.9 \\
Object V Concept 3 & \textbf{73.5 $\pm$ 2.7}$^\dagger$ & 58.4 $\pm$ 3.1 &
Squad Val & \textbf{14.5 $\pm$ 2.3}$^\dagger$ & 8.5 $\pm$ 0.8 \\
Verb V Adjective 3 & \textbf{70.1 $\pm$ 2.8}$^\dagger$ & 33.6 $\pm$ 2.9 &
Verb V Adjective 5 & \textbf{61.0 $\pm$ 3.0}$^\dagger$ & 20.8 $\pm$ 2.5 \\
\bottomrule
\end{tabular}
\end{table*}

\begin{table*}[htbp]
\centering
\caption{Performance comparison on \textbf{abstractive} in-context learning tasks (5-Shot) for Vanilla and \textsc{Hapax} by ensuring that the target answer for each test instance is not included in the 5-shot examples. 
Values show accuracy $\pm$ 95\% CI margin. Bold indicates higher performance; $\dagger$ denotes statistical significance ($p < 0.05$) according to McNemar's test.}
\label{tab:abstractive_results_nonextract}
\scriptsize
\begin{tabular}{lcc@{\hspace{1.8em}}lcc}
\toprule
\textbf{Task} & \textbf{Vanilla (\%)} & \textbf{\textsc{Hapax} (\%)} &
\textbf{Task} & \textbf{Vanilla (\%)} & \textbf{\textsc{Hapax} (\%)} \\
\midrule
AG News & 0.3 $\pm$ 0.1 & \textbf{7.4 $\pm$ 0.7}$^\dagger$ & Antonym & 0.6 $\pm$ 0.3 & \textbf{2.5 $\pm$ 0.6}$^\dagger$ \\
Capitalize (Full Word) & \textbf{78.2 $\pm$ 2.8}$^\dagger$ & 59.0 $\pm$ 3.4 & Capitalize First Letter & 36.5 $\pm$ 3.3 & \textbf{74.0 $\pm$ 3.0}$^\dagger$ \\
Capitalize Last Letter & 3.1 $\pm$ 1.2 & \textbf{6.3 $\pm$ 1.7}$^\dagger$ & Capitalize Second Letter & 0.3 $\pm$ 0.3 & \textbf{3.2 $\pm$ 1.2}$^\dagger$ \\
Chinese\textrightarrow{English} & 4.1 $\pm$ 1.2 & \textbf{6.8 $\pm$ 1.6}$^\dagger$ & CommonsenseQA & \textbf{11.2 $\pm$ 0.9}$^\dagger$ & 10.0 $\pm$ 0.8 \\
Country-Capital & 30.7 $\pm$ 6.6 & \textbf{42.3 $\pm$ 7.0}$^\dagger$ & Country-Currency & 5.4 $\pm$ 4.6 & \textbf{8.6 $\pm$ 5.7} \\
French\textrightarrow{English} & 11.0 $\pm$ 2.0 & \textbf{11.9 $\pm$ 2.0} & Indonesian\textrightarrow{English} & \textbf{1.0 $\pm$ 0.6} & 0.6 $\pm$ 0.5 \\
Italian\textrightarrow{English} & 4.8 $\pm$ 1.3 & \textbf{5.1 $\pm$ 1.4} & Japanese\textrightarrow{English} & 0.4 $\pm$ 0.4 & \textbf{2.3 $\pm$ 0.9}$^\dagger$ \\
Landmark-Country & 31.0 $\pm$ 3.1 & \textbf{39.2 $\pm$ 3.3}$^\dagger$ & Lowercase First Letter & 72.4 $\pm$ 3.1 & \textbf{76.8 $\pm$ 2.9}$^\dagger$ \\
Lowercase Last Letter & 3.7 $\pm$ 1.3 & \textbf{10.1 $\pm$ 2.1}$^\dagger$ & National Parks & 17.1 $\pm$ 3.5 & \textbf{22.8 $\pm$ 3.9}$^\dagger$ \\
Next Capital Letter & 3.3 $\pm$ 1.2 & \textbf{8.6 $\pm$ 1.9}$^\dagger$ & Next Item & 12.4 $\pm$ 4.3 & \textbf{27.6 $\pm$ 5.8}$^\dagger$ \\
Park-Country & 11.3 $\pm$ 2.3 & \textbf{16.8 $\pm$ 2.7}$^\dagger$ & Person-Instrument & 0.0 $\pm$ 0.3 & \textbf{3.5 $\pm$ 1.6}$^\dagger$ \\
Person-Occupation & 0.4 $\pm$ 0.4 & \textbf{2.6 $\pm$ 1.1}$^\dagger$ & Person-Sport & 10.1 $\pm$ 3.3 & \textbf{42.8 $\pm$ 5.4}$^\dagger$ \\
Polish\textrightarrow{English} & 1.7 $\pm$ 0.8 & \textbf{2.2 $\pm$ 0.9} & Portuguese\textrightarrow{English} & 4.7 $\pm$ 1.3 & \textbf{8.1 $\pm$ 1.7}$^\dagger$ \\
Present-Past & 55.3 $\pm$ 5.7 & \textbf{78.2 $\pm$ 4.7}$^\dagger$ & Previous Item & 2.7 $\pm$ 2.1 & \textbf{3.6 $\pm$ 2.4} \\
Product-Company & 15.5 $\pm$ 3.1 & \textbf{25.9 $\pm$ 3.8}$^\dagger$ & Sentiment & 2.7 $\pm$ 0.9 & \textbf{36.8 $\pm$ 2.8}$^\dagger$ \\
Singular-Plural & 63.9 $\pm$ 6.6 & \textbf{76.6 $\pm$ 5.8}$^\dagger$ & Spanish\textrightarrow{English} & 7.0 $\pm$ 1.6 & \textbf{11.7 $\pm$ 2.0}$^\dagger$ \\
Synonym & 1.4 $\pm$ 0.4 & \textbf{2.3 $\pm$ 0.6}$^\dagger$ & Word Length & 8.1 $\pm$ 1.9 & \textbf{15.4 $\pm$ 2.5}$^\dagger$ \\
\bottomrule
\end{tabular}
\end{table*}

\begin{table*}[htbp]
\centering
\caption{Performance comparison on \textbf{abstractive} in-context learning tasks (5-Shot) for Vanilla and Thresholded-\textsc{Hapax} by ensuring that the target answer for each test instance is not included in the 5-shot examples. 
Values show accuracy $\pm$ 95\% CI margin. Bold indicates higher performance; $\dagger$ denotes statistical significance ($p < 0.05$) according to McNemar's test.}
\label{tab:abstractive_results_nonextract_thresh}
\scriptsize
\begin{tabular}{lcc@{\hspace{1.8em}}lcc}
\toprule
\textbf{Task} & \textbf{Vanilla (\%)} & \textbf{\textsc{Hapax Thresh} (\%)} &
\textbf{Task} & \textbf{Vanilla (\%)} & \textbf{\textsc{Hapax Thresh} (\%)} \\
\midrule
AG News & 0.3 $\pm$ 0.1 & \textbf{3.4 $\pm$ 0.5}$^\dagger$ & Antonym & 0.6 $\pm$ 0.3 & \textbf{6.9 $\pm$ 1.0}$^\dagger$ \\
Capitalize (Full Word) & \textbf{78.2 $\pm$ 2.8}$^\dagger$ & 35.9 $\pm$ 3.3 & Capitalize First Letter & \textbf{36.5 $\pm$ 3.3}$^\dagger$ & 26.9 $\pm$ 3.0 \\
Capitalize Last Letter & 3.1 $\pm$ 1.2 & \textbf{8.0 $\pm$ 1.9}$^\dagger$ & Capitalize Second Letter & 0.3 $\pm$ 0.3 & \textbf{7.8 $\pm$ 1.9}$^\dagger$ \\
Chinese\textrightarrow{English} & 4.1 $\pm$ 1.2 & \textbf{10.1 $\pm$ 1.9}$^\dagger$ & CommonsenseQA & \textbf{11.2 $\pm$ 0.9}$^\dagger$ & 0.0 $\pm$ 0.0 \\
Country-Capital & \textbf{30.7 $\pm$ 6.6} & 29.1 $\pm$ 6.5 & Country-Currency & \textbf{5.4 $\pm$ 4.6} & 3.2 $\pm$ 3.4 \\
French\textrightarrow{English} & 11.0 $\pm$ 2.0 & \textbf{14.5 $\pm$ 2.2}$^\dagger$ & Indonesian\textrightarrow{English} & 1.0 $\pm$ 0.6 & 1.0 $\pm$ 0.6 \\
Italian\textrightarrow{English} & 4.8 $\pm$ 1.3 & \textbf{6.6 $\pm$ 1.6} & Japanese\textrightarrow{English} & 0.4 $\pm$ 0.4 & \textbf{3.8 $\pm$ 1.2}$^\dagger$ \\
Landmark-Country & \textbf{31.0 $\pm$ 3.1}$^\dagger$ & 25.7 $\pm$ 3.0 & Lowercase First Letter & \textbf{72.4 $\pm$ 3.1}$^\dagger$ & 61.1 $\pm$ 3.3 \\
Lowercase Last Letter & 3.7 $\pm$ 1.3 & \textbf{5.0 $\pm$ 1.5} & National Parks & 17.1 $\pm$ 3.5 & \textbf{24.2 $\pm$ 4.0}$^\dagger$ \\
Next Capital Letter & 3.3 $\pm$ 1.2 & \textbf{6.3 $\pm$ 1.7}$^\dagger$ & Next Item & \textbf{12.4 $\pm$ 4.3} & 9.8 $\pm$ 3.9 \\
Park-Country & 11.3 $\pm$ 2.3 & \textbf{16.0 $\pm$ 2.7}$^\dagger$ & Person-Instrument & 0.0 $\pm$ 0.3 & \textbf{1.6 $\pm$ 1.1}$^\dagger$ \\
Person-Occupation & 0.4 $\pm$ 0.4 & \textbf{7.8 $\pm$ 1.8}$^\dagger$ & Person-Sport & \textbf{10.1 $\pm$ 3.3}$^\dagger$ & 2.5 $\pm$ 1.7 \\
Polish\textrightarrow{English} & 1.7 $\pm$ 0.8 & \textbf{3.0 $\pm$ 1.1} & Portuguese\textrightarrow{English} & 4.7 $\pm$ 1.3 & \textbf{9.9 $\pm$ 1.9}$^\dagger$ \\
Present-Past & \textbf{55.3 $\pm$ 5.7}$^\dagger$ & 38.9 $\pm$ 5.6 & Previous Item & \textbf{2.7 $\pm$ 2.1} & 1.3 $\pm$ 1.4 \\
Product-Company & 15.5 $\pm$ 3.1 & \textbf{20.7 $\pm$ 3.5}$^\dagger$ & Sentiment & \textbf{2.7 $\pm$ 0.9}$^\dagger$ & 0.0 $\pm$ 0.1 \\
Singular-Plural & \textbf{63.9 $\pm$ 6.6}$^\dagger$ & 23.9 $\pm$ 5.8 & Spanish\textrightarrow{English} & 7.0 $\pm$ 1.6 & \textbf{14.2 $\pm$ 2.2}$^\dagger$ \\
Synonym & 1.4 $\pm$ 0.4 & \textbf{3.4 $\pm$ 0.7}$^\dagger$ & Word Length & \textbf{8.1 $\pm$ 1.9}$^\dagger$ & 2.0 $\pm$ 1.0 \\

\bottomrule
\end{tabular}
\end{table*}

\begin{table*}[t]
\centering
\caption{Task categories and their constituent tasks. The four-way categorization follows \citet{hendel-etal-2023-context}, extended with a Sentence-level family covering classification tasks that operate over full passages.}

\label{tab:icl_category_definitions}
\footnotesize
\setlength{\tabcolsep}{6pt}
\renewcommand{\arraystretch}{1.2}
\begin{tabular}{@{}p{0.17\textwidth} p{0.46\textwidth} p{0.30\textwidth}@{}}
\toprule
\textbf{Task family} & \textbf{Tasks included} & \textbf{Description} \\
\midrule
\textbf{Algorithmic} (8) & \begin{minipage}[t]{\linewidth}{\raggedright\footnotesize Capitalize (Full Word), Capitalize First Letter, Capitalize Last Letter, Capitalize Second Letter, Lowercase First Letter, Lowercase Last Letter, Next Capital Letter, Word Length}\end{minipage} & \textit{Tasks solvable by a character-level rule, such as capitalization, character selection, or counting.} \\
\textbf{Linguistic} (4) & \begin{minipage}[t]{\linewidth}{\raggedright\footnotesize Antonym, Present-Past, Singular-Plural, Synonym}\end{minipage} & \textit{Tasks that require lexical or morphological knowledge.} \\
\textbf{Knowledge} (11) & \begin{minipage}[t]{\linewidth}{\raggedright\footnotesize Country-Capital, Country-Currency, Landmark-Country, National Parks, Next Item, Park-Country, Person-Instrument, Person-Occupation, Person-Sport, Previous Item, Product-Company}\end{minipage} & \textit{Tasks that require parametric world knowledge.} \\
\textbf{Translation} (8) & \begin{minipage}[t]{\linewidth}{\raggedright\footnotesize Chinese$\rightarrow$English, French$\rightarrow$English, Indonesian$\rightarrow$English,  Italian$\rightarrow$English, Japanese$\rightarrow$English, Polish$\rightarrow$English, Portuguese$\rightarrow$English, Spanish$\rightarrow$English}\end{minipage} & \textit{Tasks where an input word is mapped from a source language to its target-language equivalent.} \\
\textbf{Sentence-level} (3) & \begin{minipage}[t]{\linewidth}{\raggedright\footnotesize AG News, CommonsenseQA, Sentiment}\end{minipage} & \textit{Classification tasks over full passages.} \\
\bottomrule
\end{tabular}
\end{table*}

\begin{figure}[htbp]
    \centering
\includegraphics[width=\linewidth]{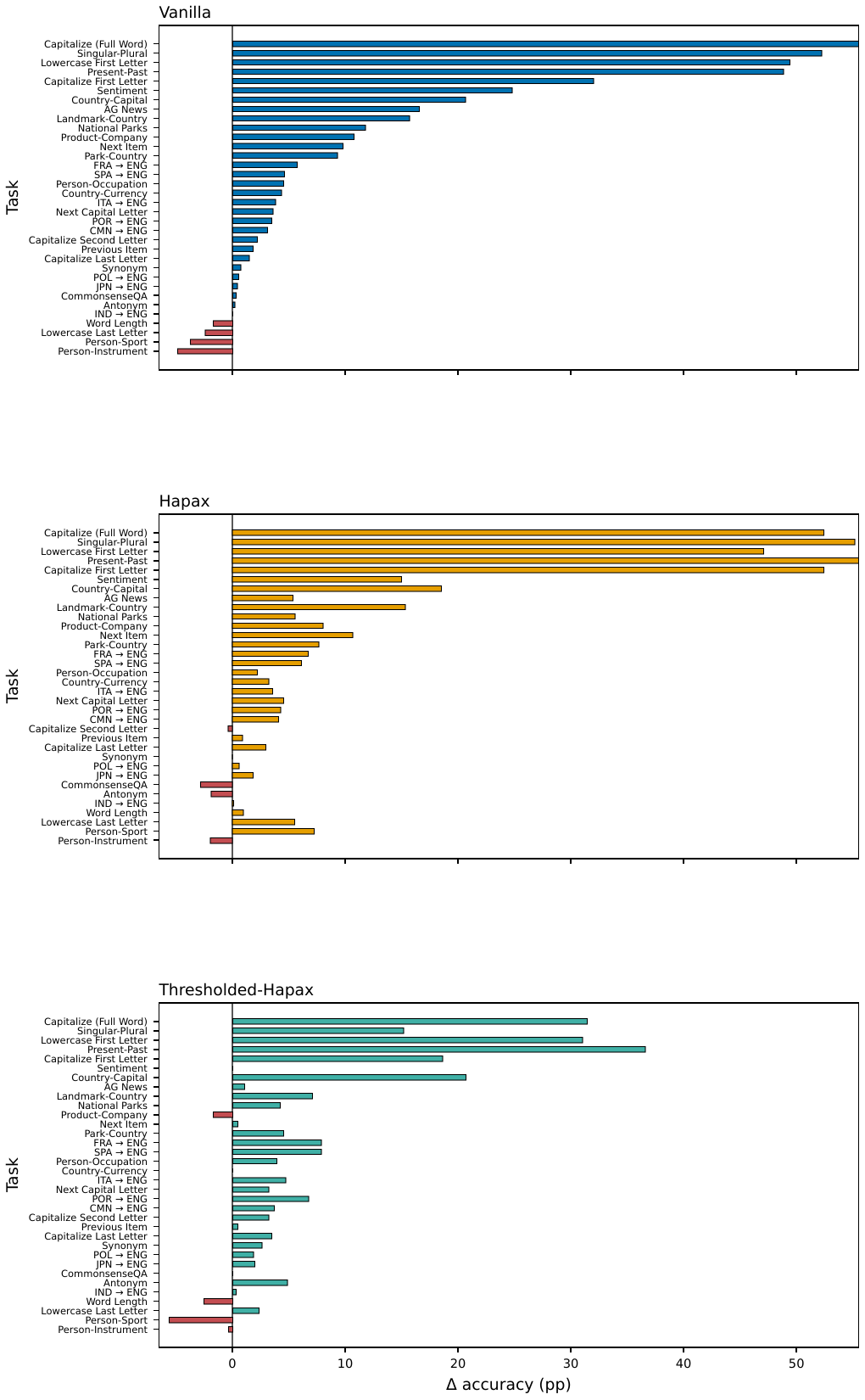}
    \caption{For Vanilla, \textsc{Hapax}, and \textsc{Thresholded-Hapax}, increasing the number of demonstrations from 1-shot to 5-shot improves accuracy on the majority of tasks. This suggests that the models benefit from the in-context demonstrations, rather than relying solely on zero-shot or parametric task knowledge.}
\label{fig:shot_scaling_per_task_stacked_1b_step19900}
\end{figure}

\subsection{Ablation Study on Copying}
\label{app:ablation_stud}

We conduct ablation studies to understand the causal impact of each individual attention heads on the expected prediction $r_s$ on a random repetition task $r_1r_2\ldots r_{s - 1}$. Mean ablation replaces the activation of a model component with its average activation from a reference distribution \citep{wang2023interpretability}. For each head, we compute the mean activation over 10000 samples from the Pile dataset that the model was trained on \citep{gao2020pile800gbdatasetdiverse}. While conducting mean ablations, we only ablate the last token position of the attention heads. We choose the last token to avoid disrupting the influence of the rest of the circuit, mainly the previous token heads. Given a sequence $r_1r_2 \ldots r_s r_1 r_2 \ldots r_{s-1}$, previous token heads store information at position $r_s$ (the final token of the first half) that induction heads use when predicting at position $r_{s-1}$ in the second half. If we were to ablate all token positions, we would also disrupt the information flow from previous token heads, which would mistakenly lead us to identify such heads as induction heads. We measure the probability difference of the target token $r_s$ before and after ablation, $p_{clean}(r_s \mid x) - p_{ablated}(r_s \mid x)$. We also include the ablation results of Vanilla and Thresholded-\textsc{Hapax} for comparison with the \textsc{Hapax} model. For interpreting the results of ablations, we chose to use probability difference metric as it correlated most accurately with task accuracy.

\begin{figure}[htbp]
    \centering
\includegraphics[width=\linewidth]{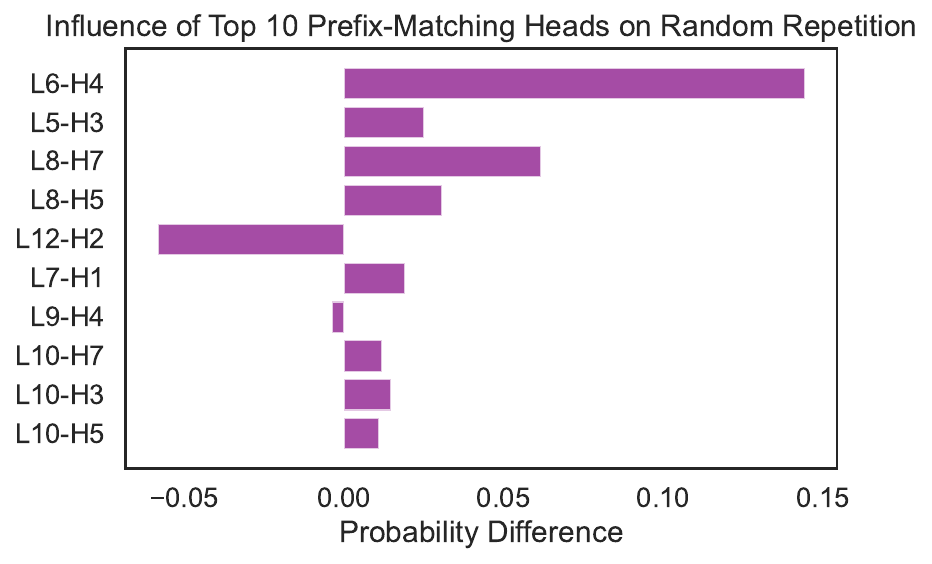}
    \caption{Probability Difference for random repetition task of the Top 10 Prefix-Matching heads for vanilla model.}
\label{fig:prefix_matching_vs_induction_clean_1b}
\end{figure}

\begin{figure}[htbp]
    \centering
        \includegraphics[width=\linewidth]{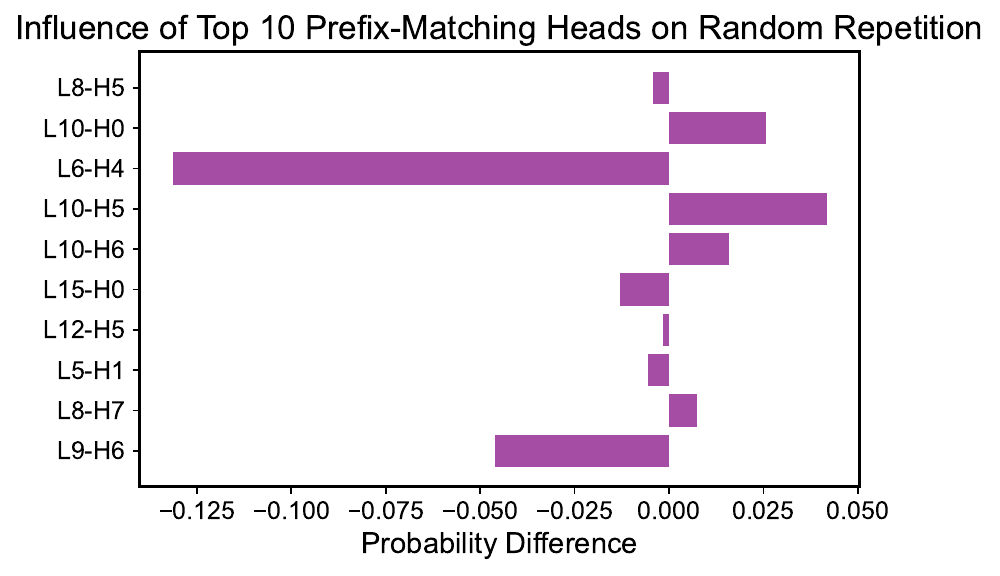}
    \caption{Probability Difference for random repetition task of the Top 10 Prefix-Matching heads for Hapax model.}
    \label{fig:prob_diff_hapax}
\end{figure}

\begin{figure}[htbp]
    \centering
\includegraphics[width=\linewidth]{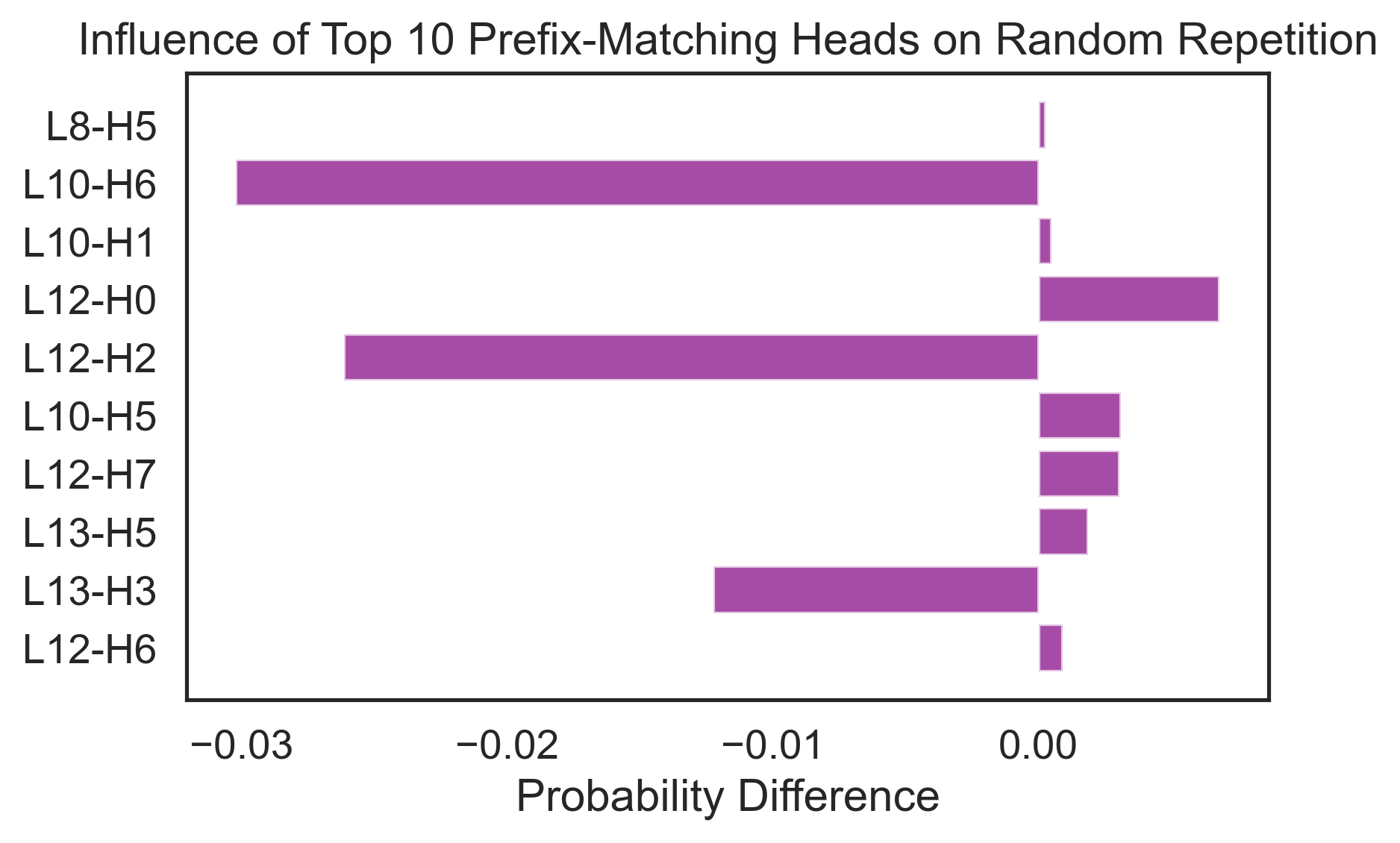}
    \caption{Probability Difference for random repetition task of the Top 10 Prefix-Matching heads for Thresholded-\textsc{Hapax}.}
\label{fig:prefix_matching_vs_induction_masked_bigram_loss_1b_thresh0.3_eq}
\end{figure}

\begin{figure}[htbp]
    \centering
\includegraphics[width=\linewidth]{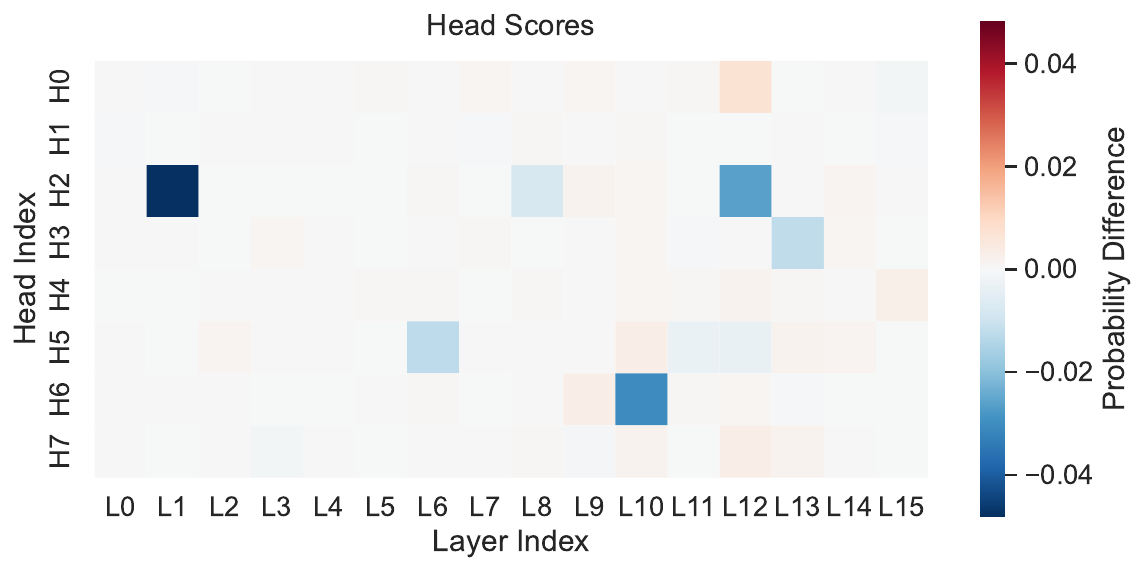}
    \caption{Probability difference values of random repetition task for Thresholded-\textsc{Hapax}.}
    \label{fig:masked_bigram_loss_1b_thresh0.3_eq_head_scores_prob_diff}
\end{figure}

\begin{figure}[htbp]
    \centering
        \includegraphics[width=\linewidth]{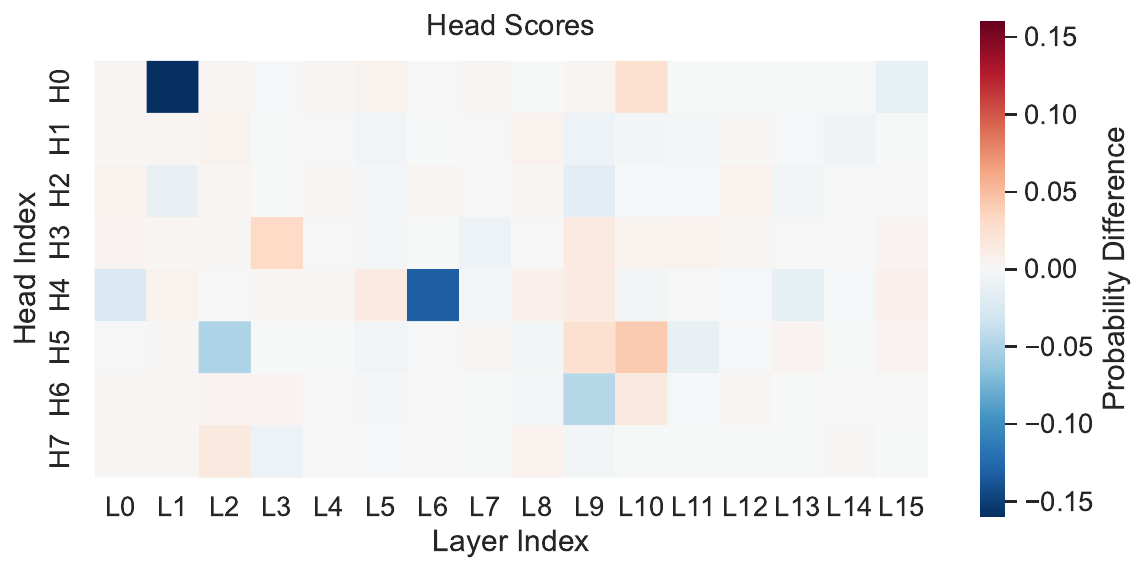}
    \caption{Probability difference values of random repetition task for \textsc{Hapax}.}
    \label{fig:masked_bigram_loss_1b_head_scores_prob_diff}
\end{figure}

\begin{figure}[htbp]
    \centering
\includegraphics[width=\linewidth]{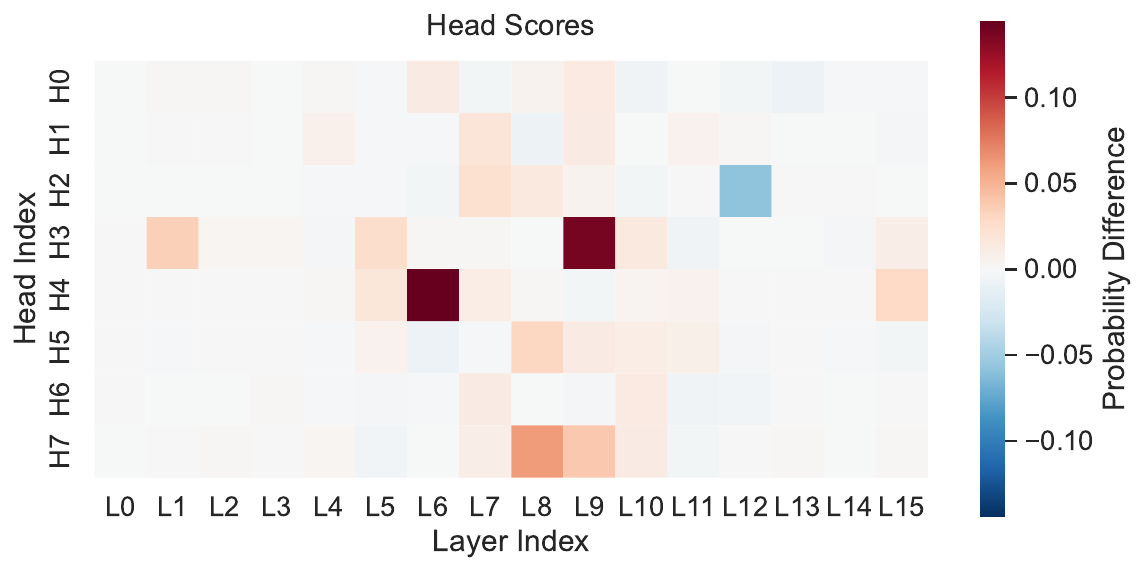}
    \caption{Probability difference values of random repetition task for Vanilla model.}
    \label{fig:clean_1b_head_scores_prob_diff}
\end{figure}

\begin{figure}[htbp]
    \centering
\includegraphics[width=\linewidth]{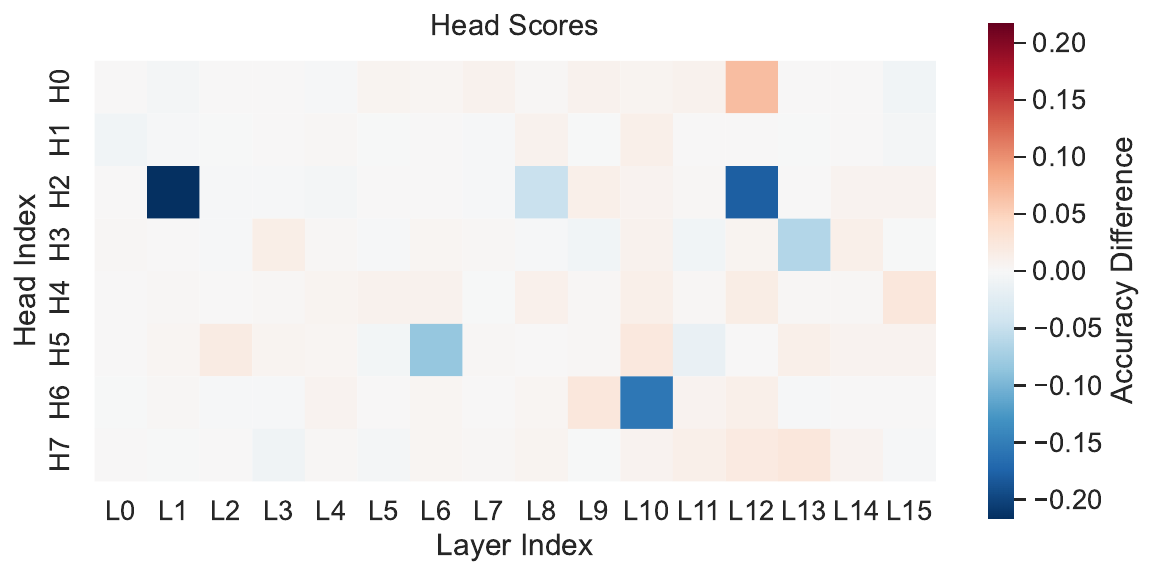}
    \caption{Accuracy difference values of random repetition task for Thresholded-\textsc{Hapax}.}
    \label{fig:masked_bigram_loss_1b_thresh0.3_eq_head_scores_acc_diff}
\end{figure}

\begin{figure}[htbp]
    \centering
\includegraphics[width=\linewidth]{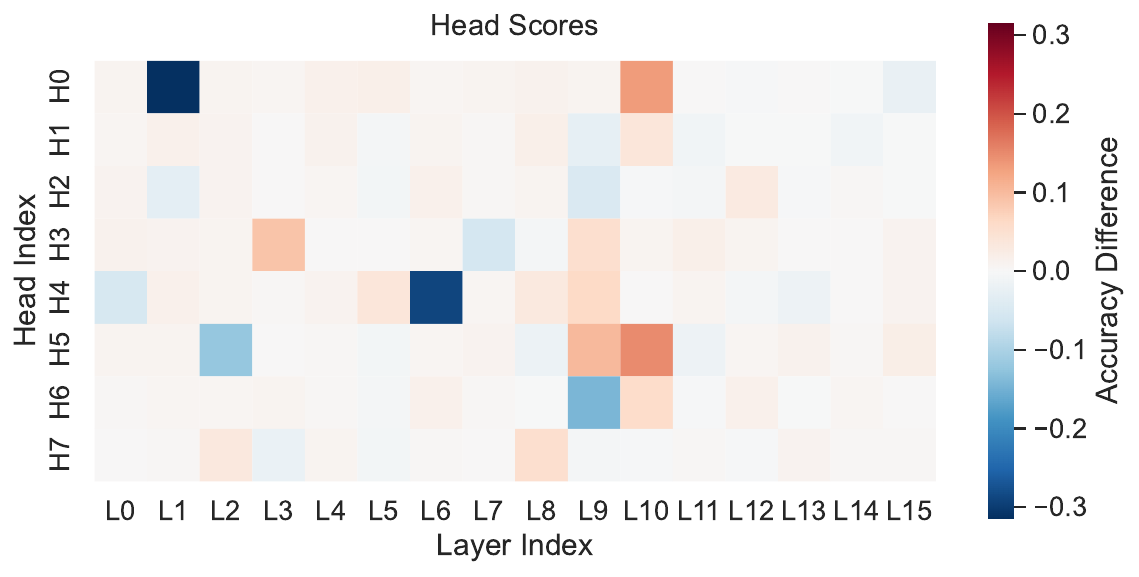}
    \caption{Accuracy difference values of random repetition task for \textsc{Hapax}}
    \label{fig:masked_bigram_loss_1b_head_scores_acc_diff}
\end{figure}

\begin{figure}[htbp]
    \centering
\includegraphics[width=\linewidth]{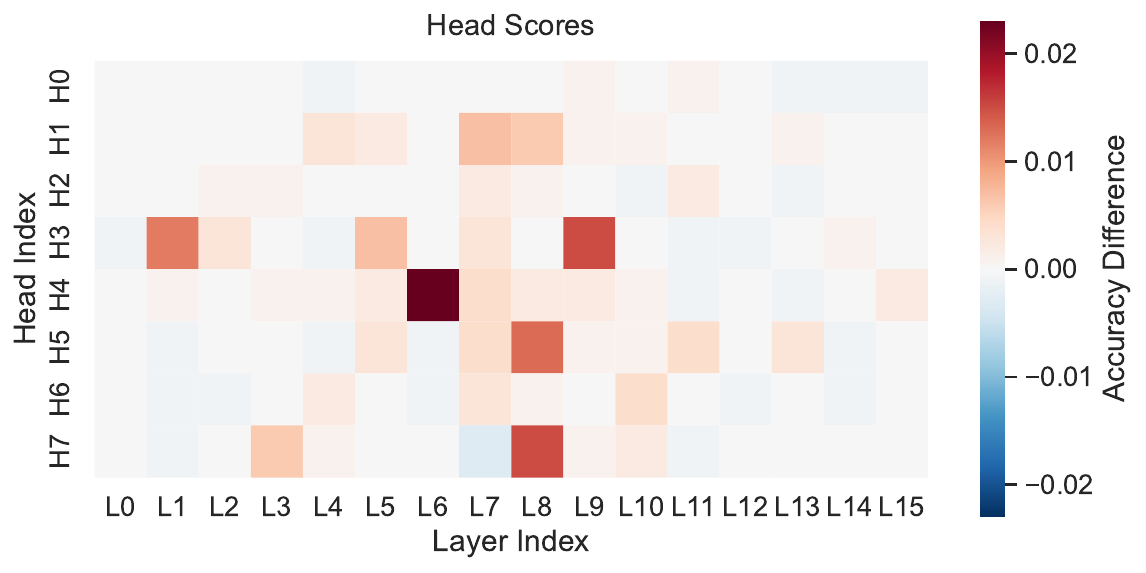}
    \caption{Accuracy difference values of random repetition task for Vanilla model.}
    \label{fig:clean_1b_head_scores_acc_diff}
\end{figure}

\begin{figure}[htbp]
    \centering
    \includegraphics[width=\linewidth]{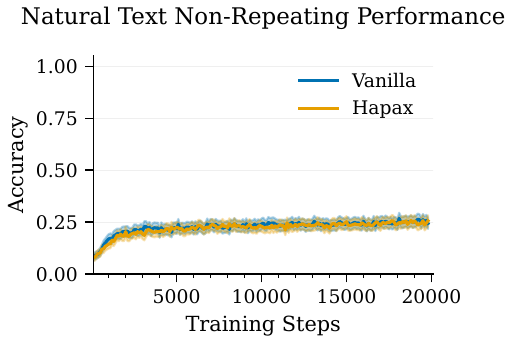}
    \caption{The non-repeating natural text repetition accuracy. Instead of giving a repeated sequence $r_1r_2...r_sr_1r_2...r_{s-1}$ we only give $r_1r_2...r_{s-1}$. This shows that the initial increase seen in the natural text repetition also occurs for non-repeating samples.}\label{fig:natural_text_non_repeating_clean_1b_vs_masked_bigram_loss_1b}
\end{figure}

\subsection{Loss Analysis}
\label{app:loss_analysis}
\begin{figure}[htbp]
    \centering
    \includegraphics[width=\linewidth]{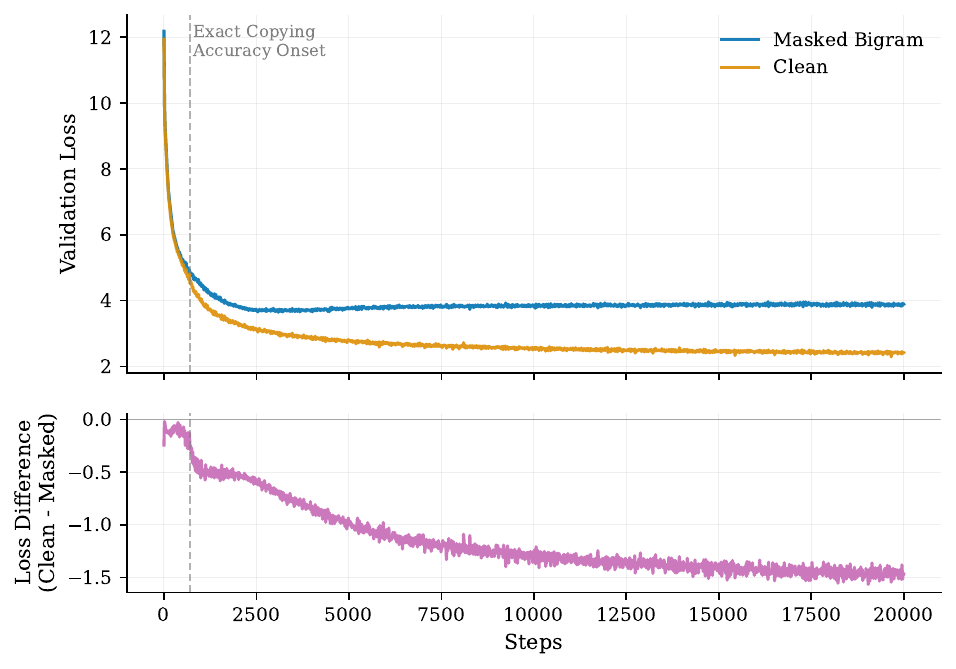}
    \caption{Validation Loss of Vanilla and Hapax model. Despite having increased loss value, as mentioned in the main paper, the validation loss also is mostly affected by the prediction of exact copying instances.}
\label{fig:validation_loss_comparison_1b}
\end{figure}

\begin{figure*}[htbp]
    \centering
        \includegraphics[width=\linewidth]{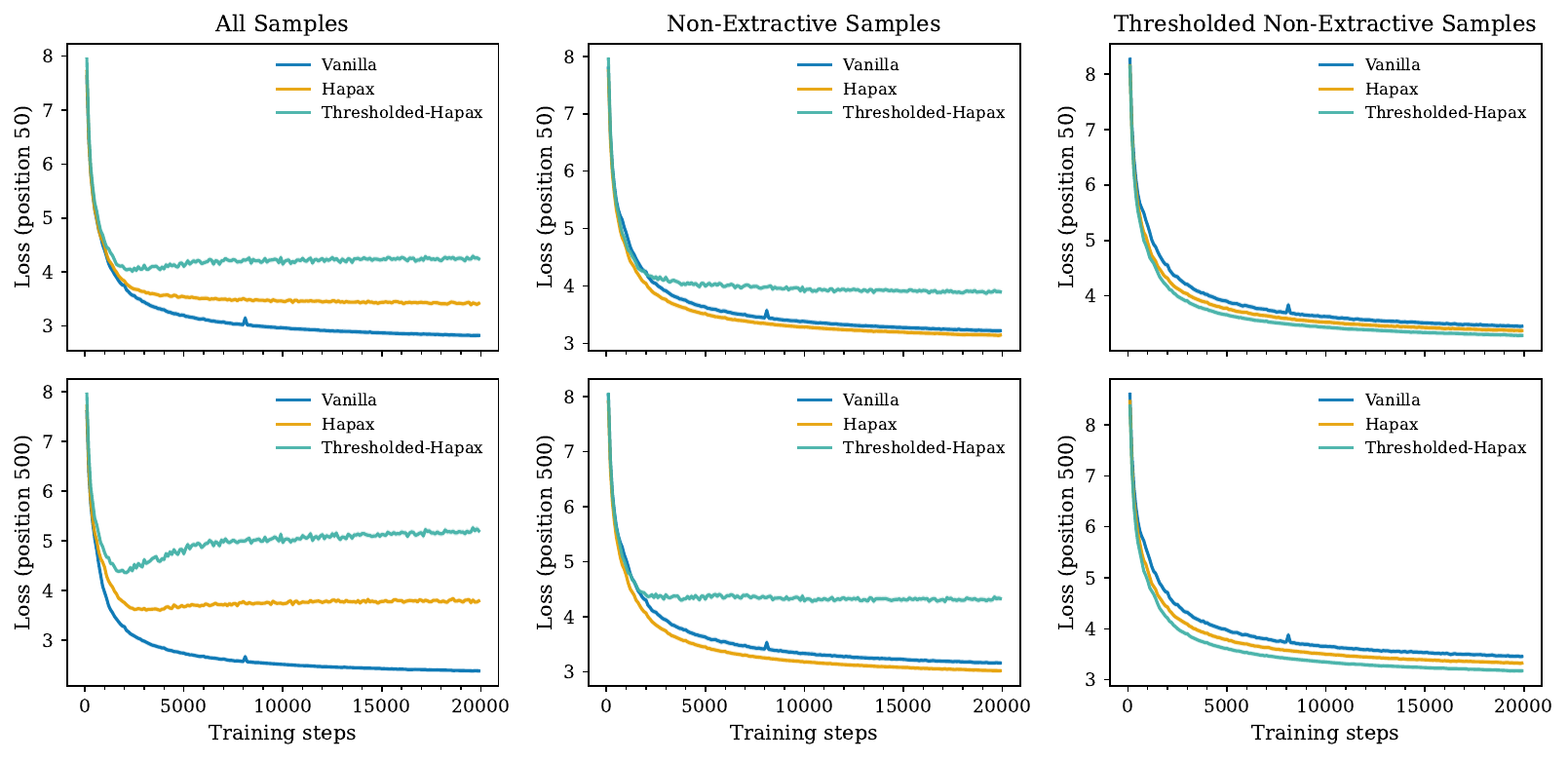}
    \caption{Loss values at individual token positions. The loss value for Non-Extractive Samples is lower for both token positions than the Vanilla in \textsc{Hapax}.}
    \label{fig:position_losses_combined}
\end{figure*}

Figure \ref{fig:validation_loss_comparison_1b} and \ref{fig:position_losses_combined} display the loss relevant analysis.

\subsection{Logit Lens on Prefix-Matching Attention Heads}
\label{app:logitlens}
\begin{figure*}[t]
    \centering
\includegraphics[width=\linewidth]{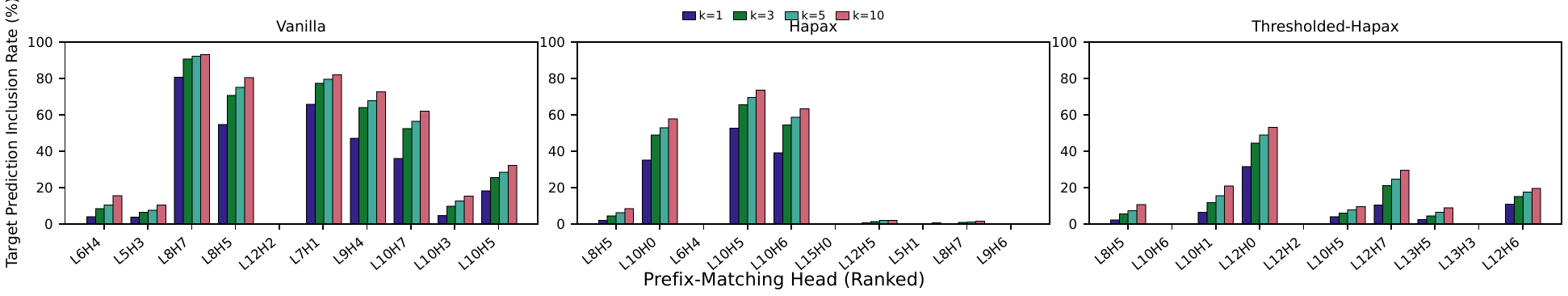}
\caption{Per-head analysis of the percentage of times the target token is included in the top-k promoted tokens for the top 10 prefix-matching heads.}
    \label{fig:logit_lens_per_head}
\end{figure*}

Within \ref{fig:logit_lens_per_head}, we give the per-head contributions of logit lens analysis. 

\subsection{Function-Vector and Concept Head Analysis}
\label{app:fv_analysis}
\begin{figure}[h]
    \centering
\includegraphics[width=\linewidth]{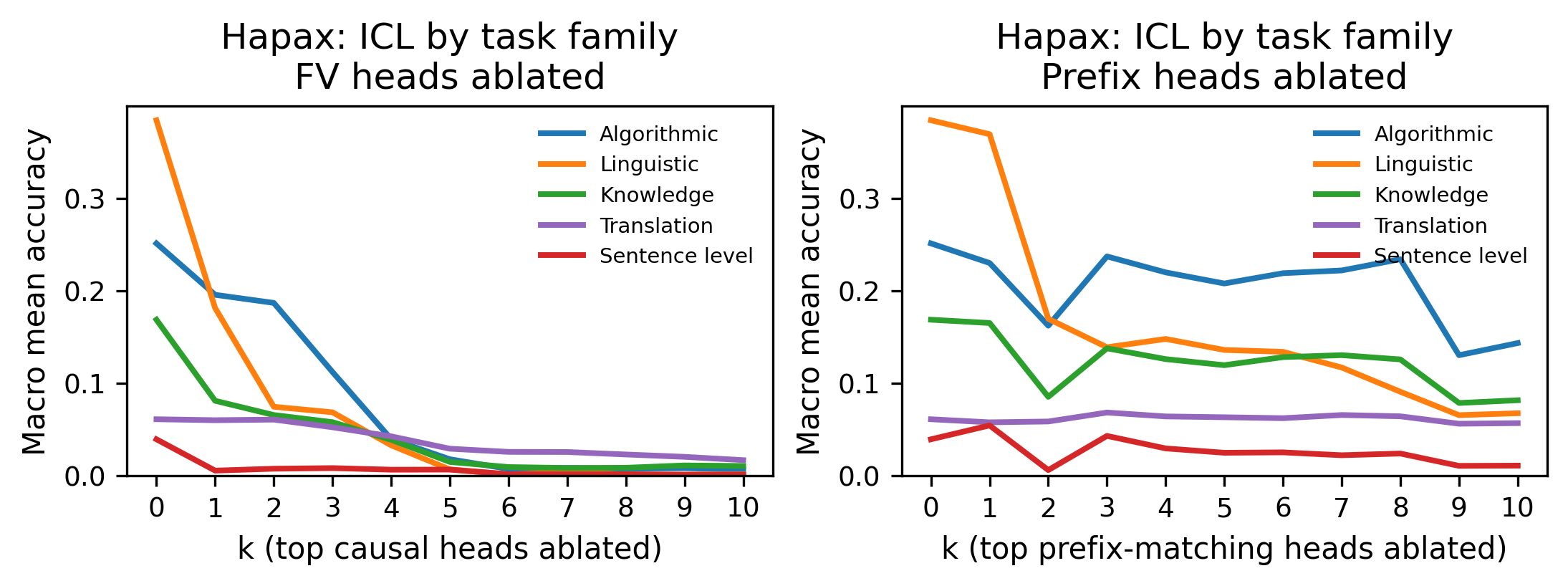}
    \caption{Similar to the findings from \cite{yin2025attentionheadsmatterincontext}, we find FV heads and find them to be causally more important than heads with high prefix-matching scores.}
\label{app:FV_ablation}
\end{figure}

\begin{figure}[h]
    \centering
\includegraphics[width=\linewidth]{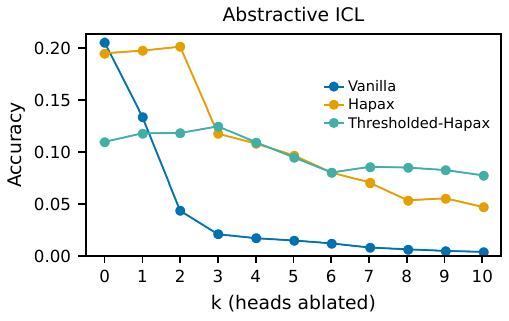}
    \caption{Ablation of previous-token head is more destructive in Vanilla, whereas the influence of ablation is progressively lower on the accuracy for hapax and thresholded hapax.}
\label{app:prev_token_abl}
\end{figure}

\begin{figure}[h]
    \centering
\includegraphics[width=\linewidth]{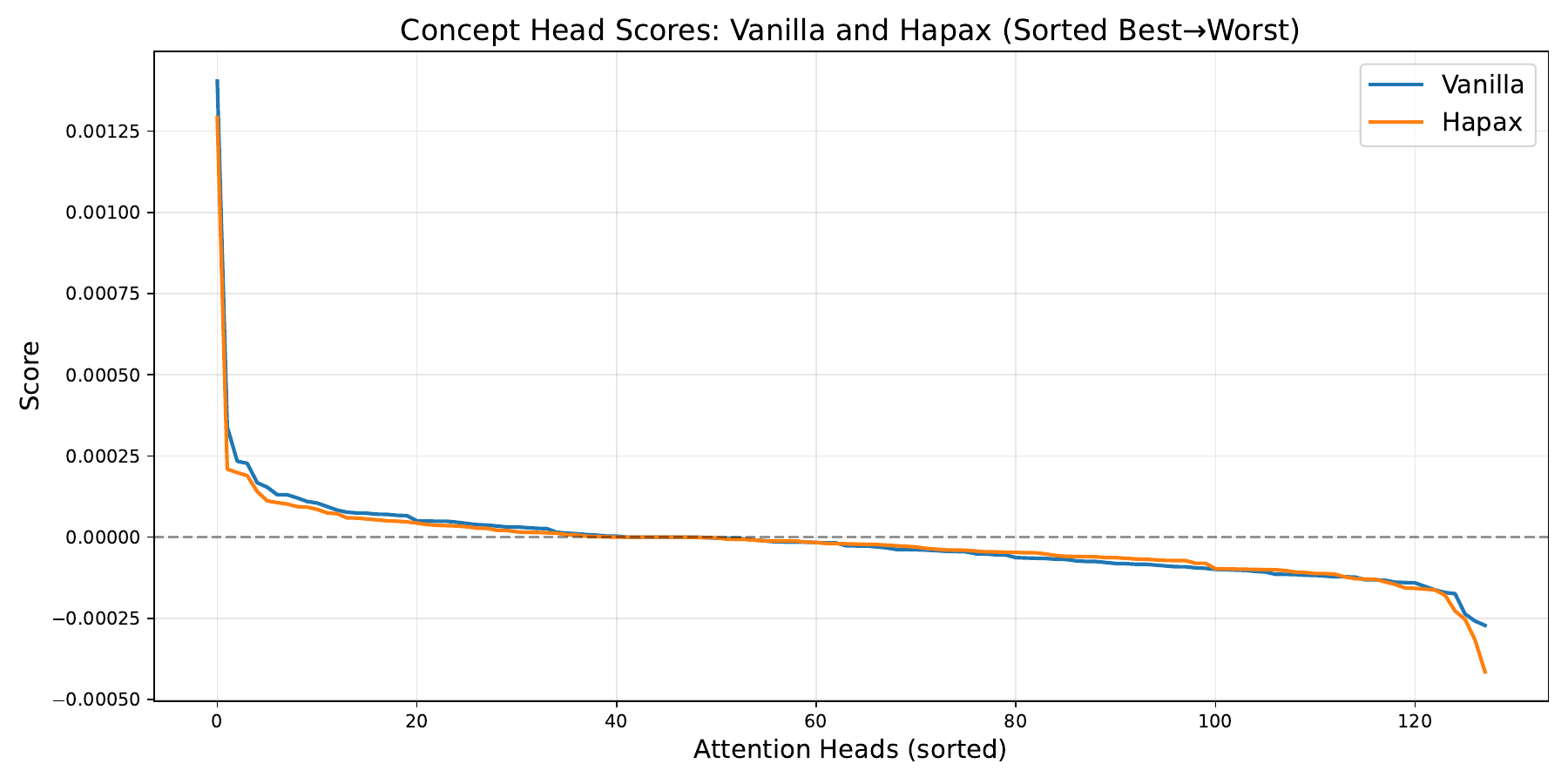}
    \caption{Using the concept head scores from \citet{feucht2025the}, we display the scores of the models from best to worst. The trends are similar for both the Vanilla and \textsc{Hapax} models, indicating that suppressing inductive copying does not noticeably affect the causal concept head metrics.}
\label{app:causal_concept_Scores}
\end{figure}

Within this section, we use Causal Mediation Analysis to find Function Vector heads proposed by \citet{todd2024function} and Concept Head Score analysis from \citet{feucht2025the} to analyze whether the \textsc{Hapax} model can develop such heads that have been found to be important for In-Context Learning (Figure~\ref{app:FV_ablation} and Figures~\ref{app:causal_concept_Scores}). Using this, we find that the \textsc{Hapax} model can develop Function Vector and Concept heads. Secondly, we verify that Function Vector heads are causally important for abstractive ICL, which is in line with \cite{yin2025attentionheadsmatterincontext} (Figure~\ref{app:FV_ablation}). Another finding is that as we ablate previous-token heads, we observe that within the Vanilla model, only ablating the top 2 previous-token heads is enough to pull the accuracy to near 0. However, with Hapax and Thresholded-Hapax, we observe that the distribution these models are trained on created a circuitry where progressively the influence of previous-token heads becomes less destructive, meaning that the training provides more distributed ICL strategies rather than the tightly coupled version in Vanilla model (Figure~\ref{app:prev_token_abl}).

\subsection{Attention Pattern Scores}
\label{app:att_pattern}
\begin{figure}[h]
    \centering
        \includegraphics[width=0.8\linewidth]{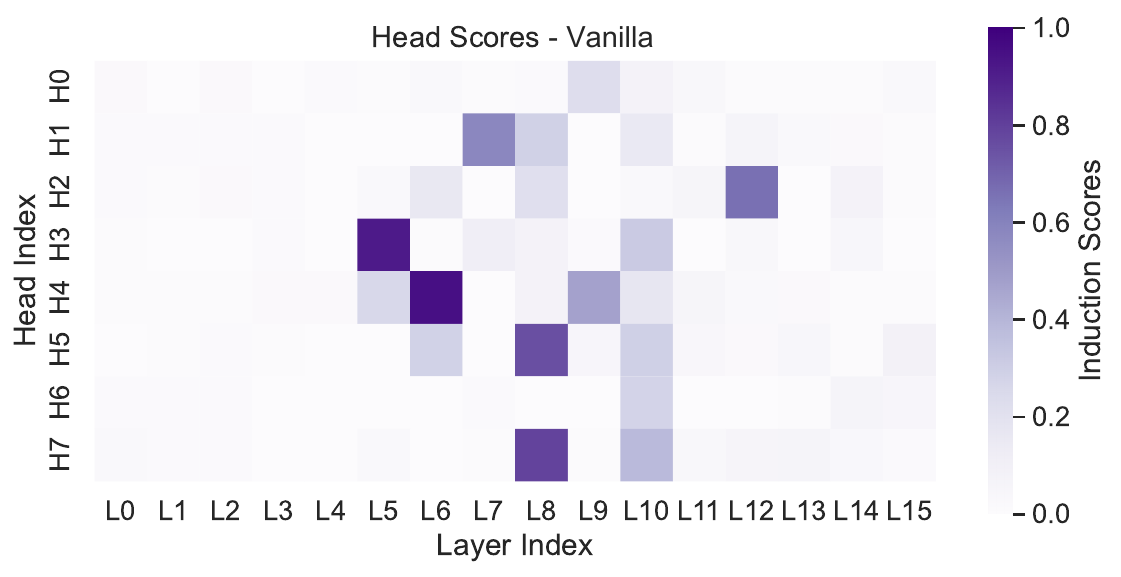}
        \includegraphics[width=0.8\linewidth]{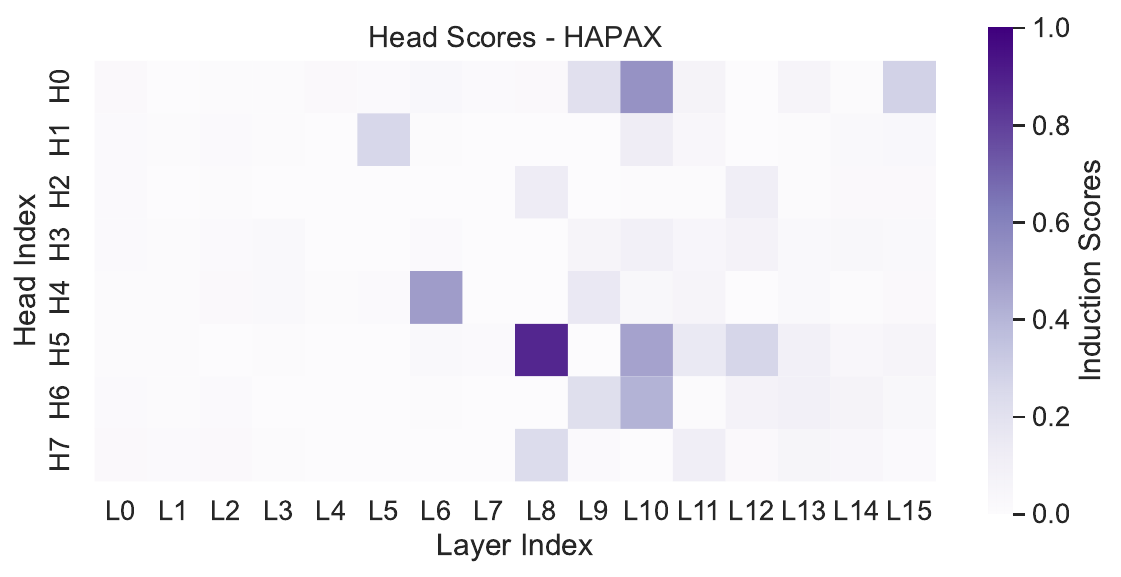}

    \caption{Prefix-Matching Scores across Vanilla and \textsc{Hapax} models. \textsc{Hapax} model has fewer heads that have a strong prefix-matching score.}
    \label{fig:comparison_across_models}
\end{figure}

\begin{figure}[htbp]
    \centering
\includegraphics[width=\linewidth]{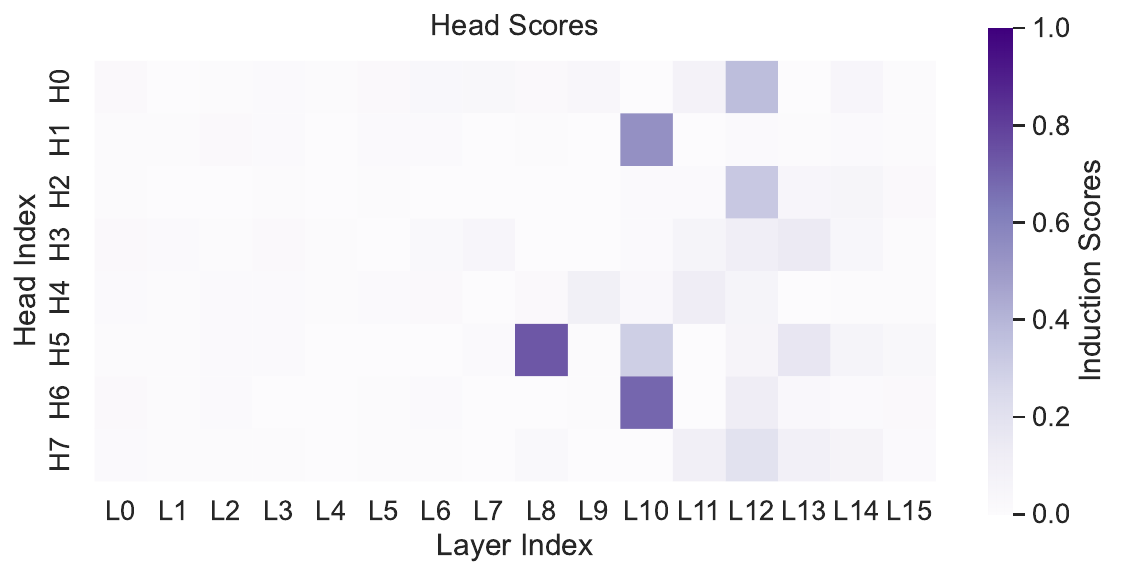}
    \caption{Prefix Matching Scores for Thresholded-\textsc{Hapax}.}
    \label{fig:masked_bigram_loss_1b_thresh0.3_eq_head_scores_induction}
\end{figure}

\begin{figure}[htbp]
    \centering
\includegraphics[width=\linewidth]{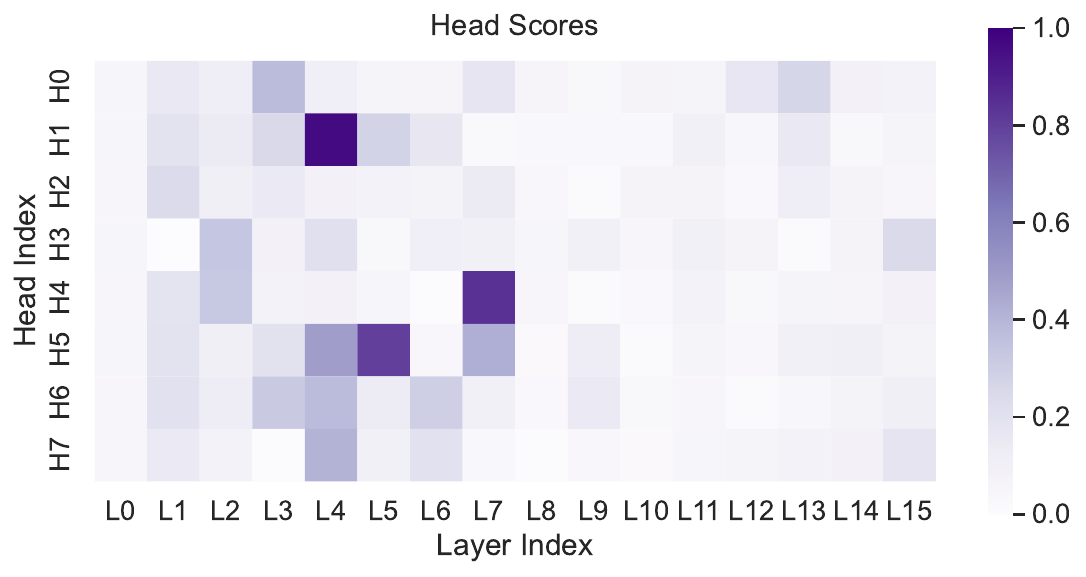}
    \caption{Previous Token Head of vanilla model at step 5000. L4 is the first layer that has a strong previous token head, which is L4 H1.}
    \label{fig:prefix-matching_step5000}
\end{figure}

Figures \ref{fig:comparison_across_models}, \ref{fig:masked_bigram_loss_1b_thresh0.3_eq_head_scores_induction}, \ref{fig:prefix-matching_step5000} display attention patterns of the trained models.

\section{Cosine Similarity Threshold Selection}
\label{app:cos_thresh}

\begin{figure}[htbp]
    \centering
    \includegraphics[width=\linewidth]{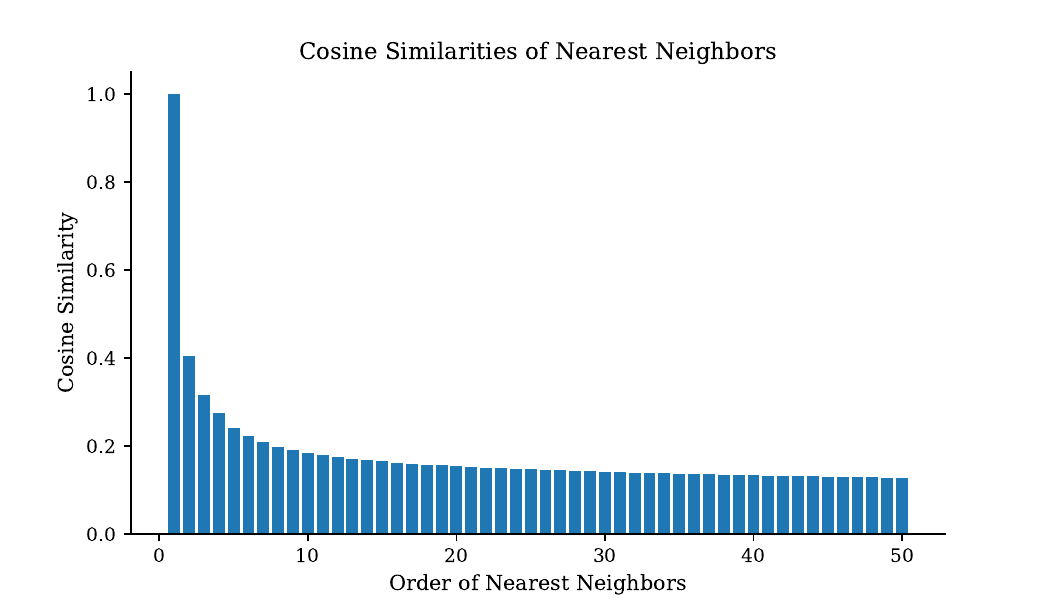}
    \caption{Cosine similarities of nearest N
    neighbors for clean model.}
\label{fig:cosine_similarities_of_nearest_neighbors}
\end{figure}

\begin{figure}[htbp]
    \centering
    \includegraphics[width=\linewidth]{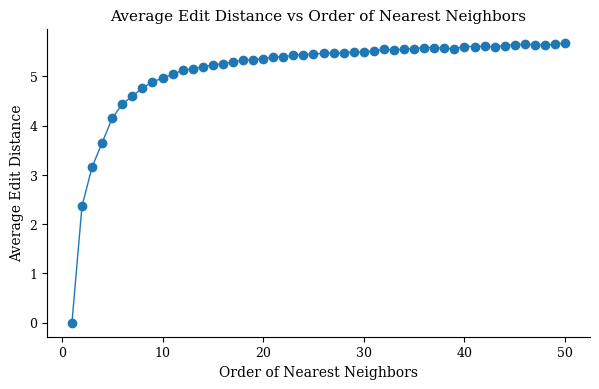}
    \caption{Average edit distance of nearest neighbors for vanilla model.}
\label{fig:avg_edit_Dist}
\end{figure}

The main purpose of Thresholded-\textsc{Hapax} is to create a stricter training regime for suppression of inductive copying. Our hypothesis is that even if we mask exact verbatim copying signals, there might still be token representations with high representational similarity, high enough to still provide a weak signal for verbatim copying. To test this, we select the cosine similarity threshold by using the embedding space of the last step of the vanilla model as a proxy for finding similar tokens. We create a cosine similarity matrix and analyze the cosine similarities of the closest neighbors and their edit distances. After doing this, we select the threshold of 0.3, which corresponds to the top-4 closest neighbors on average (Figure~\ref{fig:cosine_similarities_of_nearest_neighbors}) and an average edit distance of 3.65 (Figure~\ref{fig:avg_edit_Dist}). Tokens representing different strings may still have highly similar embeddings, so our thresholding deliberately overestimates these hypothesized similar tokens, which makes us more confident that we are removing copying signal. Given our limited compute budget, we adopt this strict threshold to ensure that the trained model consistently masks the majority of similar tokens. Although this eventually causes harm for abstractive tasks when compared to the regular \textsc{Hapax} model, we observe consistent gains for translation, suggesting that the model keeps capabilities for such instances despite the heavy masking. This control over copying also cannot solely be explained by undertraining, because the Thresholded-\textsc{Hapax} model is near zero at copying while still achieving 29.1\% accuracy on non-trivial tasks like Country-Capital. Therefore, even if it performs worse than the vanilla model, it serves as a proof of concept that the model can achieve non-trivial capabilities in abstractive tasks despite having even more reduced copying and prefix-matching scores.

\begin{figure}[htbp]
    \centering
    \includegraphics[width=\linewidth]{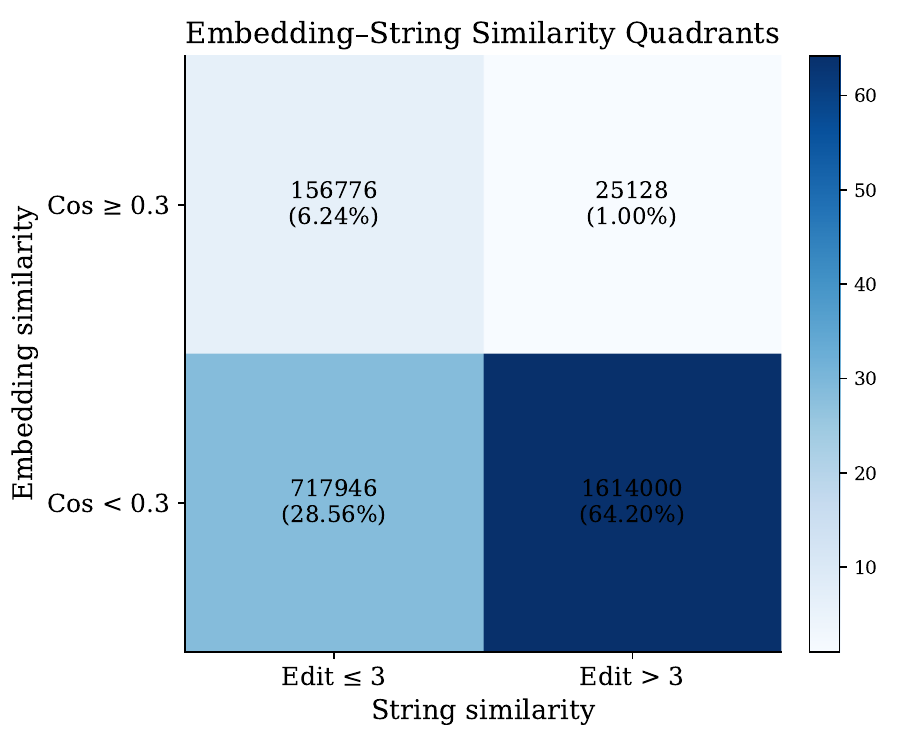}
    \caption{Relationship between string-level similarity (edit distance) and embedding similarity for the top-50 nearest neighbors. Many token pairs with low edit distance also have low embedding similarity (28.56\%). However, 1\% of the token pairs with cosine similarity above 0.3 are still not captured. Since attention heads attend by hidden state similarity, we want to make sure we are removing such similar tokens, which motivates our use of cosine similarity.}
\label{fig:conf_matrix}
\end{figure}

To further justify our use of input embedding cosine similarity (instead of edit distance), we analyze the relationship between string-level edit distance and embedding cosine similarity for the top-50 nearest neighbors in the vanilla model's embedding space (Figure~\ref{fig:conf_matrix}). With a cosine similarity threshold of 0.3, 7.24\% of the token pairs are treated as equal, and only 1\% of these pairs have an edit distance higher than 3. In contrast, if we instead threshold by edit distance at 3, 34.8\% of the pairs are treated as equal, and 28.56\% of these have a cosine similarity below 0.3. Moreover, an edit distance threshold of 3 still fails to capture 1\% of the pairs whose cosine similarity is above 0.3. Therefore, edit distance treats many more pairs as equal but still does not capture some high-cosine-similarity token pairs, which we hypothesize are more closely related to copying signal. Since attention heads attend by hidden state similarity, cosine similarity is closer to what an attention head would consider similar. We use cosine similarity as our base and use the average edit distance at different cosine similarity thresholds to guide us, so that we make sure we are removing tokens above a certain cosine similarity while also understanding how close they are on average to their represented tokens.

\begin{figure}[htbp]
    \centering
    \includegraphics[width=\linewidth]{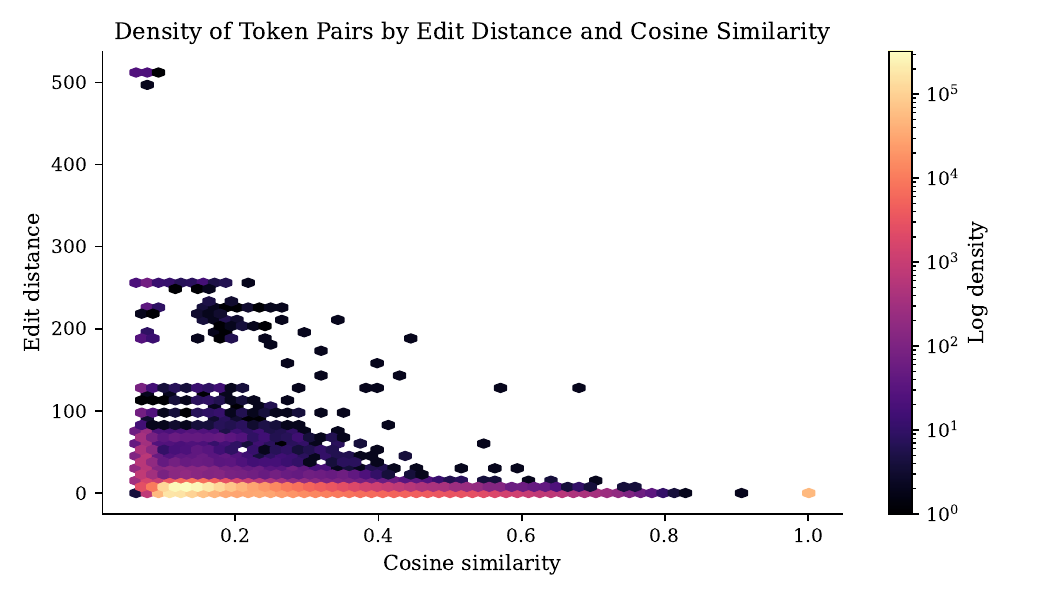}
    \caption{Low edit-distance pairs appear across a broad range of cosine similarities, suggesting that edit distance over-treats tokens as equivalent. The analysis is done on the same pairs analyzed in \ref{fig:avg_edit_Dist}.}
\label{fig:joint_distribution_of_cosine_similarity_and_edit_distance}
\end{figure}

\section{\textsc{Hapax} Fluency Evaluation}
\label{app:fluency}

\begin{figure}[h]
    \centering
    \begin{subfigure}[b]{\linewidth}
        \centering
        \includegraphics[width=\linewidth]{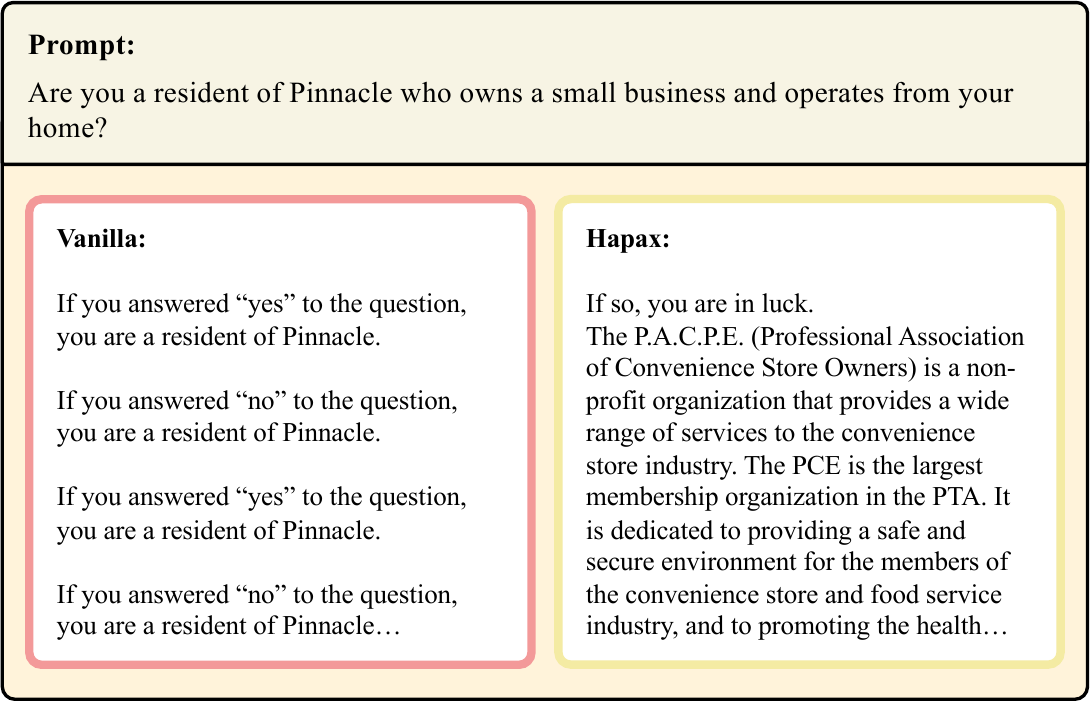}
    \end{subfigure}
    \begin{subfigure}[b]{\linewidth}
        \centering
        \includegraphics[width=\linewidth]{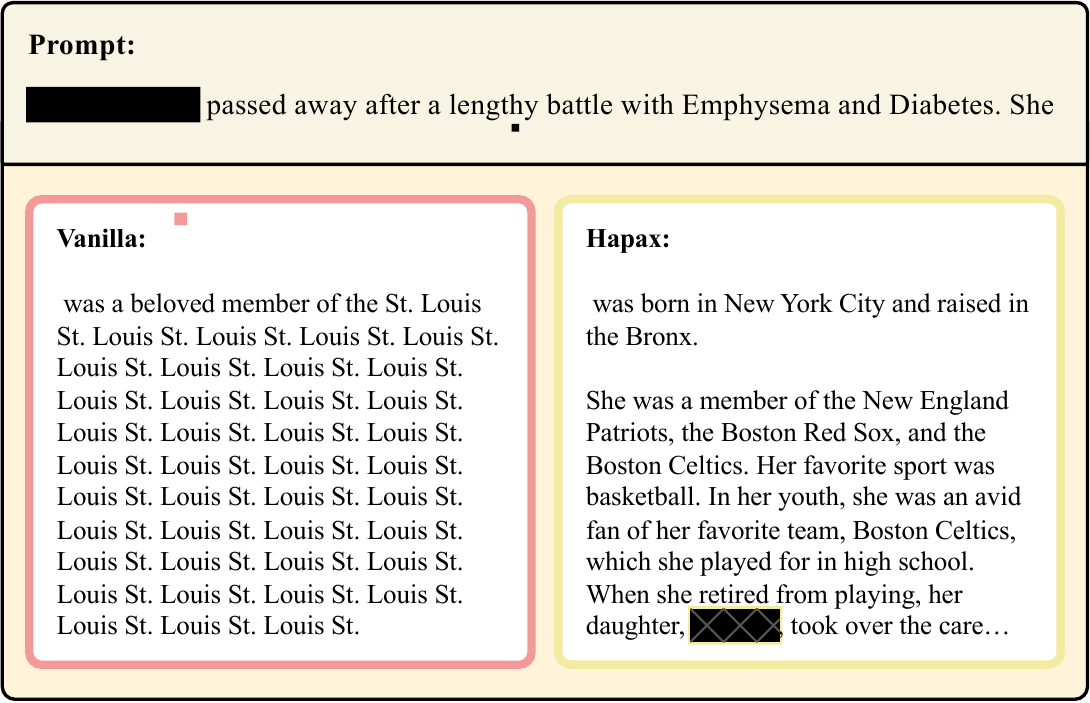}
    \end{subfigure}
    \caption{Examples of sentence completions generated for the evaluation. Each prompt seed is used to generate completions by both models. In both examples, the vanilla model's generation was rated with 0 (Not Fluent), while \textsc{Hapax} received 1 (Somewhat Fluent).  Names in the second example have been redacted.}
    \label{fig:sentence_completion_examples}
\end{figure}

We evaluate the effects of \textsc{Hapax} training on the model’s ability to generate fluent natural text and compare the results to its vanilla counterpart as a baseline. Inspired by the Fluency score from Axbench \citep{wu2025axbenchsteeringllmssimple}, we use an automated LLM-based approach to assess the fluency of sentence completions generated by each model, with discrete ratings of 0, 1, and 2 for fluent language. Sequence continuations, with a maximum length of 128 tokens, are generated based on a diverse seed of 20 tokens, sampled from the RedPajama dataset \citep{weber2024redpajama}.

\begin{figure}[htbp]
    \centering
    \includegraphics[width=\linewidth]{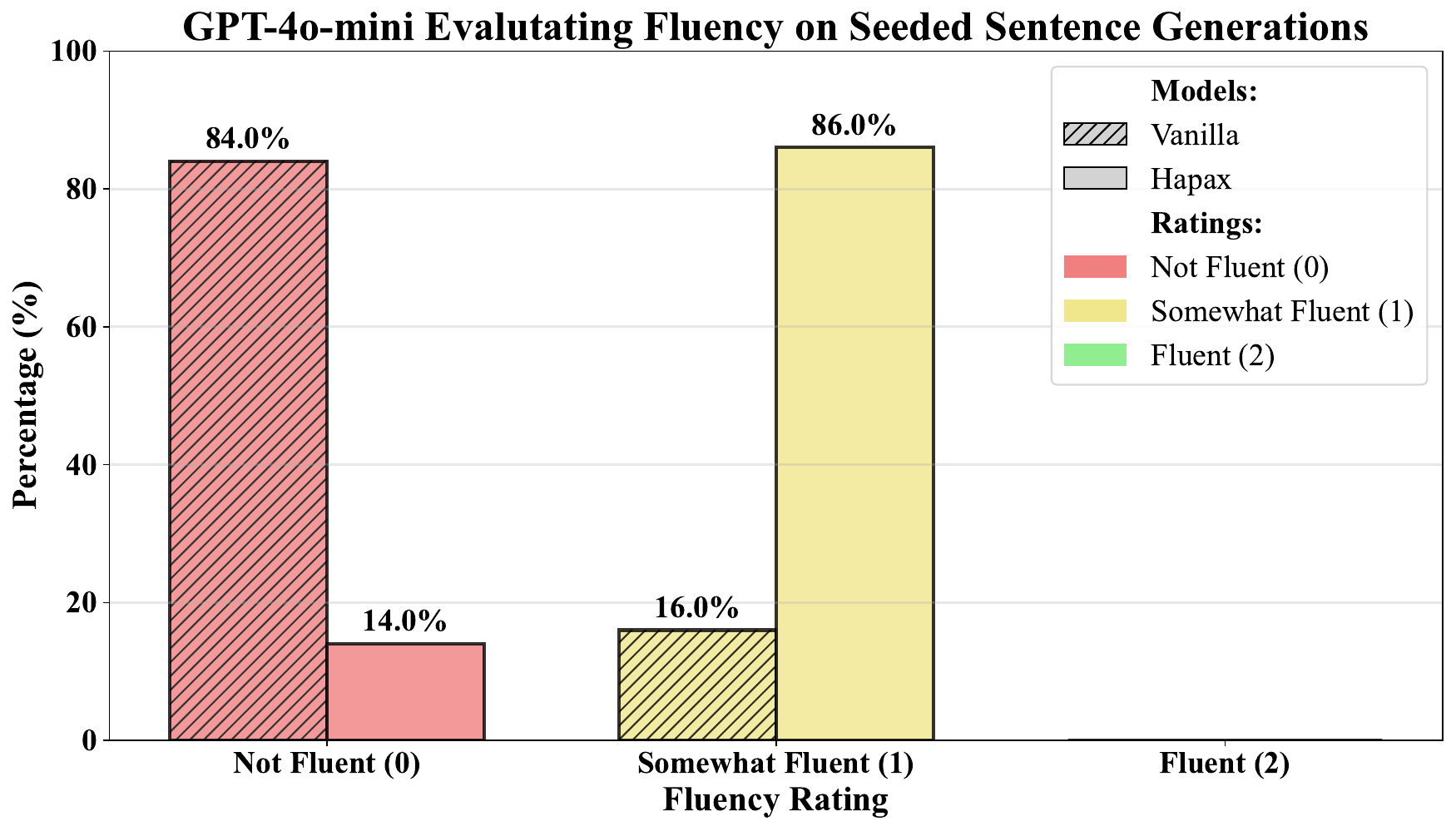}
    \caption{GPT-4o-mini as a judge rates 100 samples of natural text generated by each model based on fluency.}
\label{fig:fluency_comparison_1b}
\end{figure}

We find in Figure \ref{fig:fluency_comparison_1b} that our \textsc{Hapax} trained model generates somewhat fluent text 70\% more times than the vanilla model. The latter, performing poorly, generates non-fluent language 84\% of the time. Looking at some qualitative examples of generation in Figure \ref{fig:sentence_completion_examples}, we observe that the vanilla model defaults to repeating itself rather early, creating incoherent language, which explains its negative results. On the other hand, the \textsc{Hapax} model is able to produce chained narratives with much more fluid and natural structure.

\section{Experimental Details}
\label{sec:experimental_detail}
\subsection{Number of Samples Used \& Data Selection}
After giving the few-shot examples, we let the model generate 8 tokens. If the correct prediction requires more than 8 token predictions, we discard that sample. Secondly, since the model is trained with a context-length of 2048, samples that exceed the trained context length are also discarded. According to that logic, the number of samples available and used are given in Table~\ref{app:dataset_info}. Each few-shot demonstration is sampled randomly from the available items with a seed of 1234.

\begin{table*}[t]
\centering
\caption{Per-task evaluation sample sizes for FV ICL tasks. \textbf{Pool} is the benchmark size. \textbf{Used} is the number of samples used after the filtering logic is applied.}
\label{tab:evaluation_sample_counts}
\scriptsize
\setlength{\tabcolsep}{4pt}
\renewcommand{\arraystretch}{0.95}
\begin{tabular}{lrr@{\hspace{1.6em}}lrr}
\toprule
\textbf{Task} & \textbf{Pool} & \textbf{Used} &
\textbf{Task} & \textbf{Pool} & \textbf{Used} \\
\midrule
\multicolumn{6}{l}{\textit{FV abstractive ICL}} \\
\midrule
AG News & 7{,}600 & 5{,}000
& Antonym & 2{,}398 & 2{,}398 \\
Capitalize (Full Word) & 813 & 813
& Capitalize First Letter & 813 & 813 \\
Capitalize Last Letter & 814 & 814
& Capitalize Second Letter & 787 & 787 \\
Chinese $\rightarrow$ English & 981 & 981
& CommonsenseQA & 10{,}962 & 5{,}000 \\
Country-Capital & 197 & 189
& Country-Currency & 197 & 93 \\
French $\rightarrow$ English & 981 & 981
& Indonesian $\rightarrow$ English & 981 & 981 \\
Italian $\rightarrow$ English & 981 & 981
& Japanese $\rightarrow$ English & 981 & 981 \\
Landmark-Country & 836 & 836
& Lowercase First Letter & 814 & 814 \\
Lowercase Last Letter & 814 & 814
& National Parks & 451 & 451 \\
Next Capital Letter & 814 & 814
& Next Item & 225 & 225 \\
Park-Country & 749 & 732
& Person-Instrument & 510 & 510 \\
Person-Occupation & 821 & 821
& Person-Sport & 318 & 318 \\
Polish $\rightarrow$ English & 981 & 981
& Portuguese $\rightarrow$ English & 981 & 981 \\
Present-Past & 293 & 293
& Previous Item & 225 & 225 \\
Product-Company & 522 & 522
& Sentiment & 1{,}167 & 1{,}167 \\
Singular-Plural & 205 & 205
& Spanish $\rightarrow$ English & 981 & 981 \\
Synonym & 2{,}880 & 2{,}880
& Word Length & 814 & 814 \\
\midrule
\multicolumn{6}{l}{\textit{FV extractive ICL}} \\
\midrule
Adjective vs Verb (3) & 1{,}000 & 1{,}000
& Adjective vs Verb (5) & 1{,}000 & 1{,}000 \\
Alphabetically First (3) & 1{,}000 & 1{,}000
& Alphabetically First (5) & 1{,}000 & 1{,}000 \\
Alphabetically Last (3) & 1{,}000 & 1{,}000
& Alphabetically Last (5) & 1{,}000 & 1{,}000 \\
Animal vs Object (3) & 1{,}000 & 1{,}000
& Animal vs Object (5) & 1{,}000 & 1{,}000 \\
Choose First of 3 & 1{,}000 & 1{,}000
& Choose First of 5 & 1{,}000 & 1{,}000 \\
Choose Last of 3 & 1{,}000 & 1{,}000
& Choose Last of 5 & 1{,}000 & 1{,}000 \\
Choose Middle of 3 & 1{,}000 & 1{,}000
& Choose Middle of 5 & 1{,}000 & 1{,}000 \\
Color vs Animal (3) & 1{,}000 & 1{,}000
& Color vs Animal (5) & 1{,}000 & 1{,}000 \\
Concept vs Object (3) & 1{,}000 & 1{,}000
& Concept vs Object (5) & 1{,}000 & 1{,}000 \\
CoNLL: Location & 4{,}499 & 4{,}499
& CoNLL: Organization & 3{,}843 & 3{,}836 \\
CoNLL: Person & 3{,}544 & 3{,}539
& Fruit vs Animal (3) & 1{,}000 & 1{,}000 \\
Fruit vs Animal (5) & 1{,}000 & 1{,}000
& Object vs Concept (3) & 1{,}000 & 1{,}000 \\
Object vs Concept (5) & 1{,}000 & 1{,}000
& SQuAD Validation & 4{,}944 & 4{,}839 \\
Verb vs Adjective (3) & 1{,}000 & 1{,}000
& Verb vs Adjective (5) & 1{,}000 & 1{,}000 \\
\bottomrule
\end{tabular}
\vspace{-0.5em}
\label{app:dataset_info}
\end{table*}

\subsection{Translation Tasks}
\label{app:translation_tasks}
To perform translation evaluation, we use the multilingual translation dataset using the Open Multilingual Wordnet (OMW) 1.4 corpus via NLTK \citep{bird-loper-2004-nltk}. To select languages, we used W3Techs\footnote{\url{https://w3techs.com/technologies/overview/content_language}} and chose languages that are both present in OMW 1.4 and have above 1\% usage on the internet according to W3Techs. We select translation languages using internet usage statistics as a proxy for likely pretraining representation. Since W3Techs data update dynamically, our selection was based on June 2025 data. 981 parallel concepts are used for our tasks.  We use accuracy as our evaluation metric. For each synset, we use only the first lemma from the source language to predict the corresponding English translation. All evaluations are 5-shot, and we translate from each source language into English to ensure consistent token counts across all tasks. An example prompt is shown in Figure~\ref{fig:example_translation_prompt}. Compatible with the licensing terms, the data is only to be used for research purposes.

\begin{figure}[htbp]
\centering
\begin{tcolorbox}[width=0.9\linewidth, colback=gray!3, colframe=gray!30, boxrule=0.5pt]
\texttt{"boite de conserve" - "can"} \\
\texttt{"adresse" - "address"} \\
\texttt{"abus" - "maltreatment"} \\
\texttt{"bétail" - "livestock"} \\
\texttt{"argile" - "mud"} \\
\texttt{"chose" - "}
\end{tcolorbox}
\caption{French--English example. Accuracy uses prefix matching on the gold word (optional leading space). The continuation must equal the gold token or immediately follow it with a separator character: space, newline, tab, \texttt{'}, \texttt{,}, \texttt{.}, or end.}
\label{fig:example_translation_prompt}
\end{figure}

\subsection{Low Cardinality Label Space}
\label{app:low_cardinal_label_space}
\begin{figure}[htbp]
    \centering
    \includegraphics[width=\linewidth]{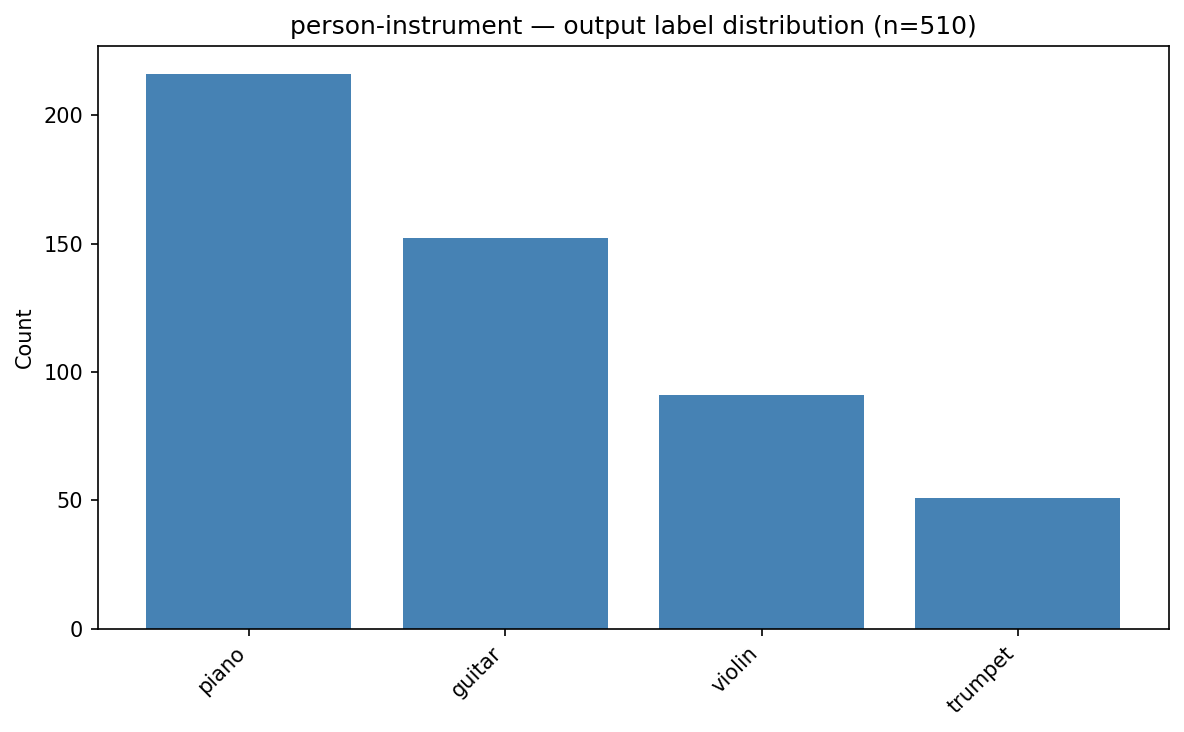}
    \caption{Label distribution of Person-Instrument task.}
\label{fig:person_instrument_dist}
\end{figure}

\begin{figure}[htbp]
    \centering
    \includegraphics[width=\linewidth]{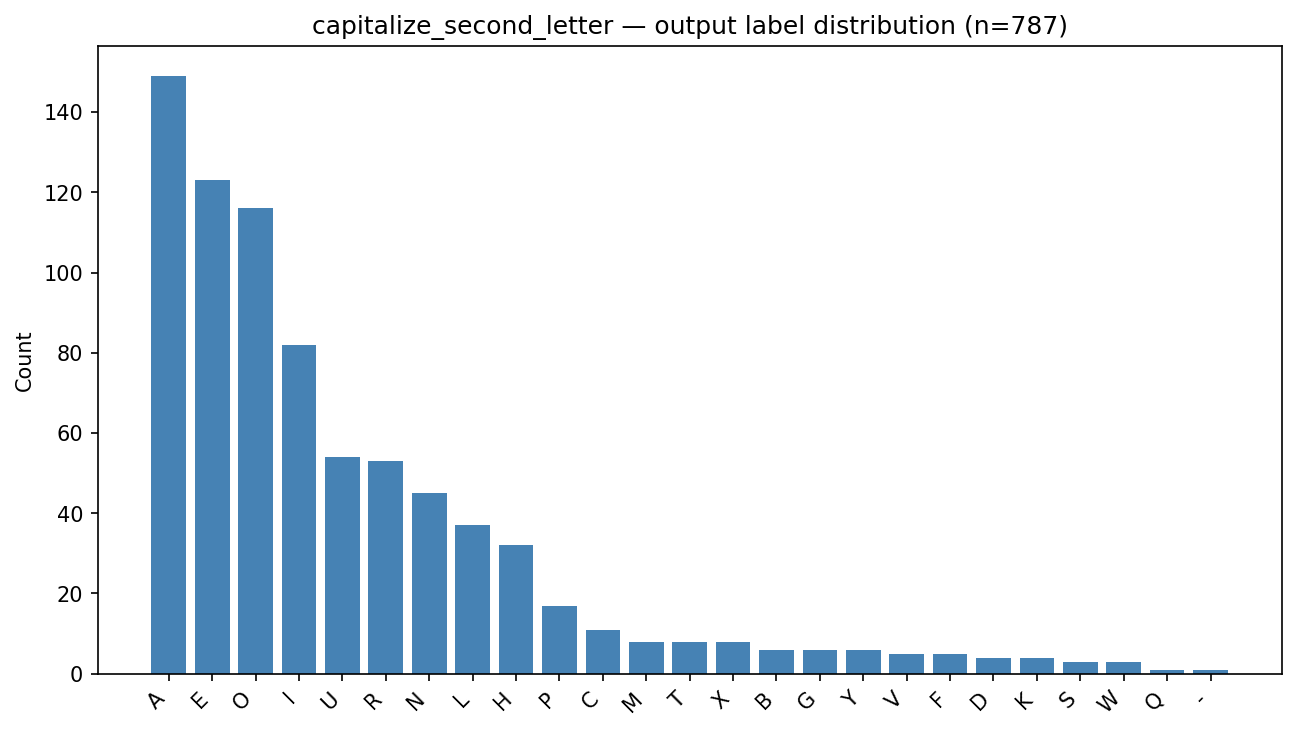}
    \caption{Label distribution of Capitalize Second Letter task.}
\label{fig:capitalize_second_letter_dist}
\end{figure}

Some function vector tasks have low cardinality label spaces, meaning the target prediction is likely to appear among the few-shot demonstrations. For instance, with 5-shot demonstrations and a 4-label task, a model could achieve correct predictions by distributional matching alone, simply sampling labels from the demonstrations. To disentangle genuine task understanding from this confound, we also evaluate with target exclusion. The Person-Instrument task has only 4 labels (Figure~\ref{fig:person_instrument_dist}), and Capitalize Second Letter, while nominally having 27 labels, is heavily skewed with the top 4 covering 59.7\% of samples (Figure~\ref{fig:capitalize_second_letter_dist}).

\section{Explored Training Variants}
\label{app:training_vars}

Before finalizing the \textsc{Hapax} training protocol, we explored several variants to understand the impact of inductive copying on learning abstractive ICL capabilities. The methods proposed here focus mainly on the first part of the induction circuit, namely the prefix-matching behavior. The motivation behind these methods is to eliminate the initial circuit component, which we expected would remove inductive copying behavior. We summarize these variants and discuss why they fail and are therefore not suitable for our purposes.

\subsection{Loss Penalty Based Interventions for Prefix-Matching Score}
To analyze the impact of inductive copying on the development of abstractive ICL capabilities, we included a loss function that penalizes the prefix-matching attention pattern commonly observed in induction heads. Given our loss function $\mathcal{L}$, we add the prefix-matching score across all heads to our loss function: \begin{equation}
\mathcal{L}_{\text{pm}}
= \lambda \, \frac{1}{L H} \sum_{l=1}^{L} \sum_{h=1}^{H}
\text{PrefixMatching}(l,h),
\end{equation}
where $\lambda$ is our hyperparameter for fine-tuning. We propose two different versions. In the first one, we compute the prefix-matching score on the synthetically generated random repetition sequence. The random sequences are given in separate forward passes along with regular training. In the second version, instead of using synthetic data, we compute the prefix-matching score on the training data itself. We use the same positions as we use to mask the loss in Section \ref{sec:training_pro}. 

\begin{figure}[htbp]
    \centering
\includegraphics[width=\linewidth]{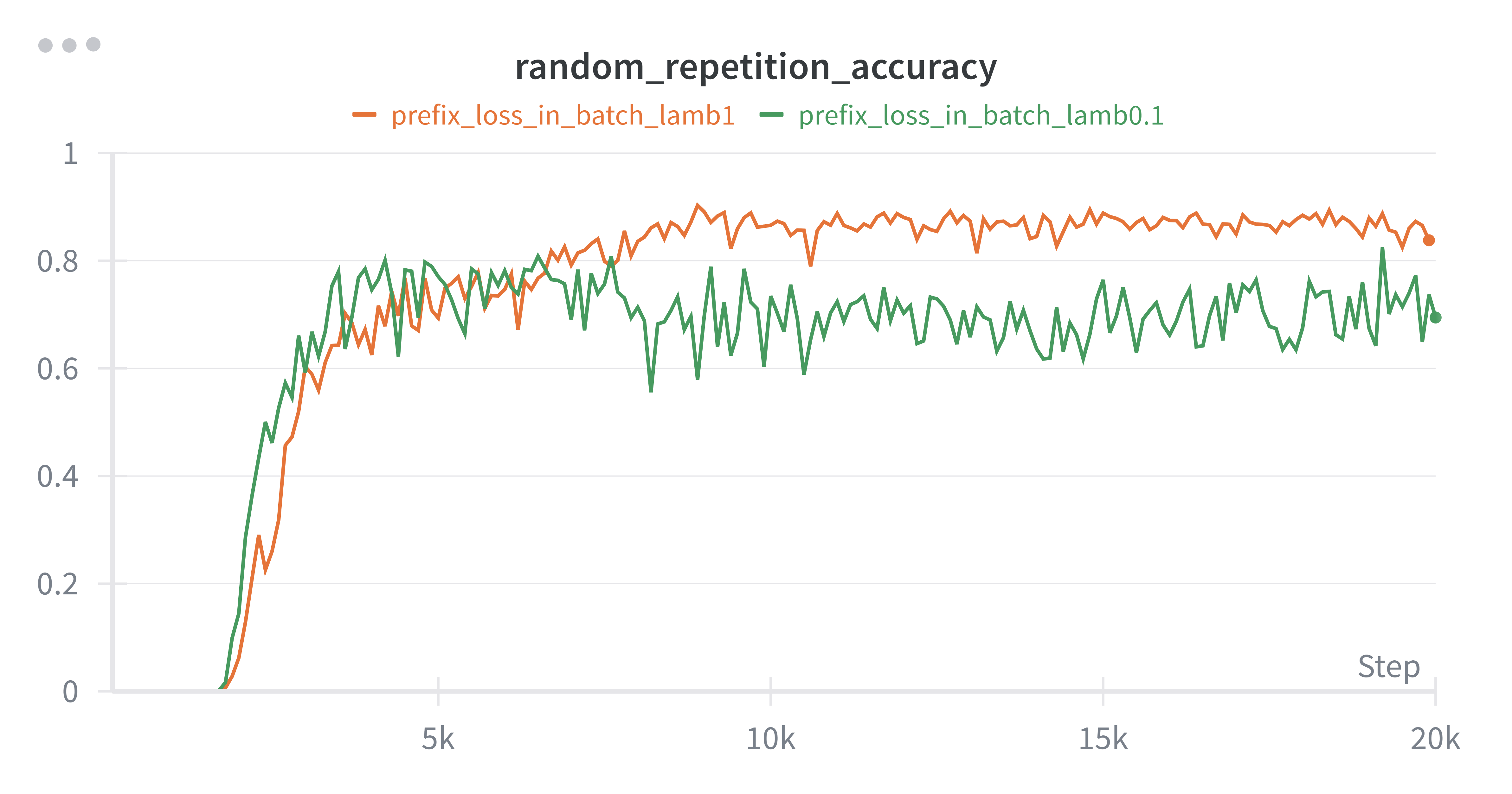}
    \caption{Random repetition accuracy for models trained with prefix-matching loss. Two lambda values that reduces the prefix-matching score are still able to learn random repetition.}
    \label{fig:prefix_loss_in_batch}
\end{figure}

\begin{figure}[htbp]
    \centering
\includegraphics[width=\linewidth]{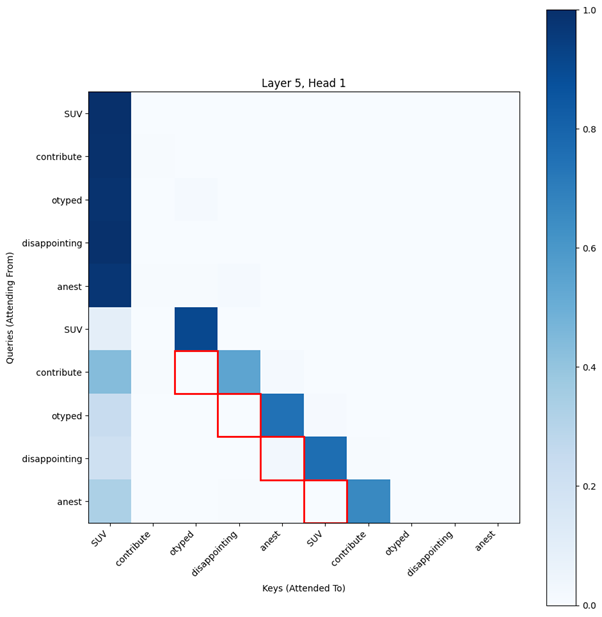}
    \caption{Example attention pattern of a model trained with a loss function that penalizes prefix-matching scores. Red squares indicate prefix-matching positions. Although the model does not display the typical attention pattern for induction, it learns to read the copying information from 1-offset positions.}
    \label{fig:prefix_loss_in_batch_attention_pattern}
\end{figure}

The initial approach overfits on the synthetic data and consequently, the prefix-matching pattern and copying is intact for natural language. The second approach reduces prefix-matching score for both synthetic and natural settings. However, although we observe that the loss function can control the prefix-matching score, the model can still do random copying. We give random repetition accuracies for two different lambda values in Figure \ref{fig:prefix_loss_in_batch} for the non-synthetic variation. When we analyze individual attention patterns on the random repetition task, we find that the model learns to copy from two-tokens ahead, instead of displaying the regular prefix-matching attention pattern. As an example, in Figure \ref{fig:prefix_loss_in_batch_attention_pattern}, red squares indicate the regular prefix-matching positions. However, when we penalize the attention pattern, the model can still read the information from the 1-offset positions. Therefore, attention-pattern based training interventions are ill-suited for our purposes. This observation also motivate us to focus on \textit{incentives}, as we did for \textsc{Hapax}, rather than directly optimizing for a specific behavior, since the model’s solution space tends to discover alternative ways to achieve the objective, similar to reward hacking.

\subsection{Reinitializing Induction Heads}
\begin{figure}[htbp]
    \centering
\includegraphics[width=\linewidth]{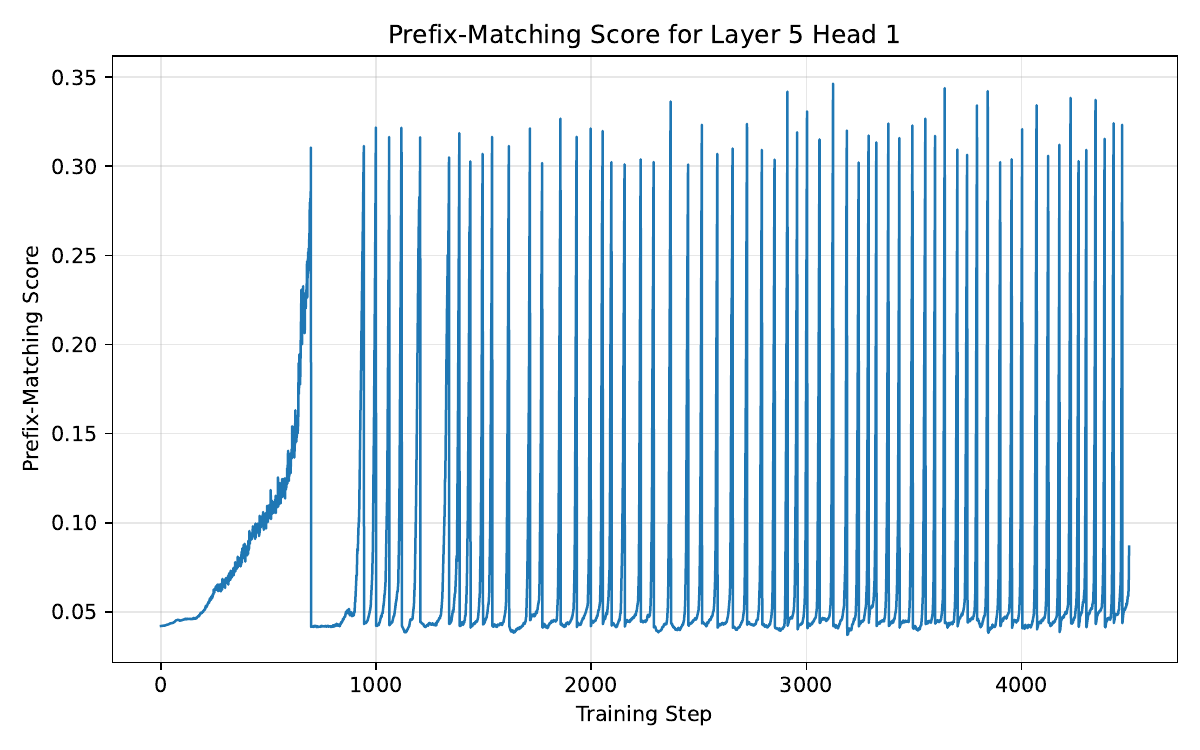}
    \caption{When the optimizer state for induction heads is  reinitialized in combination with the weights, the model gets stuck and tries to relearn induction heads. We plot prefix-matching scores before we reinitialize, which is why certain data points exceed 0.3. }
\label{fig:reinit_induction_and_optim}
\end{figure}

\begin{figure}[htbp]
    \centering
\includegraphics[width=\linewidth]{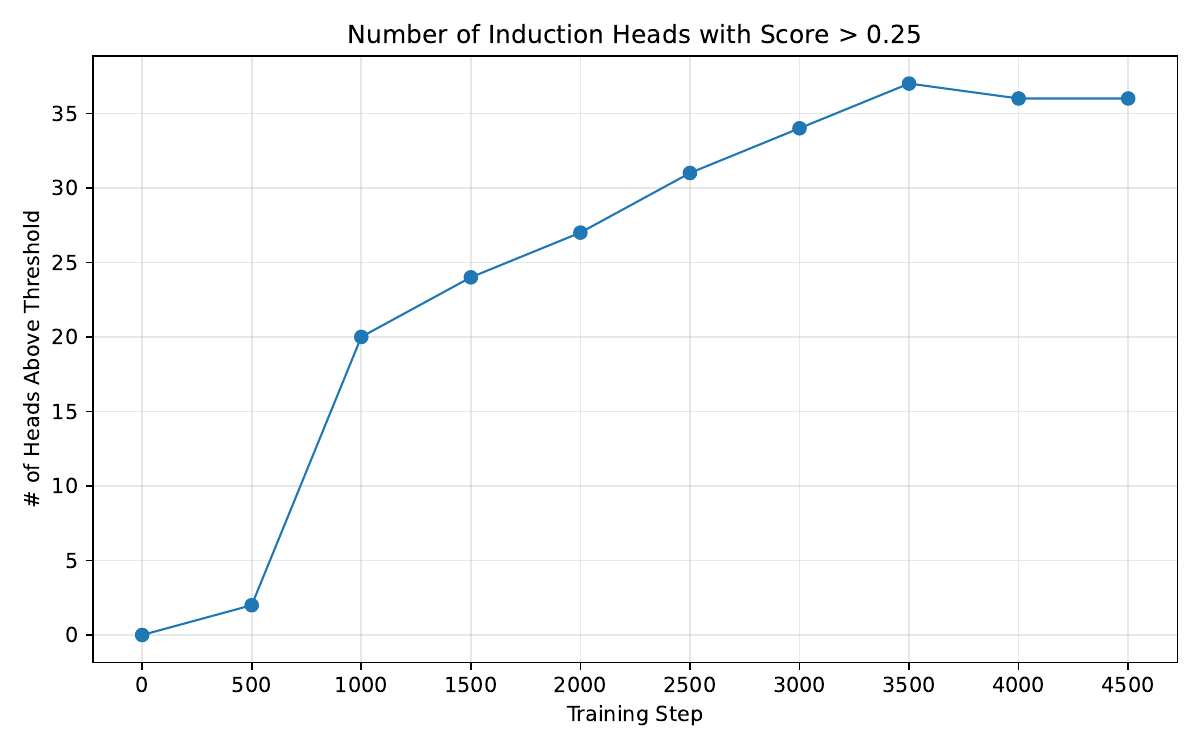}
    \caption{Number of induction heads that exceed the 0.25 threshold for the reinitialized model (160M). Since we reinitialize heads that exceed 0.3, we plot heads that exceed 0.25. The reinitialization causes the model to create three times more induction heads as compared to its vanilla version.}
\label{fig:num_heads_above_0.25_reinit}
\end{figure}

\begin{figure}[htbp]
    \centering
\includegraphics[width=\linewidth]{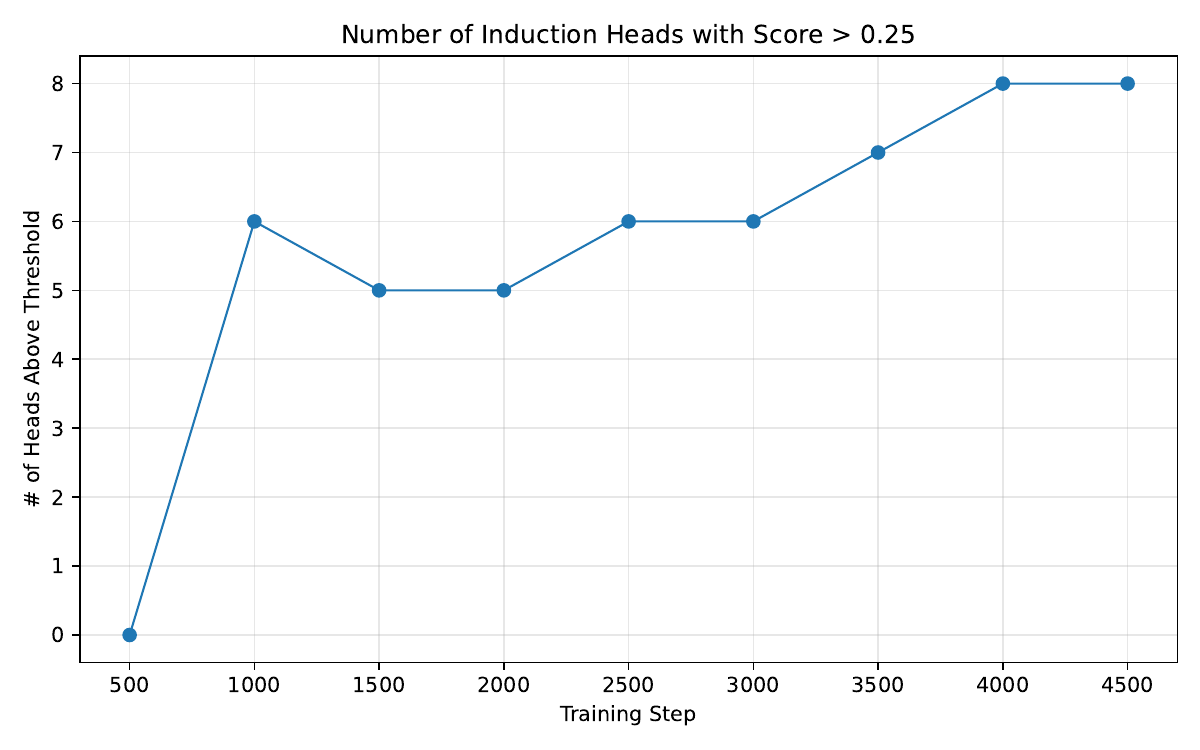}
    \caption{Number of attention heads that exceed 0.25 threshold for a vanilla model (160M).}
\label{fig:num_heads_above_0.25_vanilla}
\end{figure}

To analyze whether weight reinitialization would allow us to suppress inductive copying, we have created two different training protocols.  First, we reinitialized any attention head whose prefix-matching score was above a threshold $\tau$, which is 0.3 in our case. Secondly, in addition to reinitializing the weights, we have reinitialized the optimizer states for the corresponding attention heads that go above the threshold. For the first case, we have observed that the model can continue its training but the reinitialization creates more induction heads as compared to what it would learn in the vanilla version (Figure \ref{fig:num_heads_above_0.25_reinit}). When we also reinitialize the corresponding optimizer states, the model training procedure gets stuck (Figure \ref{fig:reinit_induction_and_optim}, \ref{fig:num_heads_above_0.25_vanilla}) and causes every attention head to have an increased prefix-matching score. Therefore, when we keep the same data distribution, the model tends to converge on similar solutions for copying behavior, which makes the presented methods unsuitable for our purposes. 

\begin{figure}[htbp]
    \centering
\includegraphics[width=\linewidth]{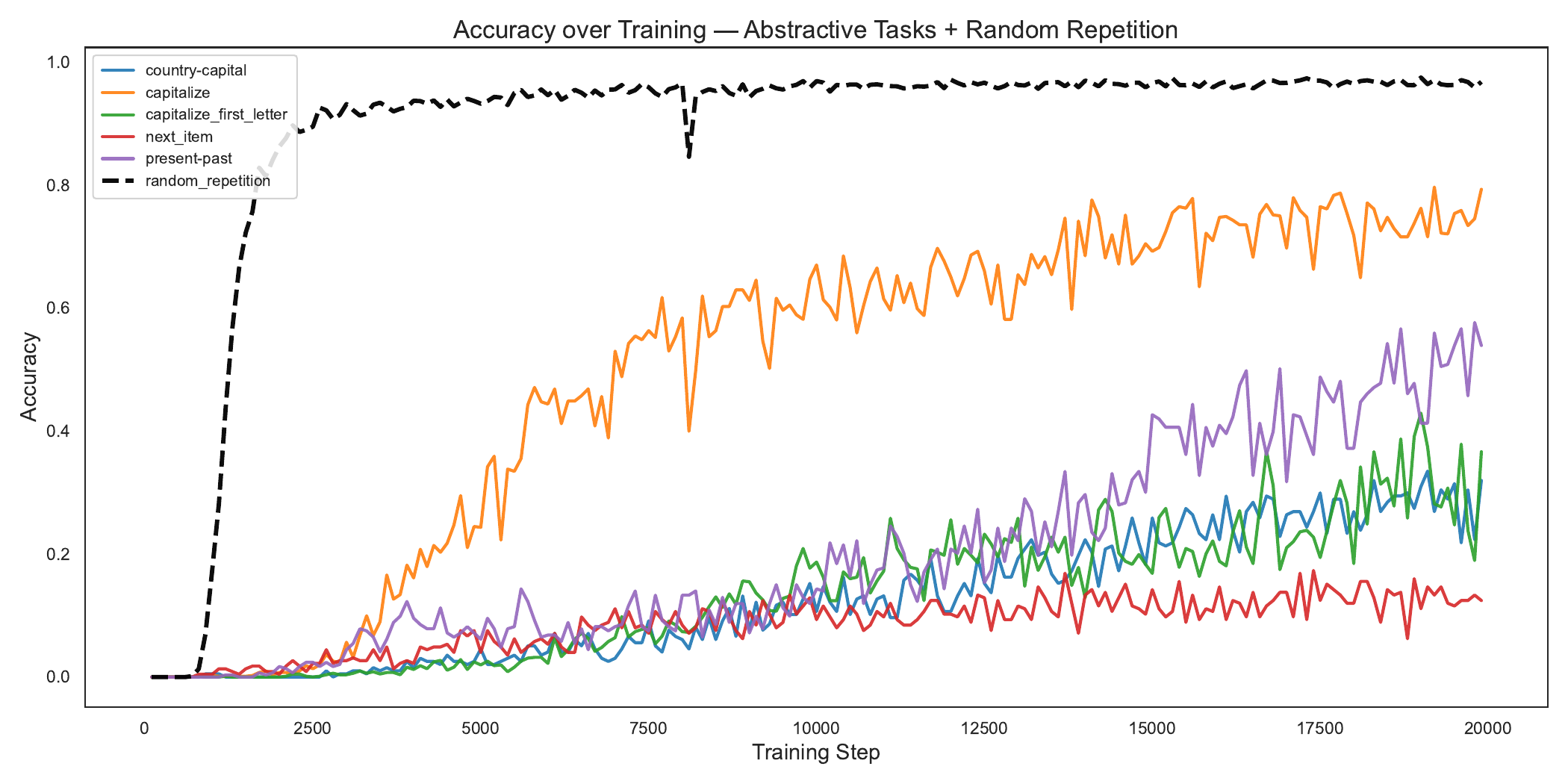}
    \caption{Accuracies of sample tasks over training for the Vanilla model. Random repetition accuracy is the first task to improve, and its rise coincides with the training step where the loss drops sharply (Figure~\ref{fig:validation_loss_comparison_1b}). This motivates examining whether the decrease in loss and the increase in inductive copying translate into meaningful gains on abstractive tasks, which according to our evidence are not tightly coupled.}
\label{fig:over_time_plot}
\end{figure}

\section{Choice of Statistical Testing}
\label{app:choice_of_statistical_test}

For downstream tasks where evaluation reduces to accuracy, each example yields a paired binary outcome across models (correct vs. incorrect). McNemar’s test is appropriate for this paired binary setting, as it assesses whether disagreements between the two models on the same examples are symmetric. Prior work has shown that McNemar’s test is well-calibrated for classifier comparison in such conditions \citep{10.1162/089976698300017197}. Accordingly, we use McNemar’s test for our main accuracy-based comparisons. To control the false discovery rate, we apply the Benjamini-Hochberg procedure at FDR level 0.05 within each table.

 \section{Additional Seed Runs}

In order to understand whether the emergence of abstractive ICL, prefix-matching scores, and TLD scores show similar qualitative patterns, we have trained two additional seeds, differing from the configuration specified in the official Pythia run. The first difference is that the actual specified seed is 1234, whereas now we use seeds 1337 and 2807. Secondly, due to compute limitations, our results do not use the same machine and configuration used for the main runs given within the paper. Differing from the main run and the original Pythia run, we have used ROCm 6.2.4, where we used the torch 2.5.1+rocm6.2 library instead of CUDA. Due to incompatibility with Flash Attention, SDPA has been used. The remaining libraries match the main run. On this hardware (MI250X), we experienced underflowing issues with fp16 training, so bf16 precision has been used throughout training. These runs took approximately 3030 MI250X-hours for 18000 steps. These runs, although not run in an equivalent environment to the main run, allow us to see whether the qualitative patterns hold.

\begin{figure*}[htbp]
    \centering       \includegraphics[width=\linewidth]{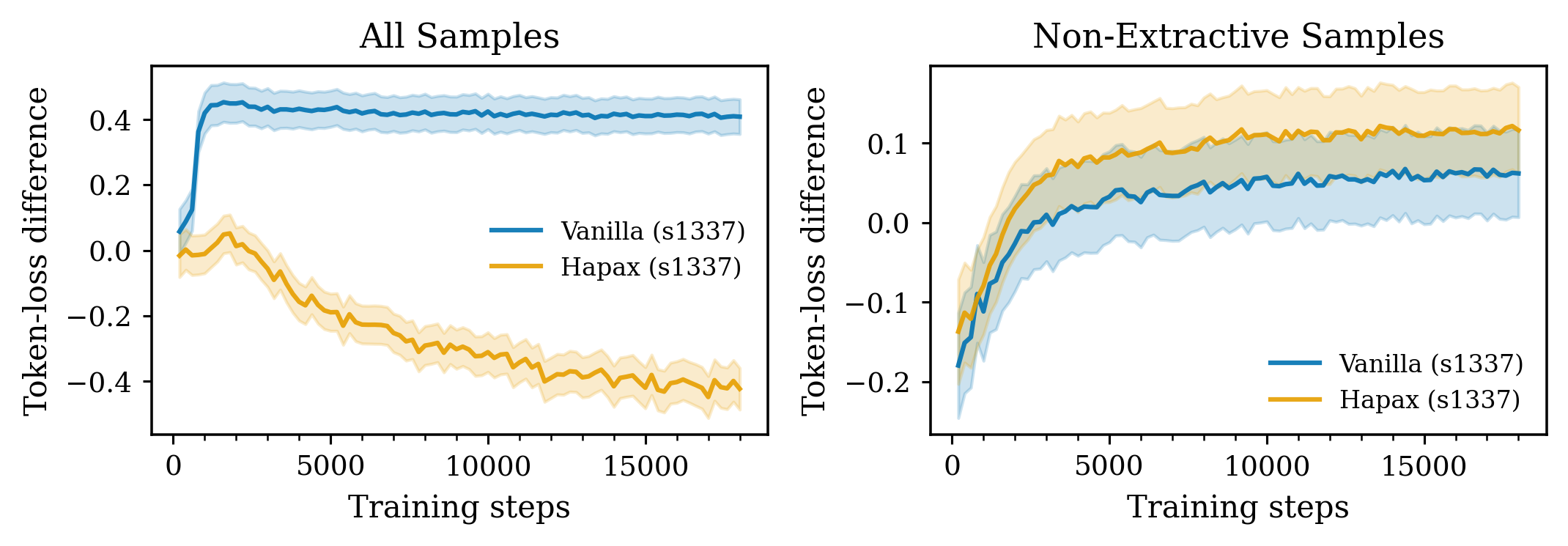}   
    \caption{Comparison of token-loss metrics across training for seed 1337.}
    \label{fig:tld_frontier_1b_seed1337}
\end{figure*}

\begin{figure*}[htbp]
    \centering       \includegraphics[width=\linewidth]{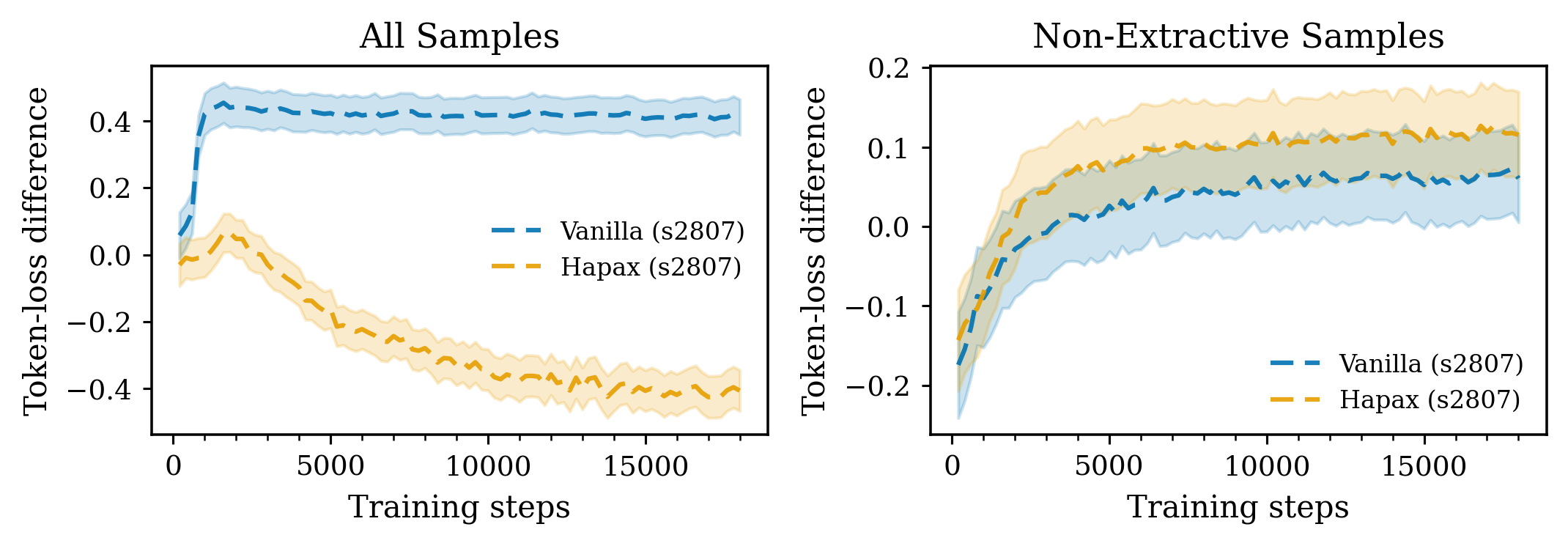}   
    \caption{Comparison of token-loss metrics across training for seed 2807.}
    \label{fig:tld_frontier_1b_seed2807}
\end{figure*}

\begin{figure*}[htbp]
    \centering       \includegraphics[width=\linewidth]{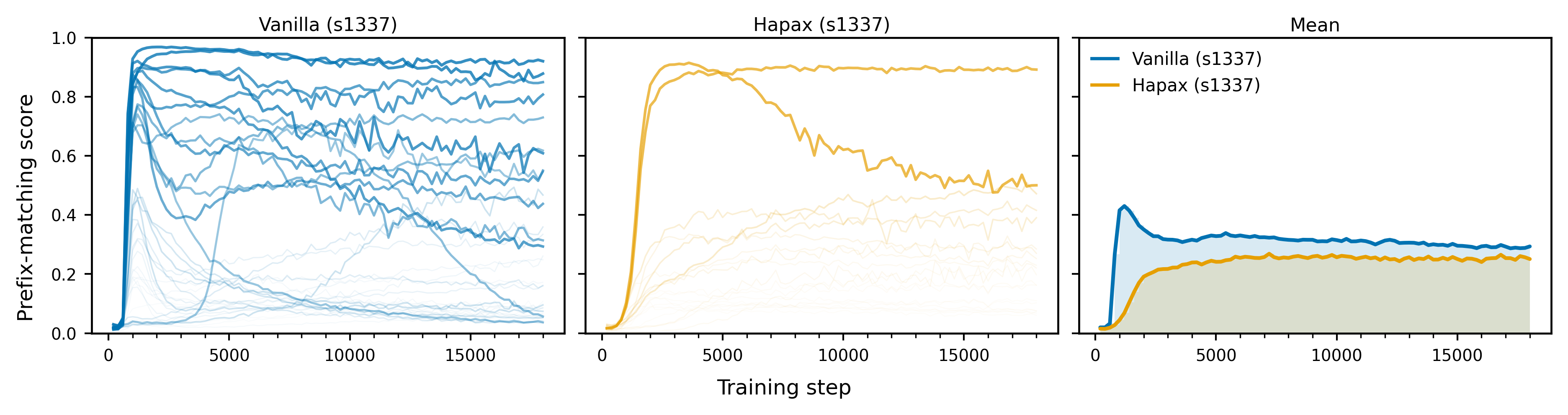}   
    \caption{Comparison of prefix-matching scores across training for seed 1337.}
    \label{fig:prefix_scores_comparison_1b-frontier-s1337}
\end{figure*}

\begin{figure*}[htbp]
    \centering       \includegraphics[width=\linewidth]{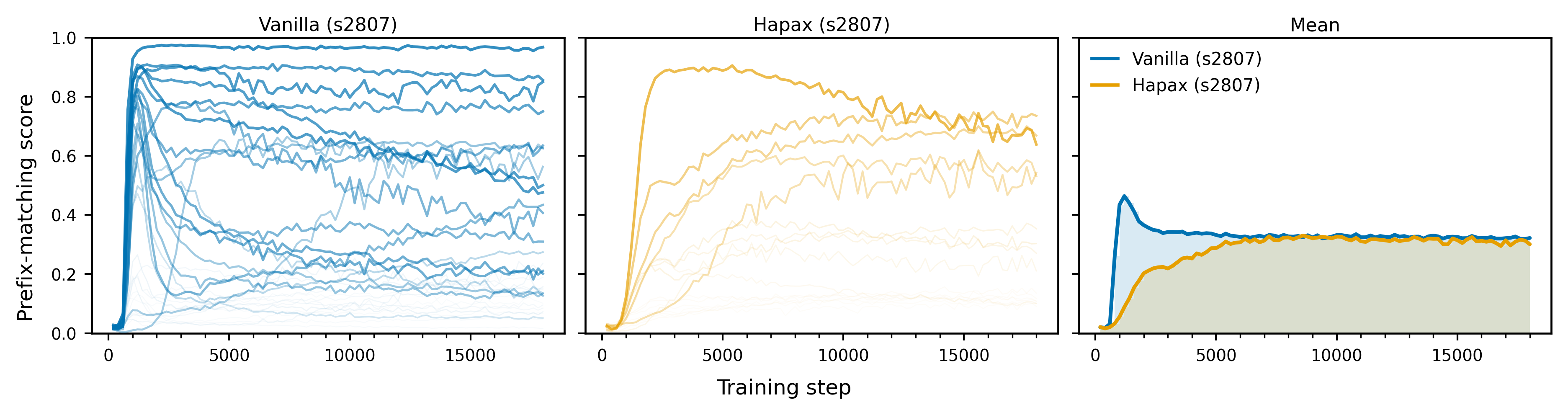}   
    \caption{Comparison of prefix-matching scores across training for seed 2807.}
    \label{fig:prefix_scores_comparison_1b-frontier-s2807}
\end{figure*}

For TLD in Figures~\ref{fig:tld_frontier_1b_seed1337} and~\ref{fig:tld_frontier_1b_seed2807}, Vanilla rises to a peak of 0.45 at step 1600 and then holds a plateau of 0.42 for the remainder of training, while HAPAX remains near zero through step 1800 and then declines steadily into negative values, in line with the decrease happening in Figure~\ref{fig:tld_plot}. As a check of the Figure~\ref{fig:induction_scores_over_time_comparison} experiment, we observe in Figures~\ref{fig:prefix_scores_comparison_1b-frontier-s1337} and~\ref{fig:prefix_scores_comparison_1b-frontier-s2807} that Vanilla training reproduces the rise-then-decay trajectory seen in Figure~\ref{fig:induction_scores_over_time_comparison}, with 8 and 9 (respectively for each seed) out of 128 heads reaching $>$80\% prefix-matching score, whereas HAPAX transitions gradually, with only 1 and 2 out of 128 heads exceeding 80\% at any point over the 18000 steps. Despite the trends observed in TLD and Prefix-Matching, the target-excluded curves of both models in Figure~\ref{fig:abstractive_icl_allseeds} emerge at similar times, where both remain below 3\% for target excluded until step 3000 and rise together thereafter. This indicates that HAPAX can develop abstractive capabilities despite the reduction in TLD and prefix-matching scores, in line with the pattern observed in Figure~\ref{fig:abstractive_icl_vanilla_vs_hapax_incl_excl_target_ci_step19900}. This gives additional evidence that HAPAX can suppress the phase transition and its associated metrics while still allowing for the emergence of abstractive ICL.

 \label{app:additional_seeds}

\section{2.8B Evaluations}
\label{app:2.8b_run}

\begin{figure*}[htbp]
    \centering
        \includegraphics[width=\linewidth]{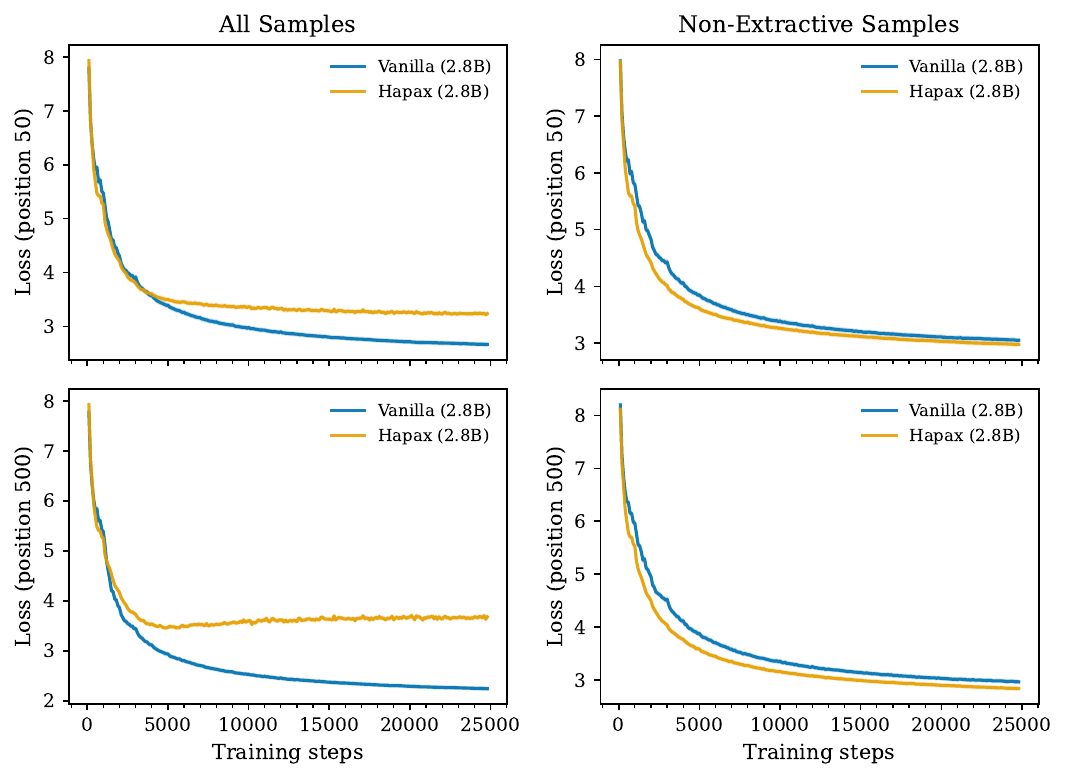}
    \caption{Loss values at individual token positions for 2.8B models. The loss value for Non-Extractive Samples is lower for both token positions than the Vanilla in \textsc{Hapax}.}
    \label{fig:position_losses_2.8b}
\end{figure*}

\begin{table*}[htbp]
  \centering
  \caption{Abstractive task performance summary for the \textsc{2.8b} model at step~24900 over 34 tasks. Macro-averaged accuracies (\%) within each family are shown with 95\% confidence intervals (normal approximation on the macro mean per category). \textit{\textbf{Sig.}} gives the number of tasks with significant accuracy differences under McNemar tests with joint Benjamini--Hochberg correction ($\alpha=0.05$). The \textit{\textbf{$>$ Vanilla}} column gives the number and fraction of those significant tasks where the corresponding variant has higher accuracy than vanilla. \textit{Standard} includes the target label in demonstrations; \textit{Target excl.} withholds the target label.}
  \label{tab:abstractive_task_summary_2.8b}

  \begin{threeparttable}
  \footnotesize
  \setlength{\tabcolsep}{3pt}
  \renewcommand{\arraystretch}{1.12}

  \begin{tabular*}{\linewidth}{@{\extracolsep{\fill}}lrrrr@{}}
    \toprule
    & \multicolumn{4}{c}{\textbf{Vanilla vs.\ \textsc{Hapax} (2.8B)}} \\
    \cmidrule(lr){2-5}
    \textbf{Task Category}
      & \textbf{Van.} & \textbf{\textsc{Hapax}} & \textbf{Sig.} & \textbf{$>$ Van.} \\
    \midrule

    \multicolumn{5}{@{}l}{\textit{Algorithmic} ($n=8$)} \\
    \quad Standard & 25.1 $\pm$ 0.9 & 30.0 $\pm$ 0.8 &  6 & 3/6\, (50\%) \\
    \quad Target excl. & 23.2 $\pm$ 0.8 & 34.8 $\pm$ 0.8 &  7 & 7/7\, (100\%) \\
    \addlinespace[4pt]
    \multicolumn{5}{@{}l}{\textit{Linguistic} ($n=4$)} \\
    \quad Standard & 36.1 $\pm$ 2.1 & 41.8 $\pm$ 1.9 &  3 & 3/3\, (100\%) \\
    \quad Target excl. & 36.1 $\pm$ 2.1 & 41.9 $\pm$ 1.9 &  3 & 3/3\, (100\%) \\
    \addlinespace[4pt]
    \multicolumn{5}{@{}l}{\textit{Knowledge} ($n=11$)} \\
    \quad Standard & 21.4 $\pm$ 1.3 & 24.9 $\pm$ 1.4 &  8 & 6/8\, (75\%) \\
    \quad Target excl. & 16.5 $\pm$ 1.2 & 28.3 $\pm$ 1.4 &  9 & 9/9\, (100\%) \\
    \addlinespace[4pt]
    \multicolumn{5}{@{}l}{\textit{Translation} ($n=8$)} \\
    \quad Standard & 9.6 $\pm$ 0.6 & 13.3 $\pm$ 0.7 &  7 & 7/7\, (100\%) \\
    \quad Target excl. & 9.6 $\pm$ 0.6 & 13.3 $\pm$ 0.7 &  7 & 7/7\, (100\%) \\
    \addlinespace[4pt]
    \multicolumn{5}{@{}l}{\textit{Sentence-level} ($n=3$)} \\
    \quad Standard & 41.2 $\pm$ 1.1 & 23.2 $\pm$ 1.0 &  3 & 0/3\, (0\%) \\
    \quad Target excl. & 12.7 $\pm$ 0.9 & 55.6 $\pm$ 0.9 &  3 & 3/3\, (100\%) \\

    \midrule
    \multicolumn{5}{@{}l}{\textbf{All} ($n=34$)} \\
    \quad Standard & 23.0 $\pm$ 0.6 & 25.2 $\pm$ 0.6 & 27 & 19/27\, (70\%) \\
    \quad Target excl. & 18.4 $\pm$ 0.5 & 30.3 $\pm$ 0.6 & 29 & 29/29\, (100\%) \\

    \bottomrule
    
  \end{tabular*}
  
  \end{threeparttable}
\end{table*}

\begin{figure}[htbp]
    \centering
    \includegraphics[width=1\linewidth]{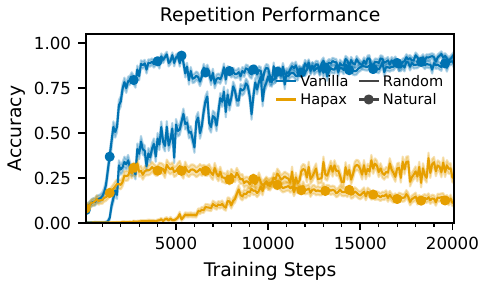}
    
    \caption{Repetition performance in both random token and natural text settings for 2.8B models. \textsc{Hapax} models struggle with repetition, which is solvable with induction circuits. This shows the \textsc{Hapax} training can control copying behavior. Accuracy is measured over 1000 randomly generated samples with $s = 25$.}
    \label{fig:random_repetition_combined_2.8b}
\end{figure}

\begin{figure}[tbp]
    \centering
    \includegraphics[width=1\linewidth]{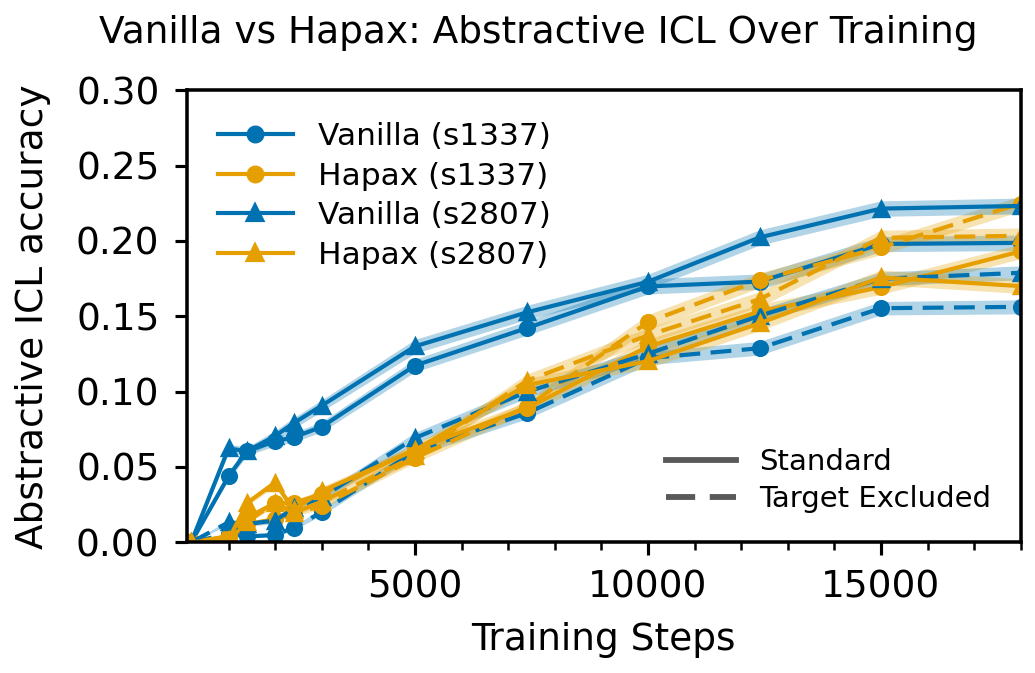}
    
    \caption{Comparison of abstractive-icl accuracy plots across training for seeds 1337 and 2807.}
    \label{fig:abstractive_icl_allseeds}
\end{figure}

\begin{table*}[t]
\centering
\caption{Performance comparison on \textbf{abstractive} in-context learning tasks (5-Shot) for Vanilla and \textsc{Hapax} 2.8B. 
Values show accuracy $\pm$ 95\% CI margin. Bold indicates higher performance; $\dagger$ denotes statistical significance ($p < 0.05$) according to McNemar's test.}
\label{tab:abstractive_results_2.8b}
\scriptsize
\begin{tabular}{lcc@{\hspace{1.8em}}lcc}
\toprule
\textbf{Task} & \textbf{Vanilla (\%)} & \textbf{\textsc{Hapax} (\%)} & \textbf{Task} & \textbf{Vanilla (\%)} & \textbf{\textsc{Hapax} (\%)} \\
\midrule
AG News & \textbf{42.0 $\pm$ 1.4}$^\dagger$ & 23.7 $\pm$ 1.2 & Antonym & 2.3 $\pm$ 0.6 & \textbf{7.9 $\pm$ 1.1}$^\dagger$ \\
Capitalize (Full Word) & 72.8 $\pm$ 3.1 & \textbf{81.3 $\pm$ 2.7}$^\dagger$ & Capitalize First Letter & 40.3 $\pm$ 3.4 & \textbf{60.8 $\pm$ 3.4}$^\dagger$ \\
Capitalize Last Letter & \textbf{7.2 $\pm$ 1.8}$^\dagger$ & 4.8 $\pm$ 1.5 & Capitalize Second Letter & \textbf{8.3 $\pm$ 1.9}$^\dagger$ & 4.1 $\pm$ 1.4 \\
Chinese\textrightarrow{English} & 11.0 $\pm$ 2.0 & \textbf{14.1 $\pm$ 2.2}$^\dagger$ & CommonsenseQA & \textbf{19.9 $\pm$ 1.1}$^\dagger$ & 17.0 $\pm$ 1.0 \\
Country-Capital & 48.7 $\pm$ 7.1 & \textbf{61.4 $\pm$ 6.9}$^\dagger$ & Country-Currency & 12.9 $\pm$ 6.8 & \textbf{21.5 $\pm$ 8.4} \\
French\textrightarrow{English} & 19.2 $\pm$ 2.5 & \textbf{25.1 $\pm$ 2.7}$^\dagger$ & Indonesian\textrightarrow{English} & 1.7 $\pm$ 0.8 & \textbf{2.3 $\pm$ 0.9} \\
Italian\textrightarrow{English} & 10.8 $\pm$ 1.9 & \textbf{14.7 $\pm$ 2.2}$^\dagger$ & Japanese\textrightarrow{English} & 2.9 $\pm$ 1.0 & \textbf{4.7 $\pm$ 1.3}$^\dagger$ \\
Landmark-Country & 31.3 $\pm$ 3.1 & \textbf{40.0 $\pm$ 3.3}$^\dagger$ & Lowercase First Letter & 55.0 $\pm$ 3.4 & \textbf{74.2 $\pm$ 3.0}$^\dagger$ \\
Lowercase Last Letter & \textbf{8.2 $\pm$ 1.9}$^\dagger$ & 5.5 $\pm$ 1.6 & National Parks & 24.6 $\pm$ 4.0 & \textbf{31.3 $\pm$ 4.3}$^\dagger$ \\
Next Capital Letter & 4.2 $\pm$ 1.4 & \textbf{4.5 $\pm$ 1.4} & Next Item & 14.7 $\pm$ 4.6 & \textbf{26.2 $\pm$ 5.7}$^\dagger$ \\
Park-Country & 10.7 $\pm$ 2.2 & \textbf{19.7 $\pm$ 2.9}$^\dagger$ & Person-Instrument & \textbf{21.0 $\pm$ 3.5}$^\dagger$ & 1.6 $\pm$ 1.1 \\
Person-Occupation & \textbf{17.5 $\pm$ 2.6}$^\dagger$ & 3.0 $\pm$ 1.2 & Person-Sport & 28.3 $\pm$ 5.0 & \textbf{37.1 $\pm$ 5.3}$^\dagger$ \\
Polish\textrightarrow{English} & 4.3 $\pm$ 1.3 & \textbf{5.9 $\pm$ 1.5}$^\dagger$ & Portuguese\textrightarrow{English} & 11.2 $\pm$ 2.0 & \textbf{17.0 $\pm$ 2.4}$^\dagger$ \\
Present-Past & 67.9 $\pm$ 5.3 & \textbf{71.3 $\pm$ 5.2} & Previous Item & 1.8 $\pm$ 1.7 & \textbf{5.3 $\pm$ 2.9} \\
Product-Company & 23.9 $\pm$ 3.7 & \textbf{26.4 $\pm$ 3.8} & Sentiment & \textbf{61.9 $\pm$ 2.8}$^\dagger$ & 28.8 $\pm$ 2.6 \\
Singular-Plural & 71.2 $\pm$ 6.2 & \textbf{82.9 $\pm$ 5.2}$^\dagger$ & Spanish\textrightarrow{English} & 15.7 $\pm$ 2.3 & \textbf{22.4 $\pm$ 2.6}$^\dagger$ \\
Synonym & 3.0 $\pm$ 0.6 & \textbf{4.9 $\pm$ 0.8}$^\dagger$ & Word Length & 4.3 $\pm$ 1.4 & \textbf{4.4 $\pm$ 1.4} \\
\bottomrule
\end{tabular}
\vspace{-0.5em}
\end{table*}

\begin{table*}[t]
\centering
\caption{Performance comparison on \textbf{abstractive} in-context learning tasks (5-Shot) for Vanilla and \textsc{Hapax} 2.8B by ensuring that the target answer for each test instance is not included in the 5-shot examples. 
Values show accuracy $\pm$ 95\% CI margin. Bold indicates higher performance; $\dagger$ denotes statistical significance ($p < 0.05$) according to McNemar's test.}
\label{tab:combined_hapax_results}
\scriptsize
\begin{tabular}{lcc@{\hspace{1.8em}}lcc}
\toprule
\textbf{Task} & \textbf{Vanilla (\%)} & \textbf{\textsc{Hapax} (\%)} & \textbf{Task} & \textbf{Vanilla (\%)} & \textbf{\textsc{Hapax} (\%)} \\
\midrule
AG News & 3.4 $\pm$ 0.5 & \textbf{35.9 $\pm$ 1.3}$^\dagger$ & Antonym & 2.4 $\pm$ 0.6 & \textbf{8.2 $\pm$ 1.1}$^\dagger$ \\
Capitalize (Full Word) & 72.8 $\pm$ 3.1 & \textbf{81.3 $\pm$ 2.7}$^\dagger$ & Capitalize First Letter & 35.9 $\pm$ 3.3 & \textbf{71.6 $\pm$ 3.1}$^\dagger$ \\
Capitalize Last Letter & 4.9 $\pm$ 1.5 & \textbf{8.1 $\pm$ 1.9}$^\dagger$ & Capitalize Second Letter & 1.9 $\pm$ 1.0 & \textbf{6.4 $\pm$ 1.7}$^\dagger$ \\
Chinese\textrightarrow{English} & 11.0 $\pm$ 2.0 & \textbf{14.1 $\pm$ 2.2}$^\dagger$ & CommonsenseQA & 14.0 $\pm$ 1.0 & \textbf{44.3 $\pm$ 1.4}$^\dagger$ \\
Country-Capital & 48.7 $\pm$ 7.1 & \textbf{61.4 $\pm$ 6.9}$^\dagger$ & Country-Currency & 14.0 $\pm$ 7.0 & \textbf{23.7 $\pm$ 8.6} \\
French\textrightarrow{English} & 19.2 $\pm$ 2.5 & \textbf{25.1 $\pm$ 2.7}$^\dagger$ & Indonesian\textrightarrow{English} & 1.7 $\pm$ 0.8 & \textbf{2.3 $\pm$ 0.9} \\
Italian\textrightarrow{English} & 10.8 $\pm$ 1.9 & \textbf{14.7 $\pm$ 2.2}$^\dagger$ & Japanese\textrightarrow{English} & 2.9 $\pm$ 1.0 & \textbf{4.7 $\pm$ 1.3}$^\dagger$ \\
Landmark-Country & 30.0 $\pm$ 3.1 & \textbf{40.9 $\pm$ 3.3}$^\dagger$ & Lowercase First Letter & 57.6 $\pm$ 3.4 & \textbf{87.6 $\pm$ 2.3}$^\dagger$ \\
Lowercase Last Letter & 5.3 $\pm$ 1.5 & \textbf{9.5 $\pm$ 2.0}$^\dagger$ & National Parks & 22.4 $\pm$ 3.8 & \textbf{33.3 $\pm$ 4.3}$^\dagger$ \\
Next Capital Letter & 3.8 $\pm$ 1.3 & \textbf{4.8 $\pm$ 1.5} & Next Item & 14.7 $\pm$ 4.6 & \textbf{26.2 $\pm$ 5.7}$^\dagger$ \\
Park-Country & 10.8 $\pm$ 2.2 & \textbf{19.0 $\pm$ 2.8}$^\dagger$ & Person-Instrument & 2.4 $\pm$ 1.3 & \textbf{5.3 $\pm$ 1.9}$^\dagger$ \\
Person-Occupation & \textbf{2.8 $\pm$ 1.1} & 2.6 $\pm$ 1.1 & Person-Sport & 20.8 $\pm$ 4.5 & \textbf{56.6 $\pm$ 5.4}$^\dagger$ \\
Polish\textrightarrow{English} & 4.3 $\pm$ 1.3 & \textbf{5.9 $\pm$ 1.5}$^\dagger$ & Portuguese\textrightarrow{English} & 11.2 $\pm$ 2.0 & \textbf{17.0 $\pm$ 2.4}$^\dagger$ \\
Present-Past & 67.9 $\pm$ 5.3 & \textbf{71.3 $\pm$ 5.2} & Previous Item & 0.4 $\pm$ 0.7 & \textbf{4.4 $\pm$ 2.7}$^\dagger$ \\
Product-Company & 14.9 $\pm$ 3.1 & \textbf{38.5 $\pm$ 4.2}$^\dagger$ & Sentiment & 20.6 $\pm$ 2.3 & \textbf{86.7 $\pm$ 1.9}$^\dagger$ \\
Singular-Plural & 71.2 $\pm$ 6.2 & \textbf{82.9 $\pm$ 5.2}$^\dagger$ & Spanish\textrightarrow{English} & 15.7 $\pm$ 2.3 & \textbf{22.4 $\pm$ 2.6}$^\dagger$ \\
Synonym & 3.0 $\pm$ 0.6 & \textbf{5.1 $\pm$ 0.8}$^\dagger$ & Word Length & 3.6 $\pm$ 1.3 & \textbf{8.8 $\pm$ 2.0}$^\dagger$ \\
\bottomrule
\end{tabular}
\vspace{-0.5em}
\end{table*}

\begin{table*}[t]
\centering
\caption{Performance comparison on \textbf{extractive} in-context learning tasks (5-Shot) for Vanilla and \textsc{Hapax} models. Values show accuracy $\pm$ 95\% CI margin. Bold indicates higher performance; $\dagger$ denotes statistical significance ($p < 0.05$) according to McNemar's test.}
\label{tab:extractive_hapax_results_2.8b}
\scriptsize
\begin{tabular}{lcc@{\hspace{1.8em}}lcc}
\toprule
\textbf{Task} & \textbf{Vanilla (\%)} & \textbf{\textsc{Hapax} (\%)} &
\textbf{Task} & \textbf{Vanilla (\%)} & \textbf{\textsc{Hapax} (\%)} \\
\midrule
Color vs Animal (3) & \textbf{56.6 $\pm$ 3.1}$^\dagger$ & 48.1 $\pm$ 3.1 & CoNLL: Location & \textbf{48.0 $\pm$ 1.5}$^\dagger$ & 35.1 $\pm$ 1.4 \\
Concept vs Object (3) & \textbf{76.7 $\pm$ 2.6}$^\dagger$ & 55.4 $\pm$ 3.1 & CoNLL: Person & \textbf{58.3 $\pm$ 1.8}$^\dagger$ & 12.1 $\pm$ 1.2 \\
Choose Middle of 3 & \textbf{40.8 $\pm$ 3.0}$^\dagger$ & 25.7 $\pm$ 2.7 & Alphabetically Last (3) & \textbf{31.5 $\pm$ 2.9}$^\dagger$ & 25.0 $\pm$ 2.7 \\
Alphabetically Last (5) & \textbf{19.2 $\pm$ 2.4}$^\dagger$ & 14.9 $\pm$ 2.2 & Object vs Concept (3) & \textbf{58.0 $\pm$ 3.1}$^\dagger$ & 37.6 $\pm$ 3.0 \\
Object vs Concept (5) & \textbf{57.6 $\pm$ 3.1}$^\dagger$ & 31.5 $\pm$ 2.9 & Fruit vs Animal (5) & \textbf{22.1 $\pm$ 2.6}$^\dagger$ & 7.8 $\pm$ 1.7 \\
Animal vs Object (3) & \textbf{50.5 $\pm$ 3.1}$^\dagger$ & 34.8 $\pm$ 3.0 & Verb vs Adjective (3) & \textbf{77.7 $\pm$ 2.6}$^\dagger$ & 64.4 $\pm$ 3.0 \\
Fruit vs Animal (3) & \textbf{37.4 $\pm$ 3.0}$^\dagger$ & 14.3 $\pm$ 2.2 & Choose First of 5 & \textbf{77.1 $\pm$ 2.6}$^\dagger$ & 42.0 $\pm$ 3.1 \\
Choose First of 3 & \textbf{72.3 $\pm$ 2.8}$^\dagger$ & 32.6 $\pm$ 2.9 & Choose Last of 3 & \textbf{63.7 $\pm$ 3.0}$^\dagger$ & 43.0 $\pm$ 3.1 \\
Color vs Animal (5) & \textbf{49.4 $\pm$ 3.1}$^\dagger$ & 41.5 $\pm$ 3.1 & Verb vs Adjective (5) & \textbf{68.1 $\pm$ 2.9}$^\dagger$ & 53.8 $\pm$ 3.1 \\
SQuAD Validation & \textbf{17.1 $\pm$ 1.1}$^\dagger$ & 10.6 $\pm$ 0.9 & CoNLL: Organization & \textbf{47.5 $\pm$ 1.7}$^\dagger$ & 17.0 $\pm$ 1.3 \\
Choose Last of 5 & \textbf{71.1 $\pm$ 2.8}$^\dagger$ & 45.6 $\pm$ 3.1 & Adjective vs Verb (3) & \textbf{54.6 $\pm$ 3.1}$^\dagger$ & 27.6 $\pm$ 2.8 \\
Alphabetically First (3) & \textbf{32.5 $\pm$ 2.9}$^\dagger$ & 23.4 $\pm$ 2.6 & Animal vs Object (5) & \textbf{38.9 $\pm$ 3.0}$^\dagger$ & 25.4 $\pm$ 2.7 \\
Concept vs Object (5) & \textbf{65.6 $\pm$ 2.9}$^\dagger$ & 47.8 $\pm$ 3.1 & Choose Middle of 5 & \textbf{20.8 $\pm$ 2.5}$^\dagger$ & 12.5 $\pm$ 2.0 \\
Adjective vs Verb (5) & \textbf{43.4 $\pm$ 3.1}$^\dagger$ & 18.1 $\pm$ 2.4 & Alphabetically First (5) & \textbf{20.9 $\pm$ 2.5}$^\dagger$ & 14.3 $\pm$ 2.2 \\
\bottomrule
\end{tabular}
\vspace{-0.5em}
\end{table*}

\begin{figure*}[h]
    \centering
    \includegraphics[width=\linewidth]{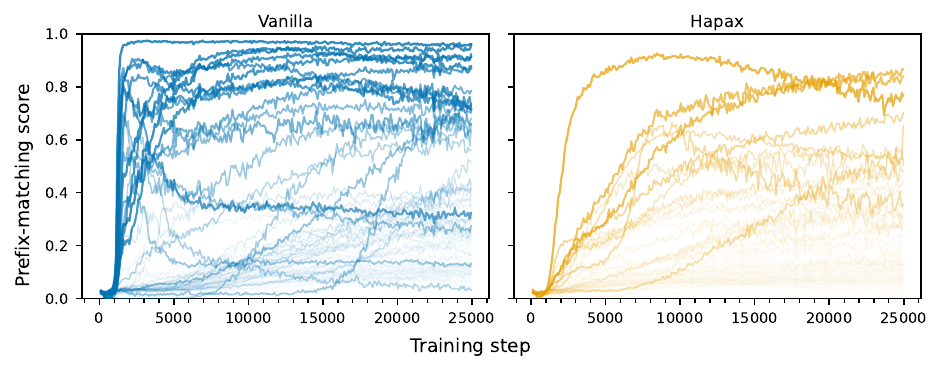}
    \caption{Prefix-matching scores of attention heads across training for 2.8B. Each line is one head. Opacity is proportional to peak score attained throughout training (heads with peak $<0.1$ omitted).}
    \label{fig:induction_scores_comparison_2.8b.pdf}
\end{figure*}

\begin{figure*}[htbp]
    \centering       \includegraphics[width=\linewidth]{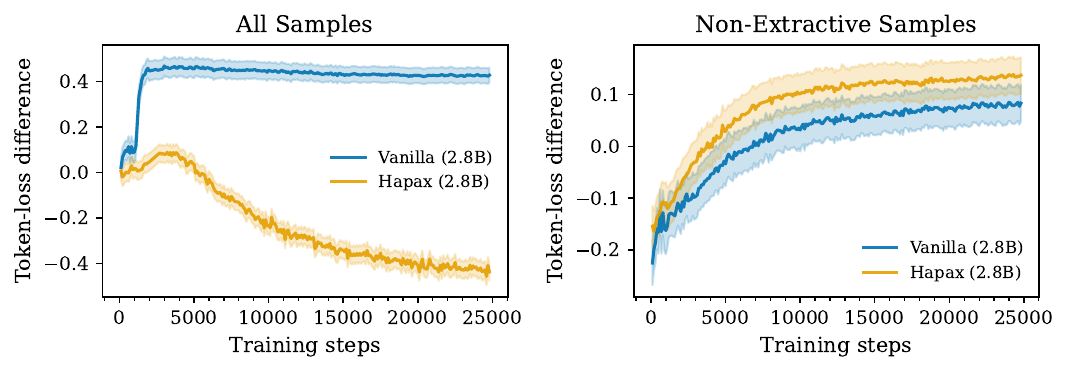}   
    \caption{Comparison of token-loss metrics across training on 2.8B models. Unlike the regular (All Samples) TLD, our proposed variants on non-extractive samples progress gradually, with each \textsc{Hapax} variant being able to achieve non-trivial ICL capabilities on its corresponding distribution. Since the regular distribution is dominated by induction-predictable positions, its gains primarily reflect exact copying rather than abstractive ICL. Computing the metric on targeted subsets thus reveals distinct emergence patterns across the distributions of interest. Metrics are computed on randomly sampled validation data.}
    \label{fig:tld_plot_2.8b}
\end{figure*}

We train additional Vanilla and \textsc{Hapax} models with 2.8B parameters to evaluate whether the preservation of abstractive capabilities is also observed at a larger model scale. The models are trained on 52.4B tokens; the compute-optimal budget estimated by \citet{NEURIPS2022_c1e2faff} is approximately 56B tokens. Due to compute constraints, we omit the Thresholded-\textsc{Hapax} variant from the 2.8B experiments. We use the same setting as in our 1B training, but train for 25K steps. Under standard evaluation, Vanilla achieves 23.0\% accuracy, whereas \textsc{Hapax} achieves 25.2\% (Table~\ref{tab:abstractive_task_summary_2.8b}). These results suggest that the preservation of abstractive capabilities under \textsc{Hapax} is also observed in this 2.8B setting, despite reduced copying and a degradation of the TLD metric (Figures~\ref{fig:position_losses_2.8b}--\ref{fig:tld_plot_2.8b}).

\section{Responsible NLP Checklist Information}

\subsection{Statistics for Data}
We discuss relevant dataset statistics in Section \ref{sec:training_pro} and  Appendix \ref{sec:experimental_detail}. Furthermore, we used a 80\%/20\% split for the training and validation data.

\subsection{Model Size and Budget}
We report the model sizes and training setup in Section~\ref{sec:training_pro}. Each 1B model was trained on 8 A100 GPUs for 72 hours, corresponding to approximately 576 A100 GPU-hours per run. Each 2.8B model was trained on 8 A100 GPUs for 168 hours, corresponding to approximately 1344 A100 GPU-hours per run.

\subsection{Artifact Use \& Licensing}
The Pile \citep{gao2020pile800gbdatasetdiverse} and the Function Vectors tasks \citep{todd2024function} are subject to the MIT License. In accordance with the licensing terms, these are solely used for research purposes. 

\subsection{AI Usage}
In accordance with the author guidelines, we made use of LLM-based services such as Claude as assistive tools for correcting grammar and refine writing.

\end{document}